\RequirePackage{fix-cm}
\documentclass[natbib,smallcondensed]{svjour3}     
\smartqed  
\usepackage{graphicx}
\usepackage{amsmath}
\usepackage{latexsym}   
\usepackage[utf8]{inputenc}
\usepackage{booktabs} 
\usepackage{footmisc}
\usepackage{pgfplots}
\usepackage{amssymb}
\usepackage{mathtools}
\usepackage{commath}
\usepackage{amsfonts}

\usepackage[flushleft]{threeparttable}
\usepackage[colorlinks=true,linkcolor=blue,citecolor=blue]{hyperref}  
\newcommand{\mycmmnt}[1]{}

\newcommand{\savefn}[2]{\footnote{\label{#1}#2}}
\newcommand{\reffn}[1]{\textsuperscript{\ref{#1}}}

\newcommand\ul[1]{\underline{#1}}

\makeatletter
\newcommand{\VATEX}{\textsc{VaTeX}}
\makeatother

\newcommand{\bfemph}[1]{\textbf{#1}}
\renewcommand{\emph}[1]{\bfemph{#1}}
\newcommand{\ie}{\textit{i}.\textit{e}., }
\newcommand{\eg}{\textit{e}.\textit{g}., }

\begin{document}

\title{
A Comprehensive Review of the\\Video-to-Text Problem
\thanks{This work has been done as part of the Stic-AmSud Project 18-STIC-09, ``Transforming multimedia data for indexing and retrieval purposes''.
Jesus Perez-Martin is funded by ANID/Doctorado Nacional/2018-21180648.
This work was partially supported by the ANID - Millennium Science Initiative Program - Code ICN17\_002, the Department of Computer Science at University of Chile, and the Image and Multimedia Data Science Laboratory (IMScience) at PUC Minas.}
}


\author{Jesus Perez-Martin\and
        Benjamin Bustos\and
        Silvio Jamil F. Guimar\~{a}es\and
        Ivan~Sipiran\and
        Jorge~Pérez\and
        Grethel~Coello~Said
}


\institute{Jesus~Perez-Martin, Benjamin~Bustos, and Jorge~Pérez \at
              IMFD, Department of Computer Science, University of Chile \at
              Beauchef 851, Santiago, Chile \\
              \email{jeperez@dcc.uchile.cl, bebustos@dcc.uchile.cl, jperez@dcc.uchile.cl}  
        \and
            Silvio~Jamil~F.~Guimar\~{a}es \at
              Computer Science Department, Pontifical Catholic University of Minas Gerais \at
              Belo Horizonte, Minas Gerais, Brazil \\
              \email{sjamil@pucminas.br}
        \and
            Ivan~Sipiran, and Grethel~Coello~Said \at
              Department of Computer Science, University of Chile \at
              Beauchef 851, Santiago, Chile \\
              \email{isipiran@dcc.uchile.cl, grethelyusel@gmail.com}  
}

\date{Received: date / Accepted: date}

\maketitle

\begin{abstract}
    Research in the Vision and Language area encompasses challenging topics that seek to connect visual and textual information.
    When the visual information is related to videos, this takes us into Video-Text Research, which includes several challenging tasks such as video question answering, video summarization with natural language, and video-to-text and text-to-video conversion.
    This paper reviews the video-to-text problem, in which the goal is to associate an input video with its textual description.
    This association can be mainly made by retrieving the most relevant descriptions from a corpus or generating a new one given a context video.
    These two ways represent essential tasks for Computer Vision and Natural Language Processing communities, called \textit{text retrieval from video task} and \textit{video captioning/description task}.
    These two tasks are substantially more complex than predicting or retrieving a single sentence from an image. 
    The spatiotemporal information present in videos introduces diversity and complexity regarding the visual content and the structure of associated language descriptions. 
    This review categorizes and describes the state-of-the-art techniques for the video-to-text problem.
    It covers the main video-to-text methods and the ways to evaluate their performance.
    We analyze twenty-six benchmark datasets, showing their drawbacks and strengths for the problem requirements.
    We also show the progress that researchers have made on each dataset, we cover the challenges in the field, and we discuss future research directions.
    \keywords{Vision-and-Language \and Video-to-text \and Video captioning \and Video description retrieval \and Matching-and-ranking \and Deep learning \and Joint multi-modal embedding \and Visual-semantic embedding \and Visual-syntactic embedding}
\end{abstract}

\section{The Video-to-Text Problem}\label{sec:introduction}


The form of communication that we humans use the most is natural language.
It is essential that systems such as interactive Artificial Intelligence (AI) and helper robots be capable of generating text, and many applications are developed to automatically generate text from non-linguistic data.
Natural Language Generation (NLG) is characterized by \citet{Reiter2000BuildingSystems} as the production of understandable texts from some underlying non-linguistic representation of information.
This definition of NLG is usually associated with the \emph{data-to-text generation} \citep{Eisenstein2019IntroductionProcessing, Gatt2018SurveyEvaluation}, assuming the exact input can vary substantially.
Today, text generation from unstructured perceptual input, such as a raw image or video, has become an important challenge.
In this review, we specifically tackle the NLG from videos 
as a fundamental problem to bridge vision and language.

The Web is the largest multimedia repository.
The search and analysis of documents (\eg audio, video, image, and 3D objects) have increased, leading to the current interest in research in this field. 
Notably, researchers devoted considerable effort to analyze and retrieve video content, being \emph{video understanding}, one of the multimedia areas that constitute an open field of research. 
Advances in computer technology make the user experience 
a priority. 


Specifically, the automatic generation of natural language descriptions of videos poses a challenge for computer vision and multimedia information retrieval communities. 
Solving it can be useful for video indexing and retrieval, surveillance systems~\citep{Sah2019UnderstandingCaptioning} (real-time generation of video captions for several simultaneous cameras), robotics (answering questions about the environment), sign language translation, and assistance for the visually impaired. 
For instance, the automatic \emph{video description/caption generation} would significantly reduce the problem of video retrieval from text to a two-step process: (1) the generation of textual stories from each video in the dataset, and (2) the text-similarity search between the query and video-related descriptions. 
This kind of model might help users filter what is attractive to them among the videos on YouTube, searching the videos by their textual stories.
However, to deal with the retrieval problem, researchers have proposed other better techniques analyzed in this review. 

\begin{figure}[t]
\centering
\includegraphics[width=.8\textwidth]{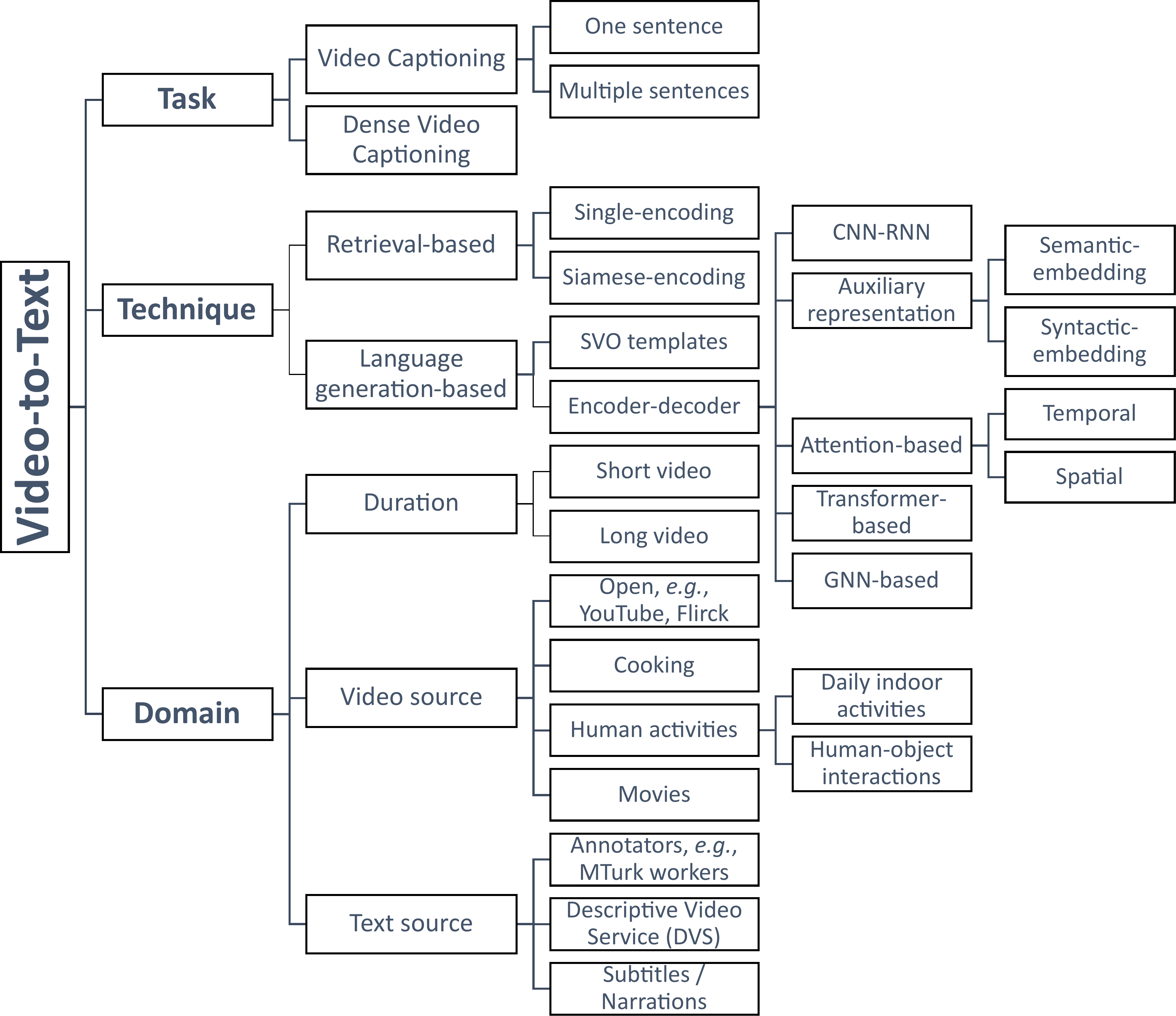}
\caption{
The Video-to-Text problem, tasks, techniques, and domains.
In this review we cover the video-to-text problem from the two fundamental techniques by which it has been addressed: retrieval-based (matching-and-ranking-based) and language generation-based (NLG-based).
}
\label{fig:visual-content-text}
\end{figure}

In Multimedia Information Retrieval, several tasks are based on relating the multimedia data and natural language.
When these tasks are focused on any kind of visual content, \eg images and videos, we lead with a recent research field, the automatic \emph{vision-language intersection}. 
In this field, to bridge vision and language, we can, for example, generate text from visual contents or vice versa, and determine (from a dataset or corpus) the most relevant multimedia data to a text query or vice versa. 
When these tasks receive videos as input, we are dealing with the \emph{video-to-text (VTT)} problem.
Figure \ref{fig:visual-content-text} shows a detailed categorization of VTT from three aspects: tasks, techniques, and domains.

The VTT research has currently grown in interest thanks to the massive success of deep learning in Computer Vision and Natural Language Processing (NLP). 
Deep learning is state-of-the-art for several vision tasks, such as activity recognition on videos~\citep{Kong2018HumanSurvey} and object detection~\citep{Ren2017FasterNetworks}.
However, these models require extensive training data to obtain high performance and are usually pre-trained on some large-scale action recognition and video classification datasets.
The recent creation of more than twenty-five \emph{video-text datasets} and benchmarks has partially removed the impediment of the lack of large-scale annotated video datasets for addressing the VTT problem.
Likewise, some institutions have proposed competitions to evaluate the VTT methods, in which the teams are asked to submit results for a set of videos. 

One of the first video-text datasets was proposed by~\citet{L.Chen2011CollectingEvaluation} in 2011, which consists of 1970 videos extracted from YouTube.  
After that, many techniques for the automatic acquisition of videos and the collection of most representative descriptions have been developed, increasing data and contributing to more diversity and realism.
In 2018, \citet{Mahdisoltani2018OnLearning} presented the Something-Something dataset, one of the largest, with 20,847 videos. 
Almost all existing large-scale video captioning/description datasets are monolingual (English), and the proposed VTT methods are restricted to English. 
However, some datasets like \VATEX{}~\citep{Wang2019VaTeXResearch} have driven the study of \emph{multilingual video-to-text} (English and Chinese), which is very important to achieve a real application of these methods worldwide.

As we will discuss in this review, the VTT problem can be mainly addressed by two types of methods: \emph{matching-and-ranking-based} techniques or \emph{NLG-based} techniques.
Although the way to evaluate the results in both categories is not the same, the metrics used in both cases come from other tasks related to the VTT problem.
In the literature, authors have reported some automatic assessments, and the majority of them have come from the metrics used for information retrieval, image captioning, and machine translation tasks.
Researchers quantitatively evaluate the models based on natural language correctness and the semantics' relevance to the respective input videos, which is not an easy task~\citep{Celikyilmaz2020EvaluationSurvey}. 
There is no standard evaluation method, and some metrics have not shown sufficient robustness for the \textit{Video Description/Captioning} task compared to the expensive human evaluation.


In the rest of this section we present the VTT tasks and techniques relevant to this review.
First, we present the automatic generation of natural language descriptions from videos, also known as \emph{video captioning} or \emph{video description}.
Next, we present two other tasks: 
the \emph{text retrieval from video} task (Section~\ref{sec:video-matching}) and the \emph{dense video captioning/description} task (Section~\ref{sec:dense-video-captioning}).
Finally, in Section~\ref{sec:other-tasks}, we mention other \emph{vision-text tasks}, which are not part of the scope of this review but can be important as complementary tasks to speed up and improve the quality of VTT solutions significantly. 


\medskip

\noindent
{\bf Outline of the document:}
The next three sections offer an overview of the fundamentals of neural models for two essential tasks of VTT, analyzing the principal strategies and models covered in the literature for description generation and matching-and-ranking techniques.
In particular, in Section~\ref{sec:visual-features}, we describe how to learn visual features from videos and re-utilize the existing image visual feature extractors.
In Section~\ref{sec:desc-gen}, an important aspect that we delve into is how sequential decoders learn to generate texts from encoded input videos' representations.
While Section~\ref{sec:matching-ranking} analyzes how state-of-the-art methods produce meaning representations in a joint space from videos and texts by two encoders.
These two sections also cover the most successful strategies for optimizing the models and present the most reported evaluation metrics for both techniques.
We cover the related competitions in Section~\ref{sec:competitions}, and we describe the standard datasets for benchmarking VTT methods in Section~\ref{sec:datasets}.
To show the state-of-the-art results in each dataset, in Section~\ref{sec:results}, we analyze and compare the reported results of the covered methods, presenting a new overall score to measure the relevance and establish a comparison between the description generation methods.
Finally, Section~\ref{sec:limitations} concludes the review, discussing the main research challenges in the VTT problem identified throughout the document.

\subsection{Video Captioning/Description}\label{sec:video-captioning}

Predicting a single sentence from an image (image captioning) has been a fundamental problem for several years~\citep{Chen2015MindsGeneration,Donahue2015Long-TermDescription,Gan2017StyleNet:Styles,Karpathy2015DeepDescriptions,Kiros2014UnifyingModels,Kuznetsova2014TREETALK:Descriptions,Mao2014Deepm-RNN,Rohrbach2013TranslatingDescriptions,Vinyals2015ShowGenerator}.
More recently, that problem was extended to generate descriptions from a video with only one event~\citep{Chen2020ASampling,Gan2017SemanticCaptioning,Gao2019HierarchicalCaptioning,Guadarrama2013Youtube2text:Recognition,Hemalatha2020Domain-SpecificCaptioning,Hou2019JointCaptioning,Kojima2002NaturalActions, Krishnamoorthy2013GeneratingKnowledge,Liu2018SibNet:Captioning,Pan2016JointlyLanguage,Pasunuru2017ReinforcedRewards,Rohrbach2013TranslatingDescriptions,Thomason2014IntegratingWild,Venugopalan2015TranslatingNetworks,Zhang2017Task-DrivenDescription}, or with multiple events \citep{Shen2017WeaklyCaptioning, Wang2018BidirectionalCaptioning, Zhou2018End-to-EndTransformer}. 
As a text generation task, video captioning is substantially more difficult than image captioning since spatial-temporal information in videos introduces diversity and complexity regarding the visual content and the structure of associated textual descriptions.
Although several proposals in the literature for video captioning can generate relevant sentence descriptions, they still have several limitations. 
Two of them are the existence of gaps in semantic representations and the often generation of syntactically incorrect sentences, which harms their performance on standard datasets.

Depending on the video duration, the problem is normally divided into \emph{short-video captioning} or \emph{long-video captioning}.
Basically, for short-video captions, we treat videos from five to 20 seconds with a single event.
Our goal is to generate a sentence to describe that event.
For longer videos, we usually deal with untrimmed videos.
These videos are typically five to ten minutes long with multiple events.
We need to automatically detect significant events in the video and generate multiple sentences to describe the different events.

\begin{figure}[t]
	\includegraphics[width=.8\textwidth]{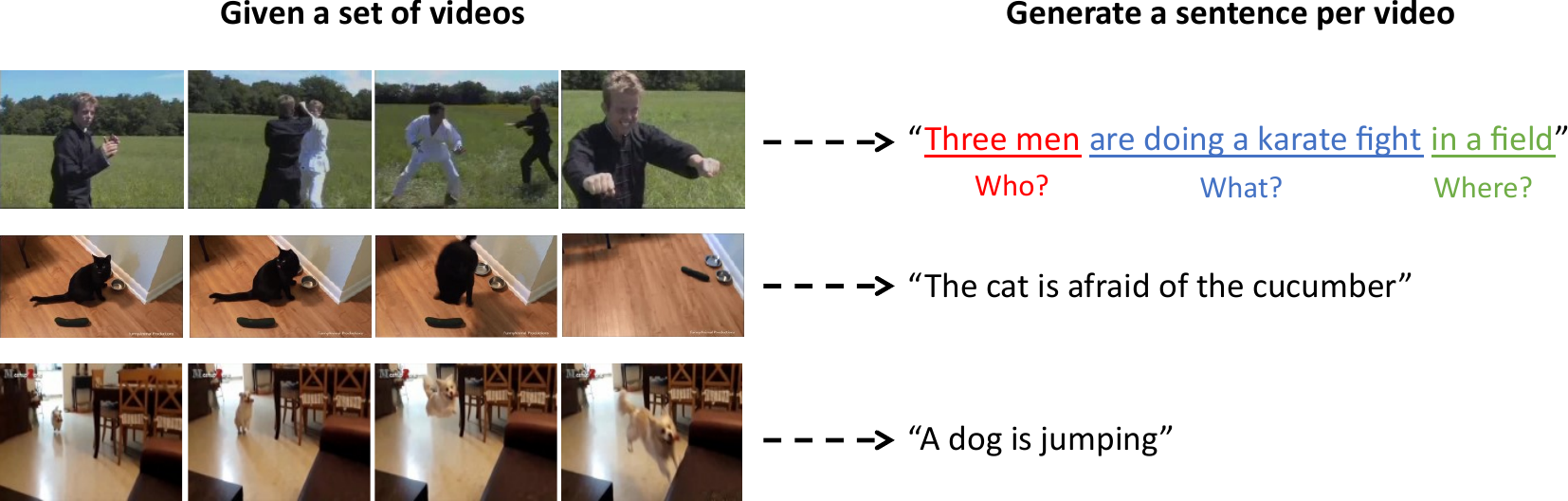}
	\centering
	\caption{Video description generation subtask: automatically generating a textual description for a video, answering questions like: Who is the video showing?, What are the objects and beings doing?, Where the situation happens?}
	\label{fig:dg-task}
\end{figure}

In general, we can divide the different video-to-text solutions into two categories (subtasks of VTT): one based on \emph{retrieval}, which selects the sentence from an available corpus through a video-to-text match (see Figure \ref{fig:mr-task}); and another based on \emph{generation} techniques, which basically generates a sentence using a captioning model (see Figure \ref{fig:dg-task}).
For several years, these two VTT subtasks have been evaluation tasks in the Annual TREC Video Retrieval Evaluation (TRECVID) Challenge, generating a great interest in the vision+language research community.
However, we must clarify that despite the common points between both tasks, they are entirely different.
Although text retrieval may seem like a viable approach for video captioning in scenarios where we have a number of manual annotations for a video and the task is to choose the best, the main goal of the video captioning task is merely to generate a new sentence from the input video.

\subsection{Text Retrieval from Video}\label{sec:video-matching}

\begin{figure}[t]
	\includegraphics[width=.8\textwidth]{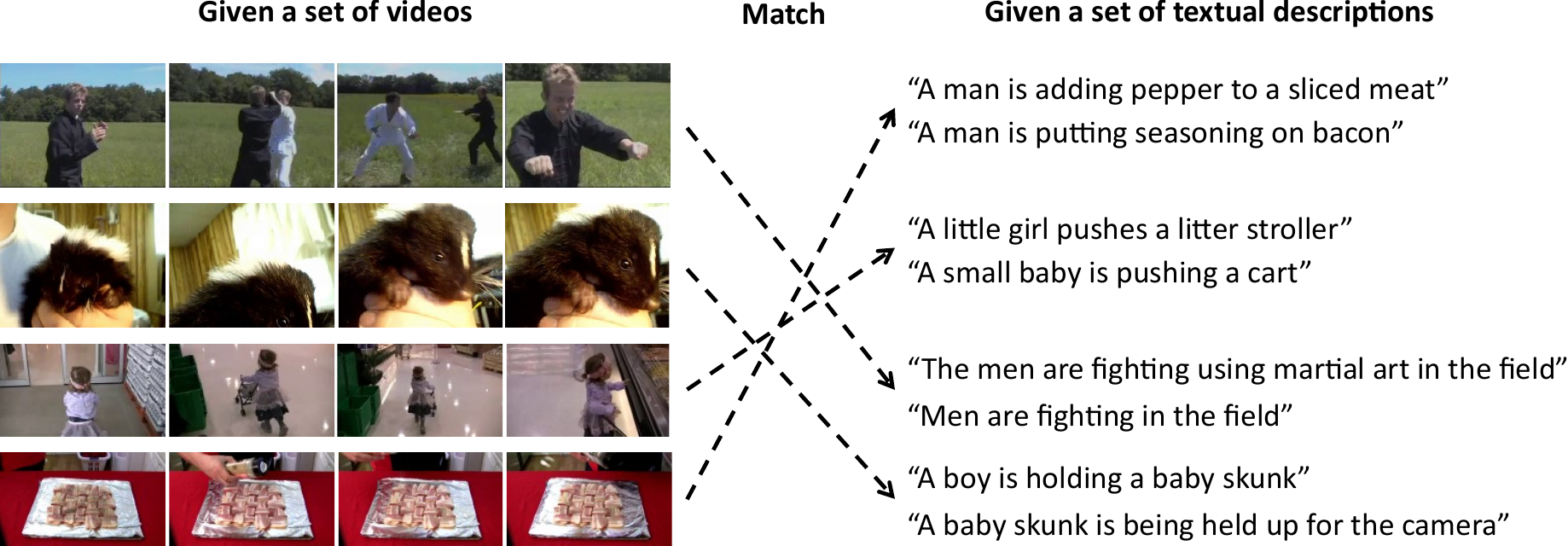}
	\centering
	\caption{Matching-and-ranking subtask: to determine the textual descriptions that best match with each video (to rank all texts for each video)}
	\label{fig:mr-task}
\end{figure}

The VTT subtask based on retrieval is also called the \emph{VTT matching and ranking subtask} and aims to rank a list of sentences for a given video based on their semantic relevance (see Figure~\ref{fig:mr-task}).
Training models for retrieving texts from videos could be considered a task without relevant real-world applications due to the current limitations for producing benchmarks with a number of textual descriptions, but this is not entirely true.
These models are usually trained in a \emph{cross-modal} strategy through learning a shared embedding space, that can indifferently embed both modalities.

This ability makes this task suitable as a pre-training technique for transferring knowledge about video descriptions to other downstream tasks~\citep{Goodfellow2016DeepLearning}, such as the \textit{cross-modal fine-grained action retrieval}~\citep{Wray2019Fine-GrainedEmbeddings}.
The actions could be treated as phrases and be represented with valuable contextual and grounded information learned from video descriptions. 
Due to this application and the rapid emergence of videos on the Web, the cross-modal retrieval between videos and texts has attracted growing attention. 

Specifically, a current dominant approach for matching models~\citep{Dong2019DualRetrieval, Ging2020COOT:Learning} is based on (but not limited to) training three components: a video encoder, a text encoder, and a joint embedding. 
However, simple joint embeddings are insufficient to represent complicated visual and textual details, such as scenes compositions, objects relations, and particular actions~\citep{Chen2020Fine-grainedReasoning}.
This review categorizes the state-of-the-art proposals according to the different directions adopted to combine these components and the \emph{levels of abstraction} proposed for encoding videos and texts.

For example, given a manner of representing frames (see Section~\ref{sec:visual-features}), to encode the visual information, three abstraction levels have been explored: (1) global, frequently based on a pooling operation over frame representations; (2) temporal, based on recurrent architectures over the frame representations; (3) local, which applies convolution operations over the sequence of temporal states for enhancing local patterns. 
These three levels have also been explored for encoding texts, replacing the frame representations with word representations produced by pre-trained word-embeddings.

Although these three different representation levels have been considered in both encodings, the fine-grained semantics and syntax have not been explicitly considered.
In our recent work~\citep{Perez-Martin2021ImprovingEmbedding}, we have applied cross-modal retrieval for learning representations with implicit syntactic information and generate more syntactically correct descriptions from videos.
Additionally, as we will detail in Section~\ref{sec:loss-mr}, the cross-modal distances in the embedding information can be used to improve the standard loss function employed for training the retrieval-based models: the \emph{ranking loss function}.

\subsection{Dense Video Captioning/Description}\label{sec:dense-video-captioning}

Dense video captioning is the task of predicting a semantic and syntactically correct sequence of words for each interesting event occurring in an input video.
We cannot introduce how to deal with this task without analyzing how the event detection methods work.
A typical event detection method usually follows a two-stage approach, including a candidate proposal generation stage and a proposal selection stage.
Many event candidates are proposed in the proposal stage by sliding window or neural networks such as single-stream temporal action proposals (SST) \citep{Buch2017SST:Proposals}.
Then, the event classifier is designated to predict event confidence for each candidate.
Proposals with confidence higher than a threshold will be selected as the final proposal of the event.

One main limitation of this approach is that the methods need to generate enough candidates, usually thousands of them, to ensure covering all correct events.
Moreover, the temporal relationships between the events are usually neglected, which results in the selection of events with high redundancy.
However, the most successful methods for dense video captioning usually work on a similar two-stage process: they first perform an events-proposal stage deciding a set of candidate intervals in the video that needs to be described, and then select the correct events and create the captions.


So, the videos where the challenging task dense video captioning/description is framed are much longer and complicated video sequences.
For evaluating these models, ActivityNet Competition includes the ``Dense-Captioning Events in Videos'' task\footnote{Dense-Captioning Events in Videos task of ActivityNet 2019 challenge website: \url{http://activity-net.org/challenges/2019/tasks/anet_captioning.html}} since 2017.
In this task, given a video as input, the participants must submit a temporally localized description for each relevant event in the video.
Then, for training models, we need videos with temporally localized descriptions.
In this sense, the ActivityNet Captions dataset, the YouCook2 dataset, and the HowTo100M dataset are the most reported dataset, covered in Section~\ref{sec:datasets}).

Deep dive into state-of-the-art for dense video captions is not in the scope of this review. 
However, we cover several works for this task that also report experimental results on video captioning or text retrieval tasks in the following sections.

\subsection{Other Vision-Text Tasks}\label{sec:other-tasks}

Bridging vision and natural language is a longstanding goal in computer vision and multimedia research, becoming a focus of research in linguistic and natural language processing (NLP) communities.
\emph{Vision-language intersection} represents a fundamental challenge for research areas like video analysis and understanding, human-computer interaction, and deep learning applications for vision and language.
Although video description generation and retrieval are the main topics of this review, it is sometimes more effective to train a simpler model to solve a complementary task and then move on to confront the final task~\citep{Goodfellow2016DeepLearning}.
This strategy that involves training simple models on different tasks before facing the difficulty of training the desired model to perform the desired task is collectively known as \emph{pre-training}.
Some complementary VTT tasks that can be considered for pre-training ---but are not part of the scope of this work--- are:
\begin{itemize}
    \item \emph{Visual question-answering}, where the input is a question posed about some visual content (image or video), and the output is the answer~\citep{Yu2015VisualAnswering, Srivastava2019VisualAnalysis, Manmadhan2020VisualReview}.
    \item \emph{Caption-based image/video retrieval}, given a caption and a pool of images, aims to retrieve the target image that is best described by the caption~\citep{Lu202012-in-1:Learning}.
    \item \emph{Grounding referring expressions}, where the inputs are a natural language expression and an image, and the output is the target region referred to by expression~\citep{Lu202012-in-1:Learning}.
    \item \emph{Video generation from text}, in which the goal is to generate a plausible and diverse video from an input text. 
    Here the broad picture and object motion must be determined by the text input~\citep{Li2017VideoText}.
    \item \emph{Multi-modal verification}, given one or more images and a natural language statement, aims to determine the correctness or predict their semantic relationship~\citep{Lu202012-in-1:Learning}.
\end{itemize}

\section{Visual Representation}\label{sec:visual-features}

Formally, while image captioning aims at converting from a \emph{fixed-length} sequence (image) to a \emph{variable-length} sequence of words, video captioning attempts to convert from a \emph{variable-length} sequence (video frames) to a \emph{variable-length} sequence of words. 
Then, we can formulate the video captioning problem as follows.
Let $x=(x_1,x_2,\ldots,x_n)$ be the sequence of video frames.
We want to construct a model, say $\mathcal{M}(\cdot)$, such that, with input $x$, the model outputs a sequence $y=(w_1,w_2,\ldots,w_m)$ of words such that $y$ correctly represents the information contained in $x$. 
In a machine learning context, we usually want to learn $\mathcal{M}(\cdot)$ from a set of examples $(x,y)$. 

Given this formulation, a critical step 
in the definition of $\mathcal{M}(\cdot)$ is the representation of the sequence of frames $x$.
The strategy to use in this step is so decisive that simply modifying it can be a major bottleneck to the entire system's performance.
These representations must be able to gather valuable information about videos in our dataset.

Visual media 
are unstructured perceptual data that inherently carry a very high-di\-men\-sion\-al representation.
This high dimensionality represents a challenge for machine learning systems trying to extract high-level semantic information directly from such visual contents. 
To address this challenge, researchers have traditionally represented images and videos with smaller feature vectors that attempt to encode the most relevant information present in them. 
This feature extraction step is crucial in any visual understanding pipeline. 
It 
serves as input for subsequent modules and can cause a drastic change in the model's performance.
This section reviews some feature extraction techniques for images and videos and identifies the best-performing ones, which will be used in designing our video understanding methods later on.

\subsection{How do we work with images?}
Traditionally, tasks such as object recognition have relied on using hand-crafted features to represent images. 
However, recently, deep Convolutional Neural Networks (CNN), which learn to extract features necessary for the task entirely from the data (with grid-like topology), have become a popular choice for image feature extraction, producing state-of-the-art performance in these traditional tasks.
The first example of that was the spectacular improvement in image classification accuracy seen on the ImageNet Large Scale Visual Recognition Challenge (ILSVRC) 2012, with the first use of CNNs in this competition. 
In this challenge, involving classifying the input images to one of thousand classes, the submission by Krizhevsky~\etal~\citep{Krizhevsky2012ImageNetNetworks} using a deep CNN outperformed all the others by a large margin. 
This set of further exploration into CNN architectures has driven up the ImageNet classification task's performance even to surpass the human classification accuracy~\citep{He2015DelvingClassification}.

More interestingly, the deep CNNs pre-trained on the large ImageNet dataset for the classification task generalize very well to other datasets and tasks~\citep{Yosinski2014HowNetworks}. 
That is, if we use the weights from CNNs pre-trained on ImageNet to initialize the networks before training them on other datasets and tasks, we can learn much better models than just using random initialization. 
Alternatively, using activations from some higher layer of an ImageNet pre-trained CNN as off-the-shelf image features has also been shown to produce state-of-the-art results~\citep{Donahue2014DeCAF:Recognition, Shetty2016Frame-Generation} on several datasets and tasks, such as object detection, instance segmentation, scene recognition, and image and video captioning. 
We will follow this idea, \ie use activations from CNNs pre-trained on ImageNet as feature input to our captioning model, without any fine-tuning of the CNNs for this task.
GoogLeNet~\citep{Szegedy2015GoingConvolutions}, VGG~\citep{Simonyan2015VeryRecognition}, and ResNet~\citep{He2016DeepRecognition} architectures, which won the different categories of later ILSVRC competitions, have been popular models for such feature extraction in the community with the immediate availability of code and pre-trained models.

\subsection{Video Features}\label{sec:video-features}

However, in the case of videos, how do we proceed?
To answer this question, we first need to understand what a video is.
Formally, a video is a 3D signal with 2D spatial coordinates and another temporal coordinate $t$. 
In the case of images, we have only the spatial dimension. 
In videos, we have temporal dimension and if we slice this cube at a specific value of $t$ at a temporal point, what we recover from that is an image that's a frame.
Understanding what a video is, we can explain what we can do to represent video and leverage the CNN architectures to process these sequences.

Before deep learning, the standard approach for representing videos involved two major stages.
In the first stage, \emph{local visual features} that describe a region of the video are extracted either \textit{densely}~\citep{Wang2011ActionTrajectories, Wang2013ActionTrajectories, Rohrbach2013TranslatingDescriptions} or at a \textit{sparse} set of interest points~\citep{Dollar2005BehaviorFeatures, Laptev2005OnPoints}.
Next, these raw features get combined into a \textbf{fixed-sized representation}.
One popular approach to doing this combination was to quantize all features using a learned k-means dictionary (codebook) and accumulate the visual words throughout the video into histograms~\citep{Laptev2008LearningMovies, Barbu2012VideoOut, Rohrbach2013TranslatingDescriptions}.
One of the first improvements to this standard approach was proposed by \cite{Rohrbach2013TranslatingDescriptions}.
They included both actions and objects on top of the \textit{dense trajectory features}, replacing the raw features with the higher-level representations of attribute classifier outputs.
While the raw video features tend to be too noisy to compute reliable distances, it has been shown that using the vector of attribute classifier outputs instead of the raw video features improves similarity estimates between videos~\citep{Regneri2013GroundingVideos}.
This difference led researchers soon after to start using the CNNs to obtain generic high-level features from videos instead of the noisy raw features.
So, we now describe how to use deep learning for extracting visual features from videos, re-utilize the existing image visual feature extractors, and incorporate these visual content representations into video captioning models.

\subsubsection{Single Frame Models}\label{sec:single-frame}
For this approach, we fit every frame to a CNN individually.
A small neural network transforms all outputs on the top of a pooling operation, \eg max, sum, or average, to obtain one feature vector for the whole video.
This is a straightforward model where we can reuse a CNN pre-trained for images and process all frames in parallel.
However, this approach's limitation is that these pooling operations are not aware of the temporal order.
This strategy is generally applied for short video clips with a single main event, 
performed from beginning to end.

\subsubsection{CNN + Recurrent Neural Networks (RNN)} 
Here, we try to leverage both worlds' best by including an RNN layer for combining the individual CNN representations instead of the pooling operation.
At the end of the input sequence, the model learns a representation, encoding the whole sequence's information.
We are aware of the signal's temporal evolution with this approach, but we have to do as many sequential steps as frames in the video.
This approach is part of the so-called \emph{dynamic encoding} strategy, and we analyze methods based on it in Section~\ref{sec:dg-dynamic-encoder}.

One aspect to consider when we try to process the sequences of frames of a video is the occurrence of cuts between the scenes.
The visual and semantic information in one video segment can be very different from that of the adjacent segments.
With RNN, we try to learn to combine this information. However, we can go further and incorporate explicit information about when these changes occur in the video.
This step is known as \emph{boundary detection} and has been incorporated into determining visual representations as a particular case of \textit{Hierarchical Recurrent Network Encoders} (HRNE).
Section~\ref{sec:desc-gen} covers works that include boundary detection as a crucial component of the description generation.

In this sense, determining the pixel-level correspondence ---mapping where each pixel goes in the next frame---, known as the \emph{optical flow} problem~\citep{Wang2019LearningTime}, contributes to obtain more discriminative visual representations.
Although it is not part of the scope of this work to detail the state-of-the-art techniques on optical flow estimation, much progress has been made in this field~\citep{Ilg2017FlowNetNetworks, Meister2018UnFlow:Loss, Varol2018Long-TermRecognition}, which can help the reader in its comprehension.
Specifically, \cite{Varol2018Long-TermRecognition} studied the impact of optical flow vector fields as a low-level representation. They demonstrated the importance of high-quality optical flow estimation for learning accurate action models.
Those authors found that using optical flows as inputs to 3D-CNNs results in higher performance than can be obtained from RGB inputs, but that the best performance could be achieved by combining RGB and optical flows.

\subsubsection{3D Convolutions (C3D)}\label{sec:c3d}
Following the deep CNN models' success on static images, why not extend these convolutions to the temporal coordinate as well?
We can add an extra dimension to standard CNN and assume that hierarchical representations of spatiotemporal data will be created.
The video needs to be split into chunks, with a fixed number of frames that fit the receptive field of C3D.
Usually, chunks of 16 frames are sufficient for representing the short temporal dynamics present in the video.
The  models need extensive training data and are usually pre-trained on some large-scale action recognition and video classification datasets, such as the Sports-1M~\citep{Karpathy2014Large-ScaleNetworks}, 20BN-something-something~\citep{Mahdisoltani2018OnLearning}, ActivityNet~\citep{Heilbron2015ActivityNet:Understanding}, and Charades~\citep{Sigurdsson2016HollywoodUnderstanding}.
The recent creation of all these datasets has partially removed the lack of enough labeled videos to address the video classification task and pre-train robust video features extractors.

\subsubsection{From 2D-CNN to 3D-CNN} 
It is not straightforward to see how to pre-train the C3D models.
We have good models on images, but if we need to train these huge 3D nets on thousands or millions of videos, it can take a considerable time.
So we want to reuse some models that we have already pre-trained.
A way to do this is taking the pre-trained 2D-CNN and simply ``inflate'' the filters, replicating the same filter on the temporal dimension.
Then we initialize the C3D with that instead of random weights and fine-tune for the specific task.
Following this idea, Carreira and Zisserman~\citep{Carreira2017QuoDataset} proposed the Inception 3D (I3D) architecture, which ``inflate'' all the 2D convolution filters used by the Inception V1 architecture~\citep{Szegedy2015GoingConvolutions} into 3D convolutions and carefully choose the temporal kernel size in the earlier layers.
They, similarly to more recent work such as R(2+1)D~\citep{Tran2018ARecognition}, ECO~\citep{Zolfaghari2018ECO:Understanding}, and Xie~\etal~\citep{Xie2018RethinkingClassification}, consider the large-scale Kinetics~\citep{Carreira2017QuoDataset} for training the model.
For a detailed examination of the architectures of various 2D-CNN with spatiotemporal 3D convolutional kernels on current video datasets, we refer the reader to~\citet{Hara2018CanImageNet}.
As a result of their analysis, they conclude that 3D-CNNs and Kinetics can contribute to significant progress in various video-related tasks such as action detection, video summarization, and optical flow estimation.

\subsubsection{Spatial Features}
Basically, no single video feature extraction method has achieved the best performance across tasks and datasets.
The recent progress achieved on challenging image-related tasks such as object detection and instance segmentation \citep{Abbas2019ASystems} has propitiated that recent video captioning methods also include spatial features based on identifying object regions on sampled frames~\citep{Pan2020Spatio-TemporalDistillation, Zhou2019GroundedDescription, Zhang2020ObjectCaptioning}.
Unlike frame-level 2D-CNN, the region features provide more fine-grained details of the video.
Usually, for each frame, methods use an object detector such as the ResNeXt-101 backbone-based Faster R-CNN~\citep{Ren2017FasterNetworks} pre-trained on Visual Genome~\citep{Krishna2017VisualAnnotations} or MSCOCO~\citep{Chen2015MicrosoftServer} datasets.

\medskip
\noindent The VTT tasks require us to identify objects and their attributes while also capturing the actions and movements that occur in the video.
Then, it is widespread to combine these features for representing the video content, encoding both the appearance and motion information.
Given the input video $x$, several state-of-the-art methods compress it into a global representation that we denote by $\rho(\cdot)$, which combines two standard visual features extractors.
Specifically, to avoid the intrinsic redundancy present in video frames, $p$ frames are sampled from $x$, and 2D-CNN feature vectors $(a_1, a_2, \ldots, a_p)$ and 3D-CNN feature vectors $(m_1, m_2, \ldots, m_p)$ are extracted, representing the appearance and motion information, respectively.
Then, these features are concatenated and averaged to produce $\rho(x)$, that is, $\rho(x)=\frac{1}{p}\sum_{i=1}^p [a_i, m_i]$.


\section{Methods for Generating Video Descriptions}\label{sec:desc-gen}

In the previous section, we talked about video captioning's difficulty compared with image captioning, but a video contains more information than an image. 
This comparison gives rise to the natural question of whether having a video implies greater or lesser difficulty in generating descriptions.
A video is a sequence of a lot of consecutive frames, sometimes with audio information. 
From this sequence, we can get much information like the motion and object transformations in time. 
We want to capture all this information for generating more relevant descriptions, but how to capture that motion information and introduce it in the generation process?
This section covers the essential methods proposed in the literature to answer these questions and solve the automatic \emph{VTT translation task}.
These methods can be grouped according to whether they are based on deep learning or not, the type of decoder they use (in the case of those based on the encoder-decoder framework), among other criteria.
The goal is to analyze distinctive aspects of each method, such as the proposed architecture, the datasets used to train the models, and the extracted visual features.



\subsection{Template-based Models}

One of the early approaches to successfully generate a video description is the template-based approach.
Here, the goal is to generate sentences within a reduced set of templates that assure grammatical correctness. 
These templates organize the results of a first stage of recognition of the relevant visual content. 
This approach is also called Subject-Verb-Object (SVO) triplets, and there are some works based on it~\citep{Kojima2002NaturalActions, Krishnamoorthy2013GeneratingKnowledge, Rohrbach2013TranslatingDescriptions,Guadarrama2013Youtube2text:Recognition, Thomason2014IntegratingWild, Xu2015JointlyFramework, Yu2015LearningInformation}.
For the sake of space, in this review, we cannot examine in detail these works, and we refer the reader to~\citet{Aafaq2019VideoMetrics} for more details.

This early work for the video captioning task, based on the two-stage SVO approach, provided several lessons that aided in the later work. 
We now analyze two of them.

An essential aspect of the first stage of visual recognition, is the explicit visual content identification.
It mainly includes the recognition and classification of actor, action, object, and scene in the video.
The success of deep learning in tasks such as \textit{object detection} and \textit{action recognition} allowed us to put the explicit visual recognition aside and reuse a CNN.
This CNN, pre-trained on a large number of images, can produce accurate \emph{global representations} and \emph{generic high-level features} from video frames, reducing noise (see Section~\ref{sec:video-features}).
However, more recent work reaffirms the importance of the explicit visual content identification and local features for video captioning.
\cite{Pan2020Spatio-TemporalDistillation} achieve state-of-the-art performance by aiming to \emph{ground} the words in the frames.
While \cite{Zhou2019GroundedDescription} and \cite{Zhang2020ObjectCaptioning} additionally aim to model the interactions between objects. 
We study these methods as \textit{Spatial Attention (SA)-based approaches} in Section~\ref{sec:dg-spatial-features}.

Another essential aspect of SVO approach is rigidly producing sentences with syntactic correctness by using a reduced set of templates.
As we explain in the next section, the most successful video captioning methods have a strong dependency on the effectiveness of semantic representations learned from visual models, but often produce syntactically incorrect sentences that harm their performance on standard datasets.
Because of this, recent deep-learning-based methods address this limitation by considering the learning of representations with syntactic information as an essential component of video captioning approaches~\citep{Hou2019JointCaptioning,Wang2019ControllableNetwork,Perez-Martin2021ImprovingEmbedding,Perez-Martin2020IMFD-IMPRESEEEmbedding}.
These methods, along with the SVO-based methods, have shown that videos, in addition to the appearance, motion, audio and semantic information, have implicit syntactic information that can be directly extracted from visual information.
We examine these works as \textit{Syntactic Guiding (SyG)-based approaches} in Section~\ref{sec:dg-syntactic-based}.

\subsection{Deep learning-based Models}

Although the template-based approach shows a pragmatic way to generate the description of videos, the sentences are very rigid and have limited vocabulary.
This limitation is a consequence of deciding a set of ``relevant'' objects, actions, and scenes to be recognized and the loss of human language's richness in the templates.
For any sufficiently rich domain, the required complexity of rules and templates makes the time-consuming manual design of templates unfeasible or too expensive.
Hence, the SVO approaches soon become inadequate in dealing with \emph{open-domain} datasets.

The high performance that deep learning models show for many areas of computer vision (\eg, action recognition task~\citep{Ji20133DRecognition, Donahue2015Long-TermDescription, Feichtenhofer2017SpatiotemporalRecognition, Kong2018HumanSurvey}) and multimedia information retrieval \citep{Sermanet2013OverFeat:Networks, Girshick2014RichSegmentation, Girshick2015FastR-CNN, Redmon2016YouDetection, Liu2016SSD:Detector, Dai2016R-FCN:Networks, Ren2017FasterNetworks} proves the effectiveness of neural networks for learning representations without requiring directly extracted features from the input data.
These networks operate as complex functions that propagate linear transformations of input values through nonlinear \textit{activation functions} (\eg the sigmoid or the hyperbolic tangent functions) to get outputs that can be further propagated 
to the upper layers of the network.
Consistent with this high-performance, deep learning-based video description methods have seen remarkable growth across all metrics in recent years.
This section analyzes the pros and cons of a vast number of ways to combine neural networks for video captioning, which are part of state-of-the-art methods.



\subsubsection{Encoder-Decoder Framework}
\begin{figure*}[t]
    \centering
    \includegraphics[width=.9\textwidth]{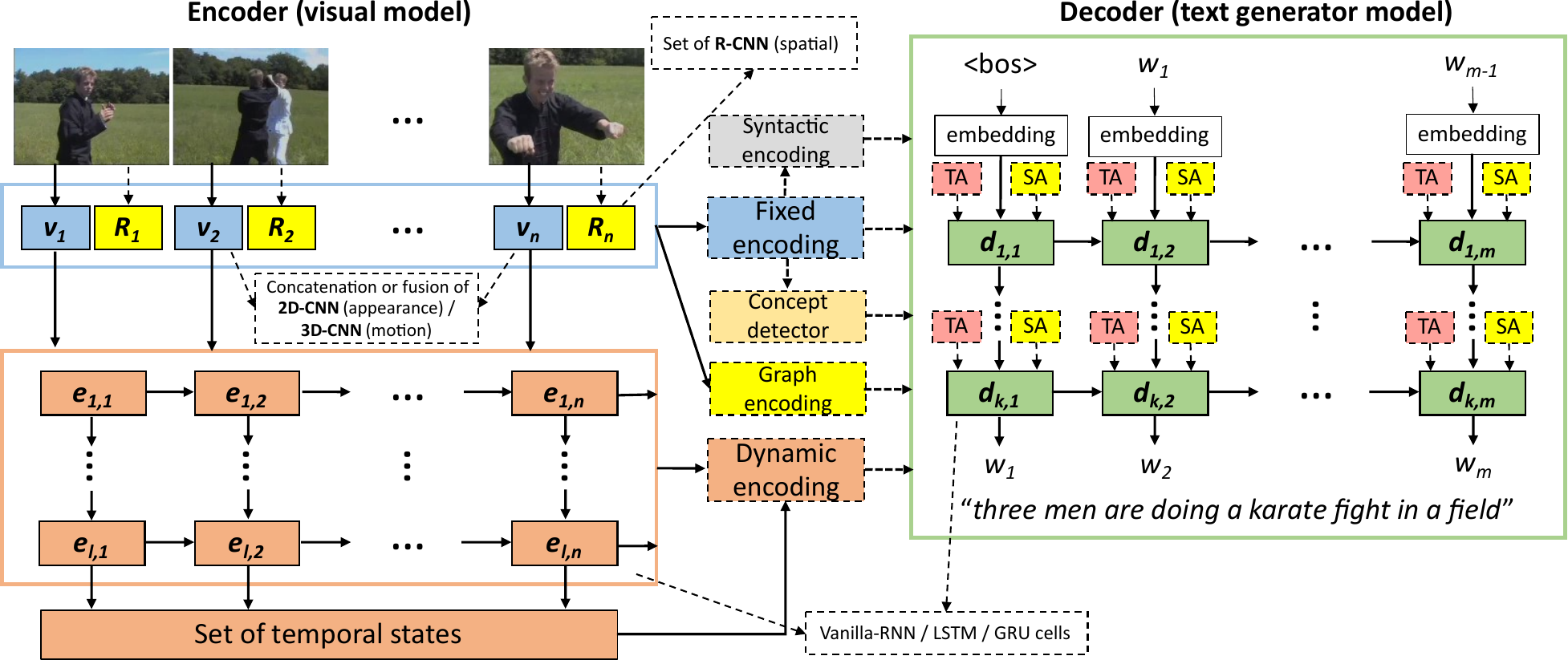}
    \caption{Video captioning/description encoder-decoder basic model.
    Dashed lines indicate components that may vary from one method to another.
    The visual model processes the video's frames and outputs a fixed representation, graph-based representation (with spatial information), or a dynamic representation (with temporal information).
    Visual features could be obtained from 2D-CNN features, 3D-CNN features, and R-CNN, or any combination of them.
    The fixed representation is computed by pooling the visual features.
    Graph encoding is usually obtained by a GNN over spatial features.
    Recurrent encoders offer dynamic representations.
    Some methods detect semantic concepts and/or obtain a syntactic encoding from fixed representation.
    Captions can be generated by including temporal or spatial attention (TA or SA) in each recurrent decoder step.}
    \label{fig:encoder-decoder}
\end{figure*}

As in neural machine translation~\citep{Eisenstein2019IntroductionProcessing}, one way of solving a video-caption problem is using neural network models based on the \emph{encoder-decoder} architecture~\citep{Cho2014LearningTranslation}.
The aim is to train a model that first constructs an \emph{encoded} representation from $x$, say $\operatorname{ENC}(x)$, and then, using $\operatorname{ENC}(x)$ as input, \emph{decodes} it to produce the sequence $y$ one word at a time.
Commonly, the \emph{encoder} and \emph{decoder} use classical building blocks such as CNNs and RNNs but with some crucial additions to mix visual, semantic, and syntactic features from the training data to boost its performance.
Some of these combinations are \emph{end-to-end trainable} deep network models where the two stages are learned simultaneously.
These models use backpropagation for going from pixels to text sequences instead of using bounding boxes, pose estimation, or any form of frame-by-frame analysis as an intermediate representation.
Figure \ref{fig:encoder-decoder} shows the main components of state-of-the-art encoder-decoder-based methods. 


\paragraph{Encoder:}
The encoder network $\operatorname{ENC}(\cdot)$ converts the source frame sequence $x$ into a real-valued representation. 
This representation holds the information, the features that represent the input. 
It might be a \emph{variable-size} or a \emph{fixed-size} set of feature vectors, a vector, or a matrix representation.
\begin{equation}\label{eq:encoder}
    z = \operatorname{ENC}(x)
\end{equation}

The selection of $\operatorname{ENC}(\cdot)$ depends on the type of input. 
For example, in machine translation, it is natural to use an RNN since the input is a \emph{variable-length} sequence of symbols. 
In contrast, in the case of selecting a fixed number of frames from all videos or using a simple pooling operation, the $\operatorname{ENC}(\cdot)$ will have a \emph{fixed-size}.



\paragraph{Decoder:}
The decoder network $\operatorname{DEC}(\cdot)$ generates the corresponding output $y$ from the encoded representation $z$ (encoding). 
As in the encoder, $\operatorname{DEC}(\cdot)$ must be chosen according to the type of output. 
In the case of video description generation, the output is a language description, and an RNN is a suitable architecture to use:
\begin{equation}\label{eq:decoder}
    y = \operatorname{DEC}(z)
\end{equation}

The decoder is typically an RNN that generates one word at a time according to an internal state, the previously generated word, and the entire input $z$ or part of it. 
So far, the proposed models use a wide range of variations of RNN in the decoder stage of the video description generation problem.
In Section~\ref{sec:loss-dg}, we review some strategies that could be used for training the entire architecture. 

Significant improvements in video captioning encoder-decoder models have been achieved by incorporating advanced techniques such as \emph{neural attention}, \emph{deep RNNs}, \emph{bidirectional RNN}, \emph{ensemble} translation models, \emph{beam search}, \emph{scheduled sampling}, \emph{professional learning}, \emph{reinforcement learning}, and the efficient \emph{self-attention} with the \emph{transformer} architecture.
The next sections review several state-of-the-art video captioning models that use these techniques and enhance the quality of generated captions by guiding the decoding process with learned representations such as semantic and syntactic representations.



\subsubsection{Fixed Encoder}
Unsurprisingly, the early deep learning-based models replaced and retained some strategies that were useful in template-based models.
Models proposed by \citet{Donahue2015Long-TermDescription} and \citet{Venugopalan2015TranslatingNetworks} incorporated two-layered Long Short-Term Memory (LSTM)-based decoders for text generation but maintained a fixed-size representation for encoding the video. 
On the one hand, \citet{Donahue2015Long-TermDescription} used the Conditional Random Field (CRF)-based approach~\citep{Rohrbach2013TranslatingDescriptions} to predict SVO triplets.
On the other hand, \citet{Venugopalan2015TranslatingNetworks} followed the single frame-based approach analyzed in Section~\ref{sec:single-frame}, using a previously trained CNN~\citep{Krizhevsky2012ImageNetNetworks}.

As we mentioned in Section~\ref{sec:visual-features}, recognizing the actions and movements that occur in the video is essential for video representation. 
\citet{Yao2015DescribingStructure} were the first to incorporate 3D-CNN features to encode the videos, but in contrast to other 3D-CNN formulations, the input to their 3D-CNN consists of features derived from three hand-designed image descriptors, \ie HoG~\citep{Dalal2005HistogramsDetection}, HoF, and MbH~\citep{Wang2009EvaluationRecognition}.
They also incorporated an attention model in the temporal dimension that we will analyze later.
The advantage of these 3D features is that they allow the encoders to accurately represent short-duration actions in a subset of consecutive frames. 

A vast number of models of the state-of-the-art have incorporated temporal representations into their encoder process.
For example, \citet{Gan2017SemanticCaptioning} proposed the Semantic Compositional Network (SCN) to understand individual semantic concepts from images effectively.
They also offered an extension for videos extracting 2D-CNN and 3D-CNN features to adequately represent the video's visual content.
The model generates two feature vectors from executing average pooling operations on all 2D-CNN features and all 3D-CNN features.
Until that moment, the video's \emph{spatiotemporal representation} combined appearance and motion, basically concatenating both vectors.

\subsubsection{Dynamic Encoder}\label{sec:dg-dynamic-encoder}
Like the template-based approach, the mentioned strategy of single-frame-based encoding suffers the loss of valuable temporal information due to the aggregation operations. 
To avoid this loss of information and obtain more meaningful representations, more sophisticated ways of encoding video features were proposed in later work, using, for example, an \emph{RNN-based encoder}~\citep{Venugopalan2015SequenceText, Venugopalan2016ImprovingText, Xu2015ANetwork, Srivastava2015UnsupervisedLSTMs, Yu2016VideoNetworks, Baraldi2017HierarchicalCaptioning}. 
For instance, \citet{Srivastava2015UnsupervisedLSTMs} have proposed an unsupervised learning model using a Long Short Term Memory (LSTM) layer in the encoder and another in the decoder.
In this unsupervised way, they achieve representations that can be used for other tasks such as reconstructing the input sequence, predicting the future sequence from the previous frames, and classification.


Recurrent encoders are intuitive to some extent, but better input encodings can be learned not only by processing the video stream from left to right by a single recurrent layer but by looking to the future or incorporating more layers. 
The Hierarchical Neural Encoder (HRNE)~\citep{Pan2016HierarchicalCaptioning} proposed by~\citet{Baraldi2017HierarchicalCaptioning} can learn to adapt its temporal connections according to the current input data. 
They built a time \textit{boundary-aware recurrent cell} on top of an LSTM unit that learns patterns with full temporal dependencies. 
When a boundary is detected, the LSTM's internal state is reinitialized, and a representation of the ended segment is given to the output.
The \textit{boundary-aware} layer's output is encoded through an additional recurrent layer and passed (as a vector) to the decoder.

To reduce the ambiguity of generated descriptions, \citet{Zhang2017Task-DrivenDescription} 
learn to construct the video's \emph{spatiotemporal representation}
by incorporating an \emph{adaptive fusion} component, which dynamically selects one of the three following combination strategies: (1) concatenation, for appea\-rance-centric entities; (2) sum or max, motion-centric entities; and (3) dynamic, for correlation-centric entities.
Although this work did not improve much over previous work, they were the first to replace the static patterns for fusing the features from different channels in the encoder~\citep{Chen2017VideoTopics, Wang2019ControllableNetwork}.
More sophisticated adaptive fusions have also been included in the decoder to select the most accurate information for generating each word~\citep{Hu2019HierarchicalCaptioning, Perez-Martin2020AttentiveCaptioning, Perez-Martin2021ImprovingEmbedding}.
We analyze them in the next section.

Methods based on recurrent encoders improved the state-of-the-art on video description generation. 
However, comparing their results with the results of techniques based on fixed encoders, the improvement is not very great. 
In this sense, authors like~\citet{Yu2016VideoNetworks} observed that the encoder has poor performance for tasks like the detection of small objects, implying incorrect object names in the sentences.
Consequently, the recurrent encoders are not the best option for videos of fine-grained activities, interaction with small objects, and datasets with very detailed descriptions, like the Charades Captions dataset~\citep{Sigurdsson2016HollywoodUnderstanding, Wang2018VideoLearning}.
The sentences in these datasets usually describe fine-grained actions that happen within a short duration and imply ambiguous information between them.

\begin{table*}[!t]
\tiny
\centering
\begin{tabular}{l|cccccc|ccccc|c}
\toprule
Method                                      & \multicolumn{6}{c|}{Encoder}                                    & \multicolumn{5}{c|}{Decoder}                       &         \\
                                            & 2D-CNN   & 3D-CNN   & R-CNN    & Fixed    & Dynamic  & Graph    & Arch.  & TA       & SA       & SeG      & SyG      &   RL    \\
\midrule 
\citet{Donahue2015Long-TermDescription}     &\checkmark&          &          &\checkmark&          &          &LSTM    &          &          &          &          &          \\
\citet{Venugopalan2015TranslatingNetworks}  &\checkmark&          &          &\checkmark&          &          &LSTM    &          &          &          &          &          \\ 
\citet{Rohrbach2015TheDescription}          &\checkmark&          &          &\checkmark&          &          &LSTM    &          &          &          &          &          \\
\citet{Xu2015JointlyFramework}              &\checkmark&          &          &\checkmark&          &          &LSTM    &          &          &          &          &          \\
\citet{Yao2015DescribingStructure}          &\checkmark&\checkmark&          &\checkmark&          &          &LSTM    &\checkmark&          &          &          &          \\
\citet{Yu2016VideoNetworks}                 &\checkmark&\checkmark&          &\checkmark&          &          &GRU     &\checkmark&          &          &          &          \\
\citet{Venugopalan2016ImprovingText}        &\checkmark&          &          &          &\checkmark&          &LSTM    &          &          &          &          &          \\
\citet{Pan2017VideoAttributes}              &\checkmark&\checkmark&          &          &\checkmark&          &LSTM    &          &          &          &          &          \\
\citet{Baraldi2017HierarchicalCaptioning}   &\checkmark&\checkmark&          &          &\checkmark&          &GRU     &          &          &          &          &          \\
\citet{Zhang2017Task-DrivenDescription}     &\checkmark&\checkmark&          &\checkmark&          &          &LSTM    &\checkmark&          &          &          &          \\ 
\citet{Gan2017SemanticCaptioning}           &\checkmark&\checkmark&          &\checkmark&          &          &LSTM    &          &          &\checkmark&          &          \\
\citet{Nina2018MTLE:Description}            &\checkmark&          &          &          &\checkmark&          &LSTM    &\checkmark&          &          &          &          \\ 
\citet{Yuan2018VideoGuiding}                &\checkmark&\checkmark&          &\checkmark&          &          &LSTM    &          &          &\checkmark&          &          \\
\citet{Wang2018VideoLearning}               &\checkmark&          &          &          &\checkmark&          &LSTM    &\checkmark&          &          &          &\checkmark\\
\citet{Chen2018LessCaptioning}              &\checkmark&          &          &          &\checkmark&          &GRU     &          &          &          &          &\checkmark\\
\citet{Gao2019HierarchicalCaptioning}       &\checkmark&\checkmark&          &\checkmark&          &          &LSTM    &\checkmark&          &          &          &          \\
\citet{Hou2019JointCaptioning}              &\checkmark&\checkmark&          &          &\checkmark&          &LSTM    &          &          &          &\checkmark&          \\
\citet{Wang2019ControllableNetwork}         &\checkmark&\checkmark&          &          &\checkmark&          &LSTM    &          &          &          &\checkmark&\checkmark\\
\citet{Zhou2019GroundedDescription}         &\checkmark&\checkmark&\checkmark&          &          &\checkmark&LSTM    &          &\checkmark&          &          &          \\
\citet{Chen2020ASampling}                   &\checkmark&\checkmark&          &\checkmark&          &          &LSTM    &          &          &\checkmark&          &          \\
\citet{Chen2020DelvingCaptioning}           &\checkmark&\checkmark&          &\checkmark&          &          &GRU     &          &          &\checkmark&          &          \\
\citet{Zhang2020ObjectCaptioning}           &\checkmark&\checkmark&\checkmark&          &          &\checkmark&LSTM    &\checkmark&\checkmark&          &          &          \\
\citet{Pan2020Spatio-TemporalDistillation}  &\checkmark&\checkmark&\checkmark&          &          &\checkmark&Transf. &          &          &          &          &          \\
\citet{Perez-Martin2020AttentiveCaptioning} &\checkmark&\checkmark&          &\checkmark&          &          &LSTM    &\checkmark&          &\checkmark&          &          \\
\citet{Perez-Martin2021ImprovingEmbedding}  &\checkmark&\checkmark&          &\checkmark&          &          &LSTM    &\checkmark&          &\checkmark&\checkmark&          \\
\bottomrule
\end{tabular}
\caption{Summary of most important encoder-decoder methods (sorted by year) presented in the literature to solve the automatic video captioning/description (translation from video to text) task.
For each method we check the main characteristics of its encoder and decoder. 
In the case of the encoder, we check the type of visual features used to represent the videos (2D-, 3D-, and/or R-CNN) and the strategy used to encapsulate the features (fixed-, dynamic-, or graph-based). 
In the case of the decoder, we show the architecture (Arch.) used, and we check if the model uses Temporal Attention (TA), Spatial Attention (SA), Semantic Guiding (SeG) or Syntactic Guiding (SyG). 
Finally, we check if the models are trained through Reinforcement Learning (RL) strategy and if the model are an ensemble of several models.}
\label{tab:dg-summary}
\end{table*}

\subsubsection{RNN-based Decoder}\label{sec:rnn-decoder}
Feed-forward and CNNs fail to adequately represent sequential structures because they can not remember previous information.
For example, they forget a frame in a video sequence just as they analyze the next one.
In contrast, the formulation of RNNs~\citep{Rumelhart1986LearningErrors} allows them to remember information for a long time as their natural behavior.

The RNN-based decoder takes the encoder output as input and starts generating text, one token at a time, conditioned on the input representation and the previously generated tokens.
In each step $t$, from the recurrent cell output, it produces a probability distribution $\hat{y}_t$ over a target vocabulary $V$ of possible output tokens, and choose one. 
The model maps this word to a vector representation using a pre-trained \textit{word embedding}, which is used as input of the recurrent cell at the next step:
\begin{equation}\label{eq:prob}
    \hat{y}_t = \text{softmax}(W_p\cdot h_t + b_p),
\end{equation}
\noindent where $h_t$ is the output of the recurrent cell at time step $t$, and $W_p$ and $b_p$ are parameters to be learned.

The decoder essentially works as a classifier, but it is aware of the whole input and the previously predicted sequence at each step.
Then, we can define the probability of generating the output sequence $Y=y_1,\ldots,y_m$ from the input sequence $X=x_1,\ldots,x_n$ as:
\begin{equation}
    p(Y|X;\Theta) = \prod_{i=1}^m p(y_i|y_1,\ldots,y_{i-1},X;\Theta),
\end{equation}
\noindent where $\Theta$ are the model parameters.

This perspective on text generation problems is very different from the template-based or pre-neural approaches. 
Here, text generation is consistent with the input and fluent in the target distribution, but it brought some exciting problems that did not exist before.
The vanilla RNN can model temporal dependency for a small-time gap, but it usually fails to capture long-term temporal information.
As for modeling natural languages, Long Short-Term Memory (LSTMs)~\citep{Hochreiter1997LongMemory} and Gated Recurrent Units (GRUs)~\citep{Cho2014LearningTranslation} have been very successful in video captioning decoders.
Two variants of these networks explored in video captioning are the Semantic-LSTM \citep{Pan2017VideoAttributes, Gan2017SemanticCaptioning, Yuan2018VideoGuiding, Chen2020ASampling} and the two-layered LSTM~\citep{Venugopalan2015SequenceText, Zhou2019GroundedDescription}. 
\citet{Venugopalan2015TranslatingNetworks} proposed to use a two-layered LSTM \citep{Graves2013SpeechNetworks}, where the hidden state of the first LSTM layer is the input to the second for caption generation. 
With this modification, the model simultaneously learns one stage of latent ``meaning'' and a more deep stage of fluid grammatical structures. 
\citet{Yu2016VideoNetworks} proposed a hierarchical-RNN (h-RNN) model to exploit \emph{spatiotemporal attention mechanisms}. 
Their model can decode through a sentence and paragraph generator. 
Specifically, a GRU layer, which simplifies the LSTM architecture, first takes the video features as input and generates a short sentence. 
Then, another recurrent layer is responsible for paragraph generation, combining the sentence generator's sentence vectors.
The paragraph generator captures the dependencies between the sentences and generates a sequence of relevant and consecutive sentences.
Table~\ref{tab:dg-summary} shows the architecture used for elaborating the state-of-the-art methods' decoders.

We now look into more advanced techniques for neural text production from videos, and show how later work focused on integrating key ideas from template-based approaches into the deep learning paradigm.
First, we show how sequential decoders can be improved using attention, external knowledge, joint embeddings, semantic guiding, and syntactic guiding.
Then we analyze some reinforcement learning-based methods and how to ensemble some models video captioning models for improving the results.

\paragraph{Temporal Attention (TA)-based approaches:}
When we defined the encoder in Equation~\ref{eq:encoder}, we mentioned that the encoding $z$ could be a \emph{variable-size} or a \emph{fixed-size} set of feature vectors, but we did not exemplify the \emph{variable-size}.
The decoders based on \emph{visual attention}, instead of taking as input a fixed global representation of the video, takes into account an input representation (usually called ``context vector'') built from the current decoder hidden state and all the input visual features (or the corresponding encoder hidden states).
With this vector, the decoder can decide ``where to look'' during word prediction in each step.
Analyzing the types of encoders, we mentioned that \citet{Yao2015DescribingStructure} used an attention mechanism as part of their model.
They incorporated a temporal attention mechanism adapted from \emph{soft attention}~\citep{Bahdanau2015NeuralTranslate} that allows the decoder to weight each temporal feature vector.
Basically, the context vector is created by the sum of the encoder state weighted by their probability.
To understand how these probabilities are obtained, we refer the reader to~\citet{Bahdanau2015NeuralTranslate}.
Intuitively, this model simulates the human attention that sequentially focuses on the most significant parts of the information over the video sequence to make predictions.

After that, the temporal-attention mechanisms have been widely used for video captioning~\citep{Gao2019HierarchicalCaptioning, Nina2018MTLE:Description, Perez-Martin2021ImprovingEmbedding, Wang2018VideoLearning, Yu2016VideoNetworks, Zhang2017Task-DrivenDescription, Zhang2020ObjectCaptioning}.
\citet{Gao2019HierarchicalCaptioning} and \citet{Perez-Martin2020AttentiveCaptioning} proposed an attention model able to decide whether to depend on the visual information or language context model. 
They proposed a hierarchical LSTM with two layers and an \emph{adaptive attention} model that extends \emph{temporal visual attention}.
\citet{Perez-Martin2020AttentiveCaptioning} get better results by including compositional LSTM and conditioning the adaptive gate and semantic layer by a temporal-attention mechanism.
Although these attention mechanisms are one of the main drivers of recent progress in video description generation models, we have a limited ability to understand how well temporal attention works in models because most datasets contain short videos.

\paragraph{Spatial Attention (SA)-based approaches:}\label{sec:dg-spatial-features}
As we mentioned at the beginning of this section, captioning methods based on object detectors perform two stages, \ie determine object proposals and fill them into predefined sentence templates.
In contrast, the deep learning-based works we have analyzed in this section neglect the explicit identification of objects in the video and only work on the frame-level and chunk-level.
An interesting point of using the object-level information is that we can also model the interaction between them by building meaningful connections and incorporate this information in the generation process for better \textit{visual grounding} capability.
Recent video captioning methods~\citep{Pan2020Spatio-TemporalDistillation, Zhang2020ObjectCaptioning, Zhou2019GroundedDescription} have explored the graph-based representations of spatial and temporal relationships between objects.
Specifically, \citet{Zhou2019GroundedDescription} processed the ActivityNet Captions benchmark for linking the captions to the evidence in the video by annotating each noun phrase with the corresponding bounding box in one of the frames.
This dataset allows producing grounding-based video captioning models that learn to jointly generate words and refine the grounding of the objects generated in the description.
These region-based annotations allow adopting a \emph{spatial-attention} mechanism similar to the temporal attention mechanisms or a more sophisticated model like \textit{self-attention}~\citep{Ging2020COOT:Learning, Lei2020MART:Captioning, Sun2019VideoBERT:Learning, Vaswani2017AttentionNeed, Zhou2018End-to-EndTransformer, Zhou2019GroundedDescription}.

Although~\citet{Zhou2019GroundedDescription} learn to attend to object regions in the frames, they only learn to ground the nouns.
They cannot explicitly model their relations, \eg the relation ``mixing'' between the nouns ``person'' and ``food''.
To model these interactions~\citet{Pan2020Spatio-TemporalDistillation} and \citet{Zhang2020ObjectCaptioning} proposed Object Relational Graphs (ORGs), which can learn the interaction among different objects dynamically.
These topological graphs can be automatically constructed according to criteria such as objects' location in the frames (spatial) and the detection of objects in consecutive frames (temporal).
After that, the node features can be updated during \textit{relational reasoning} by graph convolutional networks (GCN)~\citep{Kipf2017Semi-SupervisedNetworks}.
These three methods~\citep{Pan2020Spatio-TemporalDistillation, Zhang2020ObjectCaptioning, Zhou2019GroundedDescription} utilize an object detector such as the ResNeXt-101 backbone-based Faster R-CNN pre-trained on MS-COCO~\citep{Chen2015MicrosoftServer} or Visual Genome (VG)~\citep{Krishna2017VisualAnnotations} to extract object features in each frame.

\paragraph{Joint Embedding:}\label{sec:dg-joint-embedding}
For tasks like video-text retrieval, \ie \emph{video retrieval from descriptions} and \emph{video description retrieval from videos}~\citep{Dong2018PredictingRetrieval, Dong2019DualRetrieval, Fang2015FromBack, Ging2020COOT:Learning, Miech2018LearningData, Mithun2019JointRetrieval}, the joint visual-semantic embeddings have a successful application.
These embeddings are constructed by combining two models: a \emph{language model} that maps the captions to a language representation vector, and a \emph{visual model} that obtains a  visual representation vector from visual features.
Both models are trained for projecting those representations into a joint space, minimizing a distance function.
\citet{Dong2019DualRetrieval} 
obtain high-performance in retrieval tasks by using the same multi-level architecture for both models (based on RNNs) and training 
with the \emph{triplet-ranking-loss} function~\citep{Faghri2018VSE++:Negatives}.
\citet{Ging2020COOT:Learning} also use the same architecture (based on Transformers~\citep{Vaswani2017AttentionNeed}) for both models but produce the final embedding based on interactions between local context (clip/sentence features) and global context (video/paragraph features).
With this, they can learn the embedding from videos and paragraphs and explore different \emph{granularity} levels in the process.
In other words, they learn two embedding representations: the local context, which is learned by projecting the sentences (paragraph parts) and clips (video parts); and the global context, which is learned considering the full video and the complete paragraph and transforming the local embedding.
This hierarchical learning enforces the interactions within and between these hierarchical contextual embeddings. 
For training the model, they adopt a cross-modal strategy that we analyze in Section~\ref{sec:loss-mr}.

These embeddings have also been explored for video captioning to perform a particular form of transfer learning and domain adaptation~\citep{Gao2017VideoConsistency, Ging2020COOT:Learning, Liu2018SibNet:Captioning, Pan2016JointlyLanguage}.
These methods explore the representations captured by the joint embedding's visual models for improving the generalization in the captioning setting.
In LSTM-E~\citep{Pan2016JointlyLanguage}, a joint embedding component is utilized to bridge the gap between visual content and sentence semantics.
This embedding is trained by minimizing the \textit{relevance loss} and \textit{coherence loss} simultaneously.
This minimization guarantees, coherently and smoothly, the \textit{perplexity} of the generated sentence: the contextual relation between the sentences' words.
SibNet~\citep{Liu2018SibNet:Captioning} exploits autoencoder for visual information and a visual-semantic embedding for semantic information.
A disadvantage of these joint embeddings is that they only consider the implicit contextual relations between the words in the sentence using pre-trained word embeddings.
To improve the syntax correctness of generated sentences, \citet{Perez-Martin2021ImprovingEmbedding} learn a new representation of videos with suitable syntactic information. 
They propose a joint \emph{visual-syntactic embedding}, trained for retrieving POS tagging\footnote{\url{https://www.nltk.org/book/ch05.html}} sequences from videos.
Section~\ref{sec:dg-syntactic-based}, explains how they incorporated the \emph{visual model} in the decoder to alleviate the video content's syntactic inconsistency. 



\paragraph{Semantic Guiding (SeG)-based approaches:}\label{sec:dg-semantic-based}
Another way of exploiting informative semantics is by learning to ensemble the result of visual perception models~\citep{Chen2020ASampling, Gan2017SemanticCaptioning, Long2018VideoAttention, Pan2017VideoAttributes, Perez-Martin2020AttentiveCaptioning, Perez-Martin2021ImprovingEmbedding, Yuan2018VideoGuiding, Xu2019Semantic-filteredFeature}.
\citet{Pan2017VideoAttributes} incorporated the transferred semantic attributes learned from two sources (images and videos) inside a CNN-RNN framework. 
For this, they integrated a \emph{transfer unit} to dynamically control the impact of each source's semantic attributes as an additional input to LSTM.
\citet{Gan2017SemanticCaptioning} included the semantic meaning via a \emph{semantic-concept-detector} model, which predicts each concept's probability that appears in the video by a multi-label classification approach.
They incorporated concept-dependent weight tensors in LSTM for composing the semantic representations.

In contrast, \citet{Yuan2018VideoGuiding} proposed the Semantic Guiding Long Short-Term Memory (SG-LSTM), a framework that jointly explores visual and semantic features using two semantic guiding layers. 
These layers process three kinds of semantic features: global-, object-, and verb-semantic. 
They predict these features by selecting the 300 most-frequent subjects, verbs, and objects from the MSVD corpus and training three \text{MLP} (one for each kind of semantic feature) as standard multi-label classifiers. 
Explicitly, they compute the representations $g_a = \text{MLP}([f,m])$, $o_a = \text{MLP}(f)$ and $v_a = \text{MLP}(m)$, where $f$ is the average 2048-dimensional $pool_5$ layer from ResNet-152 pre-trained on ImageNet dataset, $m$ is the 512-dimensional $pool_5$ layer from C3D pre-trained on Sport-1M dataset and $[f,m]$ is their concatenation.

More recently, \citet{Chen2020ASampling} modified the model SCN-LSTM~\citep{Gan2017SemanticCaptioning}. 
They used $\tanh(\cdot)$ to activate the raw cell input instead of $\sigma(\cdot)$. 
Their model includes a semantic-related video feature term at each recurrent step. 
Also, these authors assumed that captions should be both accurate and concise.
They proposed a sentence-length-related loss function that keeps a balance between those criteria, incorporating a penalty weight (related to the generated sentence length) to the sum of the logarithm probability of the CELoss \citep{Goodfellow2016DeepLearning} function. 
A deficiency of this work is that the model fails when tested using a beam search strategy.
Maybe this is a result of the absence of the penalty weight in the testing phase.

The same authors, \citet{Chen2020DelvingCaptioning}, improved this method, proposing the VNS-GRU approach.
They maintained the semantic information but improved the form of computing the penalty weight in the optimization process. 
For that, they proposed \textit{professional learning} that aims, in the training algorithm, to assign higher weights to the samples with lower loss. 
An advantage of this is that the model can learn unique words and complicated grammar structures.


All these methods show the benefits of describing the videos according to dynamic visual and semantic information. 
However, these models' performance strongly depends on the quality of semantic concept detection models, 
which makes it difficult to generate words that are not included in the set of concepts to be classified.
This strong dependence can be alleviated by including an adaptive mechanism~\citep{Hu2019HierarchicalCaptioning, Perez-Martin2020AttentiveCaptioning} that selectively determines the visual and semantic information required to generate each word.

\paragraph{Syntactic Guiding (SyG)-based approaches:}\label{sec:dg-syntactic-based}
Although some recent works in image captioning~\citep{Deshpande2019FastPart-Of-Speech,He2019ImageGuidance} and video captioning~\citep{Hou2019JointCaptioning,Wang2019ControllableNetwork} have explored the use of syntactic information in the generation process, its impact has not been widely explored.
For the video captioning task, \citet{Hou2019JointCaptioning} 
created a model that first generates a sequence of POS tags and a sequence of words, and then learns the joint probability of both sequences using a probabilistic directed acyclic graph.
\citet{Wang2019ControllableNetwork} integrated a syntactic representation in the decoder, also learned by a POS sequence generator.
These models' weakness is that they neither directly exploit the relationship between syntactic and visual representations nor between syntactic and semantic representations.

In contrast, as we mentioned in Section~\ref{sec:dg-joint-embedding}, \citet{Perez-Martin2021ImprovingEmbedding} proposed a method to learn a visual-syntactic embedding and obtain syntactic representations of videos instead of learning to generate a sequence of POS tags.
They also include two compositional layers to obtain visual-syntactic-related and semantic-syntactic-related representations.
In detail, their decoder is based on three recurrent layers based on the composi\-tional-LSTM network~\citep{Chen2020ASampling, Gan2017SemanticCaptioning}, and an additional layer to combine the outputs of the three recurrent layers defined by two levels of what they called \emph{fusion gates}.
Unlike other architectures~\citep{Baraldi2017HierarchicalCaptioning,Gao2019HierarchicalCaptioning,Pan2016HierarchicalCaptioning}, how their recurrent layers are combined follows a similar scheme to the transformer's multi-head; they are not deeply connected.
So, the computation of the three layers can be done in parallel.

\subsubsection{Ensembles}
Some video captioning works have achieved outstanding performance by assembling several models~\citep{Chen2018RUC+CMU:Videos, Nina2018MTLE:Description, Song2019RUC_AIM3Text}.
This strategy is usually adopted in benchmarks competitions where participant teams can submit several runs of their model, \eg TRECVID (see Section~\ref{sec:challenge:trecvid}).
Usually, one of these runs is the ensemble of models used in the other runs.
In this sense, \citet{Chen2018RUC+CMU:Videos} proposed a model for dense video captioning task, which is composed of four components: (1) segment feature extraction (ResNet, i3D, VGGish, and LSTM); (2) proposal generation, for selection of fragments of the video; (3) caption generation; and (4) re-ranking. 
Specifically, for the third component, they ensemble three different caption generation models: (1) a vanilla caption model, based on an LSTM decoder; (2) a temporal attention caption model; and (3) a topic guided caption model, based on a semantic concept detector.
The combination of outputs of different decoders has also been adopted for other methods~\citep{Hou2019JointCaptioning, Nina2018MTLE:Description}.
The model proposed by \citet{Nina2018MTLE:Description} relies on distinct decoders that train a visual encoder following a multitask learning paradigm~\citep{Caruana1998MultitaskLearning}.
While~\citet{Hou2019JointCaptioning} (analyzed in Section~\ref{sec:dg-syntactic-based}) incorporated a decoder for POS tags generation and computed the joint probability with the natural language decoder.


\citet{Song2019RUC_AIM3Text} achieved the best results on the TRECVID VTT 2019 challenge (see Section \ref{sec:challenge:trecvid}). 
In this case, they proposed a model that consisted of four main parts: (1) video semantic encoding; (2) description generation with temporal and semantic attention; (3) reinforcement learning optimization; and (4) ensemble from multi-aspects.
The ensemble proposed for the fourth part is based on choosing the video's best-generated description by re-ranking the captions using the weighted sum of \textit{fluency} and \textit{relevancy} scores.
The third part, based on reinforcement learning, has also been adopted by several state-of-the-art methods.
We now briefly analyze some of these methods, and general aspects of this strategy are exposed in Section~\ref{sec:loss-dg}.

\subsubsection{Reinforcement Learning (RL)-based Methods}
Throughout this section, we have exposed the different strategies that avant-garde works have adopted for the video captioning task.
Some of these works also incorporate \emph{reinforcement learning} as part of their models' training process\citep{Chen2018LessCaptioning, Li2019End-to-EndLearning, Pasunuru2017ReinforcedRewards, Phan2017Consensus-basedCaptioning, Wang2018VideoLearning, Wang2019ControllableNetwork, Wei2020ExploitingCaptioning}.
In Section~\ref{sec:loss-dg}, we study how, unlike cross-entropy-based methods, these models are trained to optimize one evaluation metric (reward) directly.
An advantage of these models is that they are not limited to a specific evaluation metric.
We can extend them using other reasonable rewards.

An essential issue for video description generation is the selection of the most informative frames. 
In this sense, \citet{Chen2018LessCaptioning} have introduced the efficient PickNet method. 
PickNet is an RL-based model that chooses the most informative frames to represent the whole video.
It picks for encoding around 33.3\% of sampled frames only (6 frames for MSVD and 8 for MSR-VTT, on average), largely reducing the computation cost. 
The goal of this method is to reduce noise and increase efficiency without losing efficacy. 
The model calculates each frame selection's regard in function to minimize the visual diversity and textual discrepancy.
The authors decided only to use the appearance features because extracting motion features was very time-consuming for the model, deviating from cutting down the computation cost for video captioning. 
An advantage of this model is that it can be easily incorporated into any other model to increase efficiency.

\subsection{Loss Functions for Optimization}\label{sec:loss-dg}
In previous sections, we studied the state-of-the-art techniques for creating effective methods that produce a sequence of words from an input video.
However, we have not detailed how these models are trained.
This section analyzes the different objective functions for video captioning methods.
In particular, we include an analysis of sequence-level- and reinforcement-learning-based strategies.

After estimating each word's probability distribution in the vocabulary by Equation~\ref{eq:prob},
most video-captioning models are trained by the 
cross-entropy loss minimization (CELoss).
With $\text{CELoss}$, the model is trained to 
be good at greedily predicting the next correct word of the reference caption.
Given a target caption, $\hat{y}~=~(\hat{w}_1, \hat{w}_2, \dots, \hat{w}_L)$ describing an input video, one minimizes:
\begin{equation}
    \mathcal{L}_\Theta = -\sum_{t=1}^L \log \text{p}_\Theta(\hat{w}_t|\hat{w}_{<t}),
\end{equation}



\noindent where $\Theta$ are all the learnable parameters of the captioning model, $\hat{w}_t$ is the $t$-th word in the ground-truth caption, and $\hat{w}_{<t}$ are the first $t-1$ words in the ground-truth caption. 
$\text{p}_\Theta(\hat{w}_t|\hat{w}_{<t})$ is modeled by a logistic function as the \emph{decoder} network's output layer, understood as the probability distribution of words in the vocabulary.

However, at test time, the model has only access to its predictions, which may not be correct.
Additionally, video captioning's popular evaluation metrics are based on $n$-gram overlapping between the generation and reference captions, \eg BLEU~\citep{Papineni2002BLEU:Translation}.
This discrepancy leads to a problem called \textit{exposure bias}~\citep{Ranzato2016SequenceNetworks} in sequential decoders.
Due to this discrepancy, the word-level $\text{CELoss}$ function is not directly related to the most used evaluation metrics for video captioning~\citep{Ranzato2016SequenceNetworks, Pasunuru2017ReinforcedRewards}. 
Some state-of-the-art works show that conventional $\text{CELoss}$ cannot extract the effectiveness of their models.




To overcome this limitation while operating with explicit supervision at the sequence level, \citet{Chen2020ASampling} proposed a loss function that weights the $\text{CELoss}$ according to the length of the reference captions. 
\begin{equation}
    \mathcal{L}_\Theta' = -\frac{1}{L^\beta}\sum_{t=1}^{L}\log \text{p}_\Theta(\hat{w}_t|\hat{w}_{<t}),
\end{equation}

\noindent $L$ is the length of the ground-truth caption, and $\beta \in [0,1]$ is a hyperparameter of the model. 
The hyperparameter $\beta \in [0,1]$ regulates the length of the generated sentences.
The higher the value of $\beta$, the lower the loss produced by the function, and vice versa.
Thus, a value of $\beta$ near 0 implies that the model rapidly adapts to generate concise sentences that could affect their syntactic correctness.

Another strategy to overcome this limitation is to expose the model to its own predictions rather than just the training data, and directly optimize the task-specific evaluation metric.
The models trained following this strategy commonly use policy gradients and mixed-loss methods for \emph{reinforcement learning}~\citep{Pasunuru2017ReinforcedRewards}.
Instead of maximizing the likelihood of prediction at each step, the model first generates the whole sequence.
A reward (often one of the evaluation metrics at test time, \eg CIDEr) is estimated by comparing the output text to the ground-truth description.
The reward is then used to update the model parameters, learning to maximize the evaluation metric during training.

In particular, state-of-the-art methods based on reinforcement learning use a standard algorithm called \emph{policy learning}, which allows us to maximize a non-differentiable reward efficiently.
These methods view the description generation model as an \emph{agent} that interacts with an \emph{environment} with access to reference texts.
Each input video represents a \emph{state}, and each \emph{action} is an output text generated by the agent.
The agent predicts an output caption $\hat{y}$ using \emph{policy} $p(\hat{y}|x, \Theta)$, where $x$ is the input video and $\Theta$ are model parameters.
After each action, the agent receives an immediate \emph{reward} corresponding with how well the final caption resembles the reference text.
The transition between states is trivial because the inputs are interdependent.
The models learn to optimize the policy such that the agent can take an optimal action optimizing its reward given a state.

The reward function plays a crucial role here, which is defined such as that it captures the critical aspects of the target output concerning the benchmark dataset.
Several methods~\citep{Chen2018LessCaptioning, Li2019End-to-EndLearning, Pasunuru2017ReinforcedRewards, Phan2017Consensus-basedCaptioning, Wang2018VideoLearning, Wang2019ControllableNetwork} use the CIDEr score as the reward over the MSVD and MSR-VTT datasets.
In particular, \citet{Pasunuru2017ReinforcedRewards} proposed  CIDEnt-RL, which penalizes the phrase-matching metric (CIDEr) based reward when the entailment score is low.
While \citet{Phan2017Consensus-basedCaptioning} compared the performance using different metrics as a reward, \ie BLEU-4, METEOR, ROUGE$_L$, and CIDEr.
Although the improvement on different metrics is unbalanced because the improvements on other metrics are not as large as the specific metric directly optimized, they obtain the best performance for all metrics optimizing the CIDEr score.

\subsection{Description Generation-specific Evaluation Metrics}\label{sec:metrics-dg}

In the literature for NLG-based techniques to address the VTT problem, authors have reported some automatic assessments.
The majority of them have come from the metrics used for machine translation and image captioning tasks.
For the VTT translation task, the state-of-the-art works report several $n$-gram-overlap-based metrics to automatically evaluate the quality of the output text (candidate) with the dataset's paired descriptions (references), being the most-reported metrics: 
\begin{itemize}
    \item \emph{BLEU} (Bilingual Evaluation Understudy~\citep{Papineni2002BLEU:Translation}, 2002), based on $n$-gram precision;
    \item \emph{ROUGE} (Recall Oriented Understudy of Gisting Evaluation~\citep{Lin2004Rouge:Summaries}, 2004), based on $n$-gram recall;
    \item \emph{METEOR} (Metric for Evaluation of Translation with Explicit Ordering~\citep{Banerjee2005METEOR:Judgments}, 2005), based on $n$-gram with synonym matching; and
    \item \emph{CIDEr} (Consensus-based Image Description Evaluation~\citep{Vedantam2015CIDEr:Evaluation}, 2015), based on precision and recall by \textit{TF-IDF} weighting of $n$-gram similarity.
\end{itemize}

A range of other automated metrics has been proposed for machine translation.
The explicit semantic match metrics define the similarity between human-written references and model generated candidate by extracting semantic information units from text beyond $n$-grams.
These metrics operate on semantic and conceptual levels and correlate well with human judgments~\citep{Celikyilmaz2020EvaluationSurvey}.
The \emph{SPICE} (Semantic Proportional Image Captioning Evaluation~\citep{Anderson2016SPICE:Evaluation}, 2015) metric has also been reported for video captioning.
This metric uses scene graphs to parse the candidate sentence to semantic tokens, such as object classes, relationship types, or attribute types.
In this section, we present the definition of these metrics and briefly discuss their advantages and limitations.
For more details on the evaluation of text generation models, we refer the reader to \citet{Celikyilmaz2020EvaluationSurvey}, a recent review on NLG evaluation methods. 

Early video description generation works~\citep{Guadarrama2013Youtube2text:Recognition, Krishnamoorthy2013GeneratingKnowledge, Rohrbach2013TranslatingDescriptions} used the \emph{SVO Accuracy} to measure whether the generated SVO triplets coheres with ground truth. 
This metric does not evaluate the quality of the generated sentence. 
It aims to focus on the matching of general semantics and ignores visual and language details.


\textbf{Bilingual Evaluation Understudy computation (BLEU)}~\citep{Papineni2002BLEU:Translation} is based on the match between 
$n$-grams (a sequence of $n$ words) of the candidate sequence and one or some reference sequences.
\emph{BLEU} is a weighted geometric mean of $n$-gram \emph{precision} scores, defined as:
\begin{equation*}
p_n = \frac{\sum_s \text{min}\big(c(s, \hat{y}), c(s, y)\big)}{\sum_s c(s, \hat{y})},
\end{equation*}
\noindent where $\hat{y}$ is the candidate sequence, $y$ is the reference sequence, $s$ is a $n$-gram sequence of $\hat{y}$, and $c(s, \hat{y})$ is the number of times that $s$ appears in $\hat{y}$.
The $\text{BLEU-N}$ metric is then:
\begin{equation}\label{eq:bleu}
    \log \text{BLEU-N} = \min(1-\frac{r}{c},0) + \sum_{n=1}^N w_n \log p_n,
\end{equation}
\noindent where $c$ and $r$ are the lengths of the candidate and reference sequences, respectively. 
$N$ is the total number of $n$-gram precision scores to use, and $w_n$ 
is the weight for each precision score, which is often set to be $1/N$.

Equation \ref{eq:bleu} counts the position-independent matches among the candidate's $n$-grams and the reference's $n$-grams and computes the percent of matches (precision).
For video captioning, the most reported version is $\text{BLEU-4}$ with uniform weights $w_n = 1/N$. 
$\text{BLEU}$ was originally designed to evaluate long texts, and its use as an evaluation measure for individual sentences \textit{may not be fair}.

\textbf{Recall Oriented Understudy of Gisting Evaluation (ROUGE)}~\citep{Lin2004Rouge:Summaries} is a set of metrics designed for determining the quality of automatic summarization of long texts. 
Although mainly designed for evaluating summarization, it has also been widely used for evaluating short text production approaches, such as image captioning and video captioning.
It computes the $n$-gram \emph{recall} score of the candidate sentences with respect to those of reference~\citep{Lin2004Rouge:Summaries}. 
Similar to $\text{BLEU}$, it also is computed varying the number of $n$-grams. 
Let $c_{match}(s)$ be the number of times that the $n$-gram $s$ co-occur in the candidate sequence.
The $\text{ROUGE-N}$ score is then:
\begin{equation}
\text{ROUGE-N} = \frac{\sum_{y\in R}\sum_{s\in y}c_{match}(s)}{\sum_{y\in R}\sum_{s \in y}c(s)}
\end{equation}

\noindent where $s\in y$ is a $n$-gram sequence of the reference $y\in R$. 
The denominator $\sum_{y\in R}\sum_{s \in y}C(s)$ of the equation is the total sum of the number of $n$-grams occurring at the reference summary side.
Note that the number of $n$-grams in the denominator increases as more descriptions become available for reference.

In the same paper, the authors proposed several $\text{ROUGE-N}$ variants, such as  $\text{ROUGE-L}$, which is the most used variant for image and video captioning evaluation.
For $\text{ROUGE-L}$, the longest common subsequence (LCS)-based \emph{F-measure} is computed to estimate the similarity between the candidate sentence and the reference.
In other words, it measures the longest matching sequence of words using LCS. 
One advantage of using LCS is that it does not require consecutive matches, but it requires in-sequence matches that indicate sentence-level word order~\citep{Lin2004Rouge:Summaries}. 
The other advantage is that the $n$-gram length does not need to be predefined since ROUGE-L automatically includes the longest common $n$-grams shared by the reference and candidate text.

Formally, given two sequences $\hat{y}$ and $y$, the $\text{LCS}(\hat{y}, y)$ the length of the LCS of $\hat{y}$ and $y$.
Let $m$ the length of $\hat{y}$ and let $n$ the length of $y$.
The LCS-based \emph{F-measure} is then:
\begin{equation}
    \begin{split}
        R_{lcs} &= \frac{\text{LCS}(\hat{y}, y)}{m},\\
        P_{lcs} &= \frac{\text{LCS}(\hat{y}, y)}{n},\\
        F_{lcs} &= \frac{(1+\beta^2)R_{lcs}P_{lcs}}{R_{lcs}+\beta^2P_{lcs}},       
    \end{split}
\end{equation}
\noindent where $\beta=\frac{P_{lcs}}{R_{lcs}}$ when $\frac{\partial F_{lcs}}{\partial P_{lcs}}=\frac{\partial F_{lcs}}{\partial R_{lcs}}$.
Notice that $\text{ROUGE-L}$ is one when $\hat{y} = y$ since $\text{LCS}(\hat{y}, y) = m$ or $n$; while $\text{ROUGE-L}$ is zero when $\text{LCS}(\hat{y}, y) = 0$, \ie there is nothing in common between $\hat{y}$ and $y$.

Another metric that has been widely used for evaluating image and video captioning models is the \textbf{Metric for Evaluation of Translation with Explicit Ordering (METEOR)}~\citep{Banerjee2005METEOR:Judgments}.
Compared to BLEU (based on precision only), METEOR is based on the harmonic mean of the \textit{uni}-gram precision and recall, in which recall is weighted higher than precision~\citep{Celikyilmaz2020EvaluationSurvey}.
This metric couples BLEU scores $\forall n\in\{1,2,3,4\}$ into a single performance value, and it was designed to improve correlation with human judgments.
Some variants of this metric also introduce the semantic match for addressing the reference-translation variability problem.
These variants allow morphological variation, synonyms, stems, and paraphrases to be recognized as valid translations.

The classic version of the METEOR score between a candidate translation and a reference translation is defined as follows.
First, the $uni$-gram precision $P$ and recall $R$ are respectively calculated by:
\begin{equation}\label{eq:precision-recall}
\begin{split}
    P=&\frac{u}{u_c}, \text{ and} \\
    R=&\frac{u}{u_r},
\end{split}
\end{equation}
\noindent where $u$ is the number of $uni$-grams in the candidate translation that also appear in the reference translation, $u_c$ is the number of $uni$-grams in the candidate translation, and $u_r$ is the number of $uni$-grams in the reference translation.

Next, the $F_{mean}$ is computed by combining the precision and recall (defined in Equation~\ref{eq:precision-recall}) via a parameterized harmonic mean~\citep{Rijsbergen1979InformationRetrieval}: 
\begin{equation}\label{eq:f-mean}
    F_{mean}=\frac{P \times R}{\alpha \times P + (1-\alpha) \times R},
\end{equation}


To account for gaps and differences in word order, METEOR calculates a fragmentation penalty.
This calculation uses the total number of matched \textit{uni}-grams $u$ (averaged over candidate and reference) and the number of chunks $ch$, where a chunk is defined as a set of adjacent \textit{uni}-grams in the candidate and reference.
The penalty $p$ is then computed by:
\begin{equation}
    p=\gamma\left(\frac {ch}{u}\right)^\beta,
\end{equation}
\noindent where $\gamma$ determines the maximum penalty ($0 \leq \gamma \leq 1$), and $\beta$ determines the functional relation between the fragmentation and the penalty.
The penalty has the effect of reducing the $F_{mean}$ (defined in Equation~\ref{eq:f-mean}) if there are no $bi$-grams or longer matches.

The final METEOR score is then calculated by:
\begin{equation}\label{eq:meteor}
    \text{METEOR}=F_{mean}\times(1-p)    
\end{equation}


Another metrics that also considers both precision and recall is \textbf{Consensus-based Image Description Evaluation (CIDEr)} considers.
CIDEr was originally proposed for image captioning by \citet{Vedantam2015CIDEr:Evaluation} and is based on the consensus protocol and the average cosine similarity between $n$-grams in the candidate caption and the reference captions.

Intuitively, CIDEr measures how often $n$-grams in the candidate caption are present in the reference captions. 
For this, CIDEr assumes that $n$-grams that do not appear in the reference captions should not be in the candidate sentence.
While $n$-grams that commonly occur across the corpus should be given lower weights, they are likely to be less informative. 
Given these assumptions, a \emph{Term Frequency Inverse Document Frequency (TF-IDF)} weight is calculated for each $n$-gram.

Given the $i$-th input video $x_i$, the set of reference captions $Y_i = \{y_{i1}, \ldots, y_{im}\}$ and a candidate caption $\hat{y}$, all words in both candidate and references are first mapped to their stem or root forms.
For example, ``generates'', ``generation'', ``generated'' and ``generating'' are all reduced to ``generate''. 
Then, the TF-IDF weighting $g_k(y_{ij})$ for $k$-th $n$-gram $\omega_k$ and the $y_{ij}$ reference caption is obtained by:
\begin{equation}
    g_k(y_{ij}) = \frac{h_k(y_{ij})}{\sum_{\omega_l\in\Omega}h_l(y_{ij})}\log \Big(\frac{|V|}{\sum_{x_p\in V}\min(1, \sum_q h_k(y_{pq}))}\Big),
\end{equation}

\noindent where $h_k(y_{ij})$ represents the number of times that the $n$-gram $\omega_k$ occurs in the reference caption $y_{ij}$, $\Omega$ is the vocabulary of all $n$-grams, and $V$ is the set of all videos in the dataset.

The average cosine similarity between them calculates the $\text{CIDEr}_n$ score for a particular $n$-gram between the candidate caption $\hat{y}_i$ and the reference captions $Y_i$: 
\begin{equation}\label{eq:cider_n}
    \text{CIDEr}_n(\hat{y}_i, Y_i) = \frac{1}{m}\sum_j\frac{\textbf{g}^n(\hat{y}_i) \times \textbf{g}^n(y_{ij})}{\norm[1]{\textbf{g}^n(\hat{y}_i)}\norm[1]{\textbf{g}^n(y_{ij})}},
\end{equation}

\noindent where $\textbf{g}^n(\hat{y}_i)$ is a vector formed by the $g_k(\hat{y}_i)$ weights and $\textbf{g}^n(y_{ij})$ is a vector formed by $g_k(\hat{y}_i)$ weights, corresponding to all $n$-grams of length $n$ of the candidate $\hat{y}_i$ and the references $y_{ij}$, respectively. 
$\norm[1]{\textbf{g}^n(\hat{y}_i)}$ and $\norm[1]{\textbf{g}^n(y_{ij})}$ are the magnitudes of these vectors.

To capture richer semantics and grammatical properties, the authors proposed to compute a higher-order (longer) $n$-grams.
They combined the scores from $n$-grams varying lengths as follows:
\begin{equation}
    \text{CIDEr}(\hat{y}_i, Y_i)=\sum_{n=1}^N w_n \text{CIDEr}_n(\hat{y}_i, Y_i)
\end{equation}
\noindent Here, the authors found that uniform weights $w_n=1/N$ and $N=4$ work the best, empirically.
Likewise, for benchmark evaluation on image and video captioning, the most popular version of this metric, $\text{CIDEr-D}$~\citep{Vedantam2015CIDEr:Evaluation, Chen2015MicrosoftServer}, is used.
$\text{CIDEr-D}$, among other modifications, prevents higher scores for descriptions that erroneously fail for human judgments. CIDEr-D does this, by clipping, for a specific n-gram, the number of candidate occurrences to the number of reference occurrences.
These modifications result in the following version of Equation~\ref{eq:cider_n}:
\begin{equation}\label{eq:ciderd_n}
    \text{CIDEr-D}_n(\hat{y}_i, Y_i) = \frac{10}{m}\sum_j \Bigg(\exp \bigg(\frac{-l(\hat{y}_i)-l(y_{ij})^2}{2\sigma^2}\bigg) \times \frac{\min(\textbf{g}^n(\hat{y}_i), \textbf{g}^n(y_{ij})) \cdot \textbf{g}^n(y_{ij})}{\norm[1]{\textbf{g}^n(\hat{y}_i)}\norm[1]{\textbf{g}^n(y_{ij})}}\Bigg),
\end{equation}
\noindent where $l(\hat{y}_i)$ and $l(y_{ij})$ denote the lengths of candidate and reference sentences respectively, and $\sigma = 6$ is usually used.
This penalty and the factor ten produce that, unlike other metrics, the scores can be higher than one but lower than ten.



\subsubsection{Explicit Semantic Match-based Metrics}

Beyond $n$-grams, the semantic content matching metrics measure the similarity between the generated caption and the reference captions in the dataset by extracting explicit semantic information units.
These metrics have shown well-correlation with human judgments.

The \textbf{Semantic Proportional Image Captioning Evaluation (SPICE)}\label{metric:spice} metric~\citep{Anderson2016SPICE:Evaluation} is based on tuples of scene graphs of the generated candidate description and all reference descriptions. 
These scene graphs encode the related video in a skeleton form, parsing the descriptions using two stages: (1) syntactic dependencies extraction and (2) a rule-based graph mapping.
This skeleton represents the texts as semantic tokens, such as object classes, relationship types, or attribute types. 
After a proper parsing, SPICE then computes the \emph{F-SCORE} using the candidate and reference scene graphs over the conjunction of logical tuples representing semantic propositions in the scene graph.

Better than improved versions of METEOR and CIDEr, SPICE also uses synonyms match by the encoding into semantic scene graphs.
However, one limitation (not widely reported) of this metric is its dependency on the parsing quality.
Another major limitation is that even though SPICE correlates well with human evaluations, it ignores the generated captions' fluency~\citep{Celikyilmaz2020EvaluationSurvey, Sharif2018Learning-basedEvaluation}.

\textbf{Semantic Text Similarity (STS)} was introduced in the TREC\-VID 2016 as an experimental semantic similarity metric~\citep{Han2013UMBC_EBIQUITY-CORE:Systems}, and it has been used in this competition~\citep{Awad2016TRECVIDHyperlinking, Awad2017TrecvidHyperlinking, Awad2018TRECVIDSearch, Awad2019TRECVIDRetrieval, Awad2020TRECVIDDomains}, in addition to BLEU, METEOR, and CIDEr metrics.
This metric measures how semantically similar is the submitted description to the ground-truth description.
It is based on distributional similarity, and Latent Semantic Analysis (LSA) complemented with semantic relations extracted from WordNet.

\subsubsection{Limitations}\label{sec:eval-limitations}
Most video captioning metrics have been adopted from machine translation or image captioning and do not well-measure the quality of the generated descriptions from videos.
Automated metrics should agree well with human judgment to be useful in practice.
In this sense, BLEU and ROUGE are not optimal~\citep{Reiter2018ABLEU, Schluter2017TheROUGE} and have been shown to correlate with humans weakly~\citep{Vedantam2015CIDEr:Evaluation, Hodosh2015FramingAbstract}. 
In contrast, the METEOR and CIDEr metrics have shown a better correlation with humans when judging the overall quality of language descriptions for image and video captioning tasks.
Specifically, CIDEr shows a high agreement with consensus as assessed by humans~\citep{Vedantam2015CIDEr:Evaluation}.
Simultaneously, METEOR addresses the problem of reference description variability, allowing the recognition of morphological variants and synonyms as adequate descriptions.

Due to this discrepancy between metrics, the automatic evaluation is often accompanied by human evaluators accessing the candidate texts for their intrinsic qualities (\eg correctness, fluency, factuality, and discourse coherency and consistency)~\citep{Celikyilmaz2020EvaluationSurvey} and human preferences.
Human evaluations have several challenges.
They are expensive, time-consuming and the people who are going to evaluate will require extensive domain expertise.
To ease this process, some crowd-sourcing platforms like Amazon Mechanical Turk (AMT) exist.
We can post the method's outputs with a detailed description of how people must evaluate, and we get our results based on convinced voting.
However, that also has downsides, such as checking evaluations and evaluators' consistency and quality without knowing them.

In the VTT task of the TRECVID Challenge (see Section~\ref{sec:challenge:trecvid}), the organizers address this limitation by evaluating the automatic video descriptions with the \emph{Direct Assessment (DA)} method~\citep{Awad2020TRECVIDDomains} in addition to standard metrics.
DA is an accurate method proposed by \cite{Graham2018EvaluationAssessment} to control crowd-sourcing quality, which is efficient and cost-effective for tuning automatic evaluation metrics.
They addressed the limitations of crowd-sourced judgments by automatically modifying candidate captions and degrading their quality, and examining how human judges rated the degraded captions.
They demonstrated that this process is ten times more effective at finding significant differences between competing systems than previous methodologies.
Furthermore, the use of DA for evaluating the variable number of annual participants in the TRECVID-VTT task has shown its scalability, robustness, and replicability.


\section{Methods for Matching \& Ranking Video Descriptions}\label{sec:matching-ranking}

In the previous section, we saw how the state-of-the-art methods for video captioning produce texts from videos.
Particularly, we detailed how encoder-decoder networks could be used to model text production, where the input video is first encoded into a continuous representation that is the input to the decoder.
The decoder then predicts the output words sequence, one word at a time, conditioned both on the input representation and the previously generated word.

In contrast, the retrieval methods aim to return an ordered list (ranking), in which the text descriptions most likely to match (annotated) with the video appear first.
For achieving this aim, the models are generally trained in a \emph{multimodal learning} way~\citep{Goodfellow2016DeepLearning}.
The model must capture a representation in video modality, a representation in text modality, and in a joint distribution that relates both modalities.

This joint distribution is also called \emph{common embedding space} and is usually trained for ensuring that visual and textual samples are close if and only if they are semantically similar.
As we introduced in Section~\ref{sec:video-matching}, this common embedding space enables multiple applications in vision-language tasks.
Some of these tasks are text-to-image/video retrieval and image/video-to-text retrieval, visual question answering~\citep{Yu2017EndAnswering, Tapaswi2016MovieQA}, video summarization with natural language~\citep{Plummer2017EnhancingEmbedding}, and temporal grounding (\ie temporal localization) in videos ~\citep{Bojanowski2015WeaklyText, Hendricks2017LocalizingLanguage}.
Almost all of these works show that the vision-language embedding need not be trained on domain-specific data but can be learned from standard still image-language datasets and transferred to video. 
Therefore, understanding how video-text retrieval methods work can help you develop new models for any of these tasks.
Any technique that improves the \textbf{text-video embeddings} has the potential improve any method relying on such representations.

The authors have proposed two models, principally: (1) based on \emph{single encoding}; and (2) based on \emph{siamese encoding}.
These two categories of models implicate that the similarity rank can be computed into three different ``spaces'': the visual features space, the textual features space, and the joint embedding space.
Into each one of these categories, there are some techniques widely used in state-of-the-art research, such as \emph{concepts-guiding retrieval}, \emph{dual encoding}, and \emph{hierarchical learning}.
This section analyzes the most important works for each technique, summarized in in Table \ref{tab:MR}, according to their distinctive characteristics.

\begin{figure*}[!t]
    \centering
    \includegraphics[width=.8\textwidth]{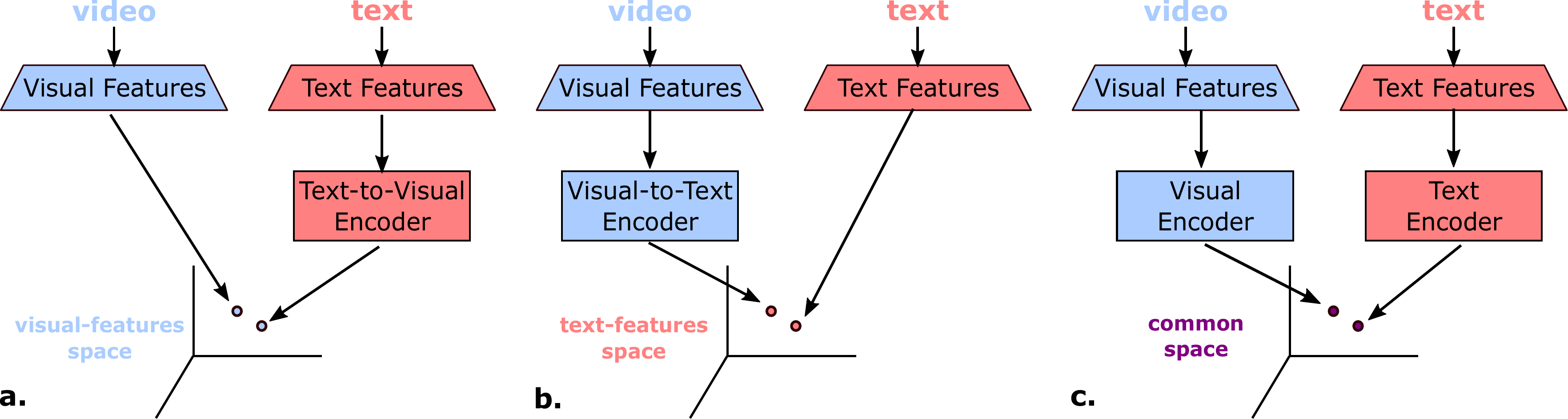}
    \caption{Matching-and-ranking techniques. In \textbf{a.}, the similarity is determined in the visual feature space, learning to map from textual features to visual features. In \textbf{b.}, the similarity is computed in the textual feature space, learning to map from visual features to textual features. And in \textbf{c.}, we learn to map from visual and textual features to a common space.}
    \label{fig:mr-spaces}
\end{figure*}

\subsection{Single Encoding-based Methods}

Early deep learning-based retrieval methods aim to learn to encode only one of the modalities (text or video). 
The other modality was encoded by pre-trained models, which were not updated in the training process. 
In other words, these models learned to convert from only one modality to a representation of the other modality computed with a pre-trained model. 
When the fixed model is the \emph{video encoder}, we say that the similarity is computed in the video feature space (see Figure \ref{fig:mr-spaces}a)~\citep{Snoek2016UniversityVideo, Snoek2017UniversityVideo}. 
In contrast, when the fixed model is the \emph{text encoder}, we say that the similarity is computed in the textual feature space (see Figure \ref{fig:mr-spaces}b)~\citep{Le2016NII-HITACHI-UIT2016, Marsden2016Dublin2016, Markatopoulou2016ITI-CERTH2016, Yang2016SemanticUnderstanding}. 

The \emph{video encoder} learned by \citet{Marsden2016Dublin2016} was based on the combination of three semantic representations of ten sampled \textit{key-frames} of the video.
These three semantic representations are: a 365-dimensional \textit{place} representation, associated with the prediction of different scene categories (\eg airport terminal, cafeteria, and hospital room); a 1,000-dimensional \textit{object} representation; and a 94-dimensional \textit{action} representation, each one referring to a crowd behavior concept (\eg fight, run, mob, parade, and protest).
The authors fine-tuned the VGG-16 architecture~\citep{Simonyan2015VeryRecognition} pre-trained on the ImageNet dataset and Places2 dataset~\citep{Zhou2018Places:Recognition} for the \textit{object} and \textit{place} representations, respectively.
While for the \textit{action} representation, the authors fine-tuned the AlexNet architecture~\citep{Krizhevsky2012ImageNetNetworks} pre-trained on the WWW dataset~\citep{Shao2015DeeplyUnderstanding}.
Finally, to retrieve the captions, they computed the similarity scores combining the word2vec~\citep{Li2015Zero-shotEmbedding} representations of video representations and the \textit{objects}, \textit{places}, and \textit{actions} that appear in each description of the corpus. 

The incorporation of semantic representation based on an explicit concept detector model has also been explored by other works~\citep{Chen2017Informedia2017, Zhang2016VIREODescription, Yang2016SemanticUnderstanding}, assuming that one important clue is whether a concept appears in the video.
However, as we analyzed for SeG-based video captioning approaches in Section~\ref{sec:dg-semantic-based}, training the detectors could be a drawback.
The training data (frames-annotated) could be not enough and can produce a harmful dependency.
These models require answering several questions to be adopted, \eg how to specify the concepts, how to train good classifiers for these concepts, and how to select relevant concepts for the video.
For that, recent state-of-the-art works attempt to deal with the retrieval task in a concept-free manner~\citep{Dong2019DualRetrieval, Ging2020COOT:Learning}.


In contrast to models that learn video encoders only, \citet{Snoek2016UniversityVideo} and \citet{Snoek2017UniversityVideo} described an approach based on capturing video representations from texts (see Figure \ref{fig:mr-spaces}a.). 
They used these representations of the corpus' captions to calculate an input video's relevance scores right there, in the video representation space.
That approach is Word2VisualVec~\citep{Dong2016Word2VisualVec:Prediction}, which uses a 500-dimensional sentence representation as input. 
The model proposed by~\citet{Snoek2016UniversityVideo} first computes this sentence representation by mean-pooling the embeddings of each word in the sentence,
In contrast, the model proposed by~\citet{Snoek2017UniversityVideo} incorporated a GRU network for processing the texts.
The word embeddings were computed with the word2vec model \citep{Li2015Zero-shotEmbedding} pre-trained on a Flickr tags' corpus.
Then, an MLP maps the sentence representation in a 2048-dimen\-sional video feature space.
The video representations are the concatenation of the average of sampled frames' visual features (extracted with a pre-trained GoogleNet model~\citep{Szegedy2015GoingConvolutions}) and a 1024-dimensional bag of quantized Mel-frequency Cepstral Coefficients (MFCC) vectors~\citep{Davis1980ComparisonSentences, Wang2003SurveyAnalysis}. 
Finally, they trained the models minimizing the \emph{Mean Squared Error (MSE)} between the vectors' sentence descriptions and vectors extracted from the video. 

\subsection{Siamese Encoding-based Methods} 

The single encoding-based methods take advantage of data but to some extent.
The representations learned by those models do not benefit from all the training data available to perform this \emph{zero-shot learning} task~\citep{Goodfellow2016DeepLearning}.
Another way to attack the retrieval of textual descriptions is to learn two mapping functions (one from video representation and another from caption representation) and a relation function between the two mapped representations (see Figure~\ref{fig:mr-spaces}c.).
For simplification, we can incorporate the relation function as part of the last operations performed by each mapping function, and the aim is then to learn the functions $f(\cdot)$ and $g(\cdot)$ (encoders) that project videos and texts into a joint embedding space\footnote{Joint embeddings are usually done by mapping semantically associated inputs from two or more domains (\textit{e.g.}, images and text) into a common vector space} respectively.
These encoders anchor the concepts in one representation in the other and vice versa~\citep{Goodfellow2016DeepLearning}.
Section~\ref{sec:loss-mr} explains how to train these models to enable \textbf{zero-data learning}, and we now analyze the architecture of some state-of-the-art works based on this approach.

One of the first solutions that train the parameters of both encoders was proposed by \citet{Le2016NII-HITACHI-UIT2016} and \citet{Phan2017NII-Hitachi-UIT2017}.
These methods are based on Embedding Convolutional Neural Network (ECNN)~\citep{Yang2016SemanticUnderstanding}, which defines $f(\cdot)$ as the 
single-frame approach (see Section \ref{sec:single-frame}) and $g(\cdot)$ as the ParagraphVector~\citep{Le2014DistributedDocuments} with the Bag-of-Words (BoW) approach.
They were among the first to employ 3D-CNN to generate the frame-level representations instead of the 2D-CNN used in ECNN~\citep{Yang2016SemanticUnderstanding}. 
While \citet{Le2016NII-HITACHI-UIT2016} trained the model minimizing the $L_2$ distances between two representations, \citet{Phan2017NII-Hitachi-UIT2017} obtained better results by defining the similarity measure $s(\cdot,\cdot)$ in the common space as the dot product:
\begin{equation}
    s(x,y) = f(x)\cdot g(y),
\end{equation}

\noindent where $x$ represents a video and $y$ represents a text. 


More recently, \citet{Mithun2017CMU-UCR-BOSCHRETRIEVAL} observed that the existing image-caption datasets are larger and more varied than the existing video-caption datasets. 
They assumed that, when retrieving a video description, a few \textit{key-frames} are enough to summarize the entire video. 
Hence, the authors proposed to extract the \textit{key-frames} of the video and treat the problem as image-text retrieval. 
For this, once the model extracts the 2D-CNN from the input video frames employing AlexNet~\citep{Krizhevsky2012ImageNetNetworks}, a sparse coding-based approach~\citep{Elhamifar2016Dissimilarity-basedSelection} is used to obtain a representative subset of frames. 
Given this subset, they tuned a joint embedding pre-trained on the MS-COCO dataset~\citep{Chen2015MicrosoftServer}. 
For encoding images/frames, the authors used the ResNet-152 model~\citep{He2016DeepRecognition}, and for encoding texts, they employed a GRU-based text encoder~\citep{Cho2014LearningTranslation}. 
For training, the authors used cosine similarity between the embedded visual features and text descriptors.
\citet{Chen2017Informedia2017} proposed a model that combines, in a distinctive way, a concept detector model with a joint embedding model, adding the scores from both models and training them simultaneously.
Specifically, for the text encoder and joint embedding, the authors proposed three methods. 
In the first method, they first calculate the dot product between the embedding of each word and the video and then obtain the final score for each caption averaging them. 
In the second method, trying to avoid a limitation of the first one (deals with each word separately), the authors used convolution operations with kernels in the count of words: one, two, and three, learning the sentence's local structure. 
The final score is then the sum of scores obtained from the representations produced for each kernel.
In the last method, they improved the second one keeping in mind that not all the words have the same weight in the sentence. 

In contrast, \citet{Yu2017Shandong2017} proposed to learn the mapping matrices $U$ and $V$ following the objective function:
\begin{equation}
    \text{min}_{U,V} f(U,V,X,Y) = C(U,X,Y) + L(U,V,X) + N(U,V),
\end{equation}
\noindent where $C(\cdot,\cdot,\cdot)$ is a linear regression for keeping the frames with the same semantics close, $L(\cdot,\cdot,\cdot)$ is a correlation analysis term, and $N(\cdot,\cdot)$ is the regularization term.

Specifically, the authors defined $C(\cdot,\cdot,\cdot)$ as:
\begin{equation}
    C(U,X,Y) = \beta ||X^TU-Y||^2_F,
\end{equation}
\noindent where $0\leq\beta\leq1$, $X$ and $Y$ denote the feature matrices of frames and semantics respectively, and $U$ is a mapping matrix of weights to be learned. 
To obtain $Y$, the authors pre-trained a semantic detector on Wikipedia and Pascal Sentence image-vector pairs datasets~\citep{Rashtchian2010CollectingTurk} with ten and twenty categories, respectively. 
They extracted 2048-dimensional CNN features for frames, and for captions, they extracted 100-dimensional features with the Sentence2Vector model~\citep{Saha2017Con-S2V:Sen2Vec}.



\subsubsection{Dual Encoding-based Methods}

More recently, \citet{Dong2019DualRetrieval} and \citet{Ging2020COOT:Learning} proposed to use the same architecture for both encoders. 
The architecture proposed by \citet{Dong2019DualRetrieval} is based on multi-level encoding.
In contrast, the architecture proposed by \citet{Ging2020COOT:Learning} is based on multi-contextual embeddings. 
In the first one~\citep{Dong2019DualRetrieval}, three encoding levels are combined inside each branch, and then, the relationship between these combinations is captured in the joint embedding. 
In the second one~\citep{Ging2020COOT:Learning}, two distinct levels of granularity of texts and videos are combined using two contextual embeddings, and then, the final embedding is a combination of them.

Specifically, the encoding levels explored by \citet{Dong2019DualRetrieval} are: a global encoding obtained by mean pooling, a temporal-aware encoding computed by bi-directional GRU (biGRU), and a local-enhanced encoding resulted from a multi-kernel convolution of the biGRU outputs.
While the granularity levels by contextual embeddings explored by \citet{Ging2020COOT:Learning} are: local context learned by projecting the sentences (parts of the paragraph) and clips (segments of the video), and global context learned considering the full video and the complete paragraph and transforming the local embedding.
As we mentioned in Section~\ref{sec:dg-joint-embedding}, the contextual embeddings are obtained by transformer-based models~\citep{Vaswani2017AttentionNeed}, and their optimization process is analyzed in the next Section~\ref{sec:loss-mr}.


Based on the multi-level encoding~\citep{Dong2019DualRetrieval}, \citet{Song2019RUC_AIM3Text} also proposed a model of three encoding levels.
However, in the text-side encoder, they use in the first level a mean pooling of \textbf{word embeddings} initialized by GloVe instead of word2vec~\citep{Mikolov2013DistributedCompositionality, Li2015Zero-shotEmbedding}. 
Besides, they based the architecture on BERT instead of biGRU and fine-tuned its last layer. 
Finally, they employed the TRECVID VTT 2016~\citep{Awad2016TRECVIDHyperlinking}, TRECVID VTT 2017~\citep{Awad2017TrecvidHyperlinking}, TGIF~\citep{Li2016TGIF:Description}, MSR-VTT~\citep{Xu2016MSR-VTT:Language}, and \VATEX{}~\citep{Wang2019VaTeXResearch} video captioning datasets as the train set, and TRECVID VTT 2018~\citep{Awad2018TRECVIDSearch} as the validation set. 
For video representation, they extracted features from ResNeXt-101~\citep{Xie2017AggregatedNetworks}, I3D, and audio (VGGish).


\subsection{Loss Functions for Optimization}\label{sec:loss-mr}
As we have explained throughout this section, recent methods for matching \& ranking video descriptions use two encoders $f(\cdot)\in \mathbb{R}^d$ and $g(\cdot)\in \mathbb{R}^d$ for mapping the videos and captions into a joint space, respectively.
After defining these mapping functions' structure, we need to determine how to train them to obtain accurate representations that capture both modalities' relationship.
The \emph{triplet-ranking-loss} function is a standard margin-based loss for training retrieval models.
Motivated on the nearest-neighbor classification technique~\citep{Weinberger2005DistanceClassification}, \citet{Schroff2015FaceNet:Clustering} proposed this loss function for the Face Recognition problem~\citep{Parkhi2015DeepRecognition}.

Let $(x,y)\in \mathcal{D}$ a video-description pair of the dataset and assume that $y^\star$ is an arbitrary description in the dataset not associated with $x$ (a negative example).
Their projections in the joint space are $f(x) = z_x$, $g(y) = z_y$, and $g(y^\star) = z_{y^\star}$.
Assume $\text{dist}(\cdot,\cdot)$ is a distance function in $\mathbb{R}^d$.
This loss function tries to ensure that the distance between the anchor $z_x$ and positive encoding $z_y$ is less than the distance between the anchor and any negative encoding. 
We would like the following to hold:
\begin{equation}
    dist(z_x, z_y) + \alpha < dist(z_x, z_{y^\star}),
\end{equation}
\noindent where $\alpha$ is a margin that is enforced between positive and negative pairs.
This gives  rise  to  a  natural  optimization problem in which one wants to minimize:
\begin{equation}
    L_\Theta = \max\bigg\{\, 0,\; \text{dist}\big(z_x,\,z_y\big) + \alpha - \text{dist}\big(z_x,\,z_{y^\star}\big)\bigg\}
\end{equation}



This formulation is the classic version of triple-ranking loss and is used in many different applications with the same formulation or minor variations. However, in the literature, it can be found under other names such as ranking loss, margin loss, contrastive loss, triplet loss, and hinge loss, which can be confusing.
For VTT, researchers have adapted several distance functions and triplets selection approaches.
For example, \citet{Dong2019DualRetrieval} proposed to use the improved marginal ranking loss~\citep{Faghri2018VSE++:Negatives}, which penalizes the model taking into account the hardest negative examples.
In this improved version, one considers a tuple 
$(x,y,x^\star,y^\star)$ where $(x,y)\in \mathcal{D}$, $y^\star$ is the closest negative example for $x$, and $x^\star$ is the closest negative example for $y$.
Their projections in the joint space are $f(x) = z_x$, $g(y) = z_y$, $f(x^\star) = z_{x^\star}$, and $g(y^\star) = z_{y^\star}$.
They trained the model attending to the loss function:
\begin{equation}\label{eq:loss:improved-triplet-loss}
\begin{split}
    L_\Theta' = &\max\bigg\{\, 0,\; \text{dist}\big(z_x,\,z_y\big) + \alpha - \text{dist}\big(z_x,\,z_{y^\star}\big)\bigg\} \\
                & + \max\bigg\{\, 0,\; \text{dist}\big(z_x,\,z_y\big) + \alpha - \text{dist}\big(z_{x^\star},\,z_y\big)\bigg\}
\end{split}
\end{equation}
\noindent where the distance function $\text{dist}(\cdot,\cdot)$ they used was the cosine distance function.



Recently, \citet{Ging2020COOT:Learning} have proposed another form of training their cooperative hierarchical models.
In the context of video-paragraph datasets, they introduced a \emph{cross-modal cycle-consistency loss} to enforce the semantic alignment between clips and sentences.
The cycle-consistency uses transitivity as an objective for training~\citep{Chen2019Cycle-ConsistencyAnswering, Dwibedi2019TemporalLearning, Wang2019LearningTime, Zhu2017UnpairedNetworks}, and it has been used for tasks like temporal video alignment~\citep{Dwibedi2019TemporalLearning} and visual question answering~\citep{Chen2019Cycle-ConsistencyAnswering}.
In contrast to previous models~\citep{Dwibedi2019TemporalLearning} that only work on the video domain, \citet{Ging2020COOT:Learning} aligned video and text in a cross-modal way.
They defined the cyclical structure as follows.

Given a sequence of video clip embeddings $Z_x=\{z_{x1}, \ldots, z_{xn}\}$ of an input video $x$ and sentence embeddings $Z_y=\{z_{y1}, \ldots, z_{ym}\}$ of the reference paragraph $y$.
We first find the soft nearest neighbor~\citep{Dwibedi2019TemporalLearning} of each sentence embedding $z_{yi}$ in the clip sequence by:
\begin{equation}
    \hat{z_x} = \sum_{j=1}^n \alpha_j z_{xj},
\end{equation}
\noindent where $\alpha_j$ represents the similarity score of clip embedding $z_{xj}$ to sentence embedding $z_{yi}$, computed by:
\[
\alpha_j = \frac{\exp(-\norm[1]{z_{yi}-z_{xj}}^2)}{\sum_{k=1}^n\exp(-\norm[1]{z_{yi}-z_{xk}}^2)}
\]

\noindent We then cycle back from $\hat{z_x}$ to the sentences embeddings sequence $Z_y$ and calculate the soft location by:
\begin{equation}
    \hat{i} = \sum_{j=1}^m \beta_j j,
\end{equation}
\noindent where $\beta_j$ represents the similarity score of sentence embedding $z_{yj}$ to the soft clip embedding $\hat{z}_{x}$, computed by:
\[
\beta_j = \frac{\exp(-\norm[1]{\hat{z}_x-z_{yj}}^2)}{\sum_{k=1}^n\exp(-\norm[1]{\hat{z}_{x}-z_{yk}}^2)}
\]

Given these to locations $i$ and $\hat{i}$ in the sequence of sentence embeddings $Z_y$, we assume that the embedding $z_{yi}$ is semantically cycle-consistent if and only if it cycles back to the original location, \ie $i = \hat{i}$.
This loss aims to penalize deviations from cycle-consistency for sampled sets of clips and sentences, which encourages semantic alignment between vision and text representations in the joint embedding space.
The objective is then, the distance between the source index $i$ and the soft destination index $\hat{i}$:
\begin{equation}
    \mathcal{L}_{CMC} = \norm[2]{i-\hat{i}}^2
\end{equation}

For the final loss of the model, the authors combined $\mathcal{L}_{CMC}$ with four other losses.
Three of these losses are based on ranking losses (Equation~\ref{eq:loss:improved-triplet-loss}) obtained from clip-sentence level, video-paragraph level, and global context level.
The other one, proposed by \citet{Zhang2018Cross-ModalText}, models the clustering of low-level and high-level semantics in the joint embedding space:
\begin{equation}
    \mathcal{L}_{cluster} = \mathcal{L}_{low} + \mathcal{L}_{high}
\end{equation}

\noindent Here, the $\mathcal{L}_{low}$ part models the low-level semantics clustering by the sum of $L_\Theta'$ (defined in Equation~\ref{eq:loss:improved-triplet-loss}) for all tuples $(z_{xi}, z_{yi}, z_{x'j}, z_{y'j})$ such that $(x', y')\in\mathcal{D}$ and $x' \neq x$.
So, $z_{x'j}$ and $z_{y'j}$ are clip-sentence-level intra-modality negative examples such that $z_{x'j}$ is a negative example for $z_{xi}$ computed by the video encoder $f(\cdot)$, and $z_{y'j}$ is a negative example for $z_{yi}$ computed by the text encoder $g(\cdot)$.
\[
\begin{split}
\mathcal{L}_{low} = \sum_i \sum_{j, (x', y')\in \mathcal{D}, x' \neq x} \Bigg( 
&\max\bigg\{\, 0,\; \text{dist}\big(z_{xi},\,z_{yi}\big) + \gamma - \text{dist}\big(z_{xi},\,z_{x'j}\big)\bigg\} \\
&+ \max\bigg\{\, 0,\; \text{dist}\big(z_{xi},\,z_{yi}\big) + \gamma - \text{dist}\big(z_{y'j},\,z_{yi}\big)\bigg\} \Bigg),
\end{split}
\]
\noindent where $\gamma$ is a constant margin.

The $\mathcal{L}_{high}$ part captures the high-level semantics clustering by the sum of $L_\Theta'$ (defined in Equation~\ref{eq:loss:improved-triplet-loss}) for all tuples $(z_x, z_y, z_{x'}, z_{y'})$ such that $(x', y')\in\mathcal{D}$ and $x' \neq x$.
So, $z_{x'}$ and $z_{y'}$ are video-paragraph-level intra-modality negative examples such that $z_{x}$ is a negative example for $z_{x}$ computed by the video encoder $f(\cdot)$, and $z_{y'}$ is a negative example for $z_{y}$ computed by the text encoder $g(\cdot)$.
\[
\begin{split}
\mathcal{L}_{high} = \sum_i \sum_{(x', y')\in \mathcal{D}, x' \neq x} \Bigg( 
&\max\bigg\{\, 0,\; \text{dist}\big(z_x,\,z_y\big) + \rho - \text{dist}\big(z_x,\,z_{x'}\big)\bigg\} \\
&+ \max\bigg\{\, 0,\; \text{dist}\big(z_x,\,z_y\big) + \rho - \text{dist}\big(z_{y'},\,z_y\big)\bigg\} \Bigg),
\end{split}
\]
\noindent where $\rho$ is a constant margin.

\subsection{Retrieval-specific Evaluation Metrics}\label{sec:metrics-mr}

As we explained for description generation-specific evaluation metrics in Section~\ref{sec:metrics-dg}, almost all metrics used to evaluate the VTT methods come from other related tasks.
Specifically, for evaluating the output of retrieval-based methods, authors have reported some automatic assessments from the information retrieval field.
Below we review the most-reported measures for these methods: Mean Reciprocal Rank (MIR) and Recall at K (R@K).

A standard evaluation metric for information retrieval methods is the \emph{Mean Inverted Rank}~\citep{Craswell2009MeanRank}.
It is a statistical measure that calculates the average over a set of $n$ queries.
Different scores are attributed inversely proportional to the rank of the first correct answer in the answer list.
\begin{equation}\label{ec:MIR}
\text{MIR} = \frac{1}{n}\sum_{i=1}^{n}\frac{1}{rank_{i}},
\end{equation}
\noindent where $rank_{i}$ is the rank position of the first correct answer of the $i$-th query.

$\text{MIR}$ has been used in the Matching \& Ranking subtask of the TRECVID challenge from 2017 but has not been frequently reported in the literature of video-text retrieval methods.
It is appropriate to evaluate models where there is only one relevant result or only cares about the highest-ranked one. 

In the context of recommendation systems, we are most likely interested in recommending top-$k$ items to the user.
So it makes more sense to compute \emph{precision} and \emph{recall} metrics in the first $k$ items instead of all the items.
Thus, in the context of VTT, the notion of precision and recall at $k$, where $k$ is a user-definable integer to match the top-$k$ recommended descriptions.
\emph{Recall at k} is then the proportion of relevant descriptions found in the top-$k$ recommendations:
\begin{equation}
    \text{R@k} = \frac{r@k}{r},
\end{equation}
\noindent where $r@k$ is the number of recommended descriptions at $k$ that are relevant, and $r$ is the total number of relevant descriptions. 

Several state-of-the-art works report R@k results for some values of $k$, \eg R@1, R@5, R@10, R@50.
These works also report the \emph{mean average precision (mAP)} and \emph{median rank (MR)} scores.
The \emph{mAP} score considers whether all of the relevant items tend to get ranked highly.
Then, when there is only one relevant answer in the corpus, $\text{MIR}$ and $\text{mAP}$ are exactly equivalent under the standard definition of $\text{mAP}$.
For comparing methods performance, higher $\text{R@k}$ and $\text{mAP}$, and smaller $\text{MR}$ indicates better performance.
Usually, for evaluating the quality of learned joint embeddings, the authors also report these scores for the opposite problem: video retrieval from an input text.


\subsection{Summary}

\begin{table*}[t]
    \fontsize{4.7pt}{4.7pt}\selectfont
    \centering
    \begin{tabular}{c|c|ccc|cc}
\toprule
 Space           & Method                                       &\multicolumn{3}{c|}{Visual branch}                              &\multicolumn{2}{c}{Textual branch}\\
                 &                                              & Features                   & CD            & Encoder           & Features      & Encoder           \\
 \midrule
Textual features& \citet{Marsden2016Dublin2016}                & VGG, AlexNet, SVO          &               & weighted sum      & word2vec      & average           \\
                & \citet{Yang2016SemanticUnderstanding}        & CaffeNet                   &\checkmark     & ECNN              & word2vec      & average           \\
\midrule            
  Visual features & \citet{Snoek2016UniversityVideo}             & GoogLeNet, MFCC            &               & average           & word2vec      & Word2VisualVec    \\
                & \citet{Snoek2017UniversityVideo}             & ResNeXt-101                &               & average           & BoW, word2vec & GRU               \\
                & \citet{Chen2017Informedia2017}               & ResNet, i3D                &\checkmark     & average           & word2vec      & TCNN, attention   \\ 
\midrule    
Common          & \citet{Zhang2016VIREODescription}            & VGG, C3D                   &\checkmark     & SAN               & word2vec      & TCNN              \\
                & \citet{Le2016NII-HITACHI-UIT2016}            & C3D                        &               & ECNN              & BoW           & ParagraphVector  \\ 
                & \citet{Yu2017Shandong2017}                   & CNN                        &\checkmark     & FC                & BoW,word2vec  & sentence2vec, FC  \\ 
                & \citet{Mithun2017CMU-UCR-BOSCHRETRIEVAL}     & ResNet                     &               & average, MLP      & word2vec      & GRU               \\ 
                & \citet{Nguyen2017VIREOHyperlinking}          & ResNet, C3D                &               & average, FC       & word2vec      & LSTM              \\
                & \citet{Li2018RenminRetrieval}                & ResNext, ResNet            &               & average, FC       & BoW, word2vec & GRU, FC           \\
                & \citet{Dong2019DualRetrieval}                & ResNeXt-101                &               & biGRU-CNN         & one-hot       & biGRU-CNN         \\
                & \citet{Song2019RUC_AIM3Text}                 & ResNeXt-101, I3D, VGGish   &               & BERT              & GloVe         & BERT              \\
                & \citet{Ging2020COOT:Learning}                & ResNet-152, C3D            &               & Transformer       & GloVe         & Transformer       \\
                
\bottomrule
    \end{tabular}
    \caption{Summary of most important matching-and-ranking methods (sorted by year) presented in the literature to solve the video caption/description retrieval task. For each method we mention the main characteristics of its visual and textual branches. We mention the type of visual/textual features used to represent the videos/descriptions in the Features columns. Besides, we summarize the strategy used to encapsulate and project these features in Encoder columns. And in the case of visual branch, we also check if the method uses concept detectors (CD) as a high-level representation.}
    \label{tab:MR}
\end{table*}

Table \ref{tab:MR} shows a summary of the papers we described for the retrieval-based methods. 
We can point out that there are authors such as \citet{Snoek2016UniversityVideo} that, in 2016, based their proposal on the CNN scheme only, but in 2017, incorporated a recurrent-based encoder for the textual branch (CNN+RNN combination). 
Also, we can notice that, over the years, there has been a tendency to develop methods based on learning a generic representation into a joint space. 
For this, the researchers have proposed different ways of encoding each modality. 
Recent works~\citep{Dong2019DualRetrieval, Ging2020COOT:Learning, Song2019RUC_AIM3Text} proposed to use models with the same structure to compute both encodings with transformer-based architectures~\citep{Ging2020COOT:Learning, Song2019RUC_AIM3Text}.

\section{Evaluation Benchmarks}\label{sec:competitions}

Stimulating  research on the vision-language intersection, researchers have proposed conferences and workshops where VTT is a central topic.
Some of these meetings and some companies have organized exciting VTT competitions (challenges) in recent years.
These contests can be centered on describing videos from any domain or videos from a specific domain, such as LSMDC for movie description or ActivityNet for human activities description.
This section covers these competitions, introducing their evaluation tasks and datasets and explaining how the participant teams are evaluated.
Table~\ref{tab:challenge:participants} shows the number of annual participant teams reported for each competition task.
In the following Section~\ref{sec:datasets}, the reader can find the details on the construction and composition of related datasets.

\begin{table}[t]
    \scriptsize
    \centering
    \begin{tabular}{lccc}
    \toprule
    Challenge, Task & Year & \multicolumn{2}{c}{Participants} \\
                    &      & public test & blind test \\
    \midrule
    LSMDC, Movie Description	            & 2015 & 13	& 13    \\
    LSMDC, Movie Retrieval	                & 2017 & 5	& -     \\
    LSMDC, Movie Multiple-Choice Test	    & 2017 & 11	& -     \\
    LSMDC, Movie Fill-in-the-Blank	        & 2017 & 5	& -     \\
    ActivityNet, Dense Captioning Events	& 2017 & -	& 10    \\
    ActivityNet, Dense Captioning Events	& 2018 & -	& 18    \\
    ActivityNet, Dense Captioning Events	& 2019 & -	& 33    \\
    ActivityNet, Dense Captioning Events	& 2020 & -	& 15    \\
    ActivityNet, Dense Captioning Events	& 2021 & -	& 10    \\
    \VATEX{}, English Captioning	        & 2019 & -	& 8     \\
    \VATEX{}, Chinese Captioning	        & 2019 & -	& 8     \\
    \VATEX{}, English Captioning	        & 2020 & 13	& 8     \\
    \VATEX{}, Chinese Captioning	        & 2020 & 6	& 4     \\
    TRECVID, VTT	                        & 2016 & -	& 11    \\
    TRECVID, VTT	                        & 2017 & -	& 16    \\
    TRECVID, VTT	                        & 2018 & -	& 12    \\
    TRECVID, VTT	                        & 2019 & -	& 10    \\
    TRECVID, VTT	                        & 2020 & -	& 6     \\
    \bottomrule
    \end{tabular}
    \caption{Number of annual participants reported for each competition task. Some contests report public and blind test set's participation, but the majority only report the blind test set's participation.
    The blind test track is not a viable evaluation for retrieval tasks.}
    \label{tab:challenge:participants}
\end{table}

\subsection{The Large Scale Movie Description Challenge (LSMDC)} \label{sec:challenge:lsmdc}

The LSMDC competition and workshop have been organized in conjunction with ECCV'16 and ICCV'15, '17\savefn{note:lsmdc}{LSMDC 2017 challenge website: \url{https://sites.google.com/site/describingmovies/lsmdc-2017}}, and '19\footnote{LSMDC 2019 challenge website: \url{https://sites.google.com/site/describingmovies/lsmdc-2019}}.
The competition is based on a unified version of the MPII-MD~\citep{Rohrbach2015ADescription} (see Section~\ref{sec:dataset:mpii-md}) and M-VAD~\citep{Torabi2015UsingResearch} (see Section~\ref{sec:dataset:mvad}) movie datasets. 
This combined dataset is considered one of the most challenging datasets due to the low performance achieved for all metrics. 
These datasets use Audio Descriptions (AD) / Descriptive Video Service (DVS) resources for the visually impaired, transcribed and aligned with the video.
The original task implies the generation of single-sentence descriptions for each movie fragment. 
The challenge consisted of two phases (test sets): public evaluation and blind (without ground-truth descriptions) evaluation.

In the first edition\footnote{ICCV Workshop for LSMDC 2015 challenge website: \url{https://sites.google.com/site/describingmovies/workshop-at-iccv-15}}, hosted on ICCV'15, the winners were decided based on the human evaluation only.
Additionally, the challenge included an automatic evaluation server based on MS COCO Caption Evaluation API\footnote{MS COCO Caption Evaluation: \url{https://github.com/tylin/coco-caption}}.
For human evaluation, each evaluator was asked to rank four generated sentences and a reference sentence from 1 (lower) to 5 (better) concerning four criteria: \emph{grammar}, \emph{correctness}, \emph{relevance}, and \emph{helpful for blind}.

Two other tasks were assessed in the posterior challenge's editions (2016 and 2017): (1) movie annotation and retrieval, and (2) movie fill-in-the-blank.
In the last edition, hosted on ICCV'19, organizers presented a new competition version focused on \textit{Multi-sentence Movie Description Generation}.
For this new challenge, it is essential to determine ``who is who'' and identify characters to provide a coherent and informative narrative.
The tasks assessed in this edition were:
\begin{itemize}
    \item \emph{Multi-Sentence Description}, focusing on the description of videos, and putting generic ``SOMEONE''-s in place of all the occurring character names.
    \item \emph{Fill-in the Characters}, focusing on \textit{filling-in} the character IDs locally (within a set of five clips).
    \item \emph{Multi-Sentence Description with Characters}, which combines both description generation and \textit{filling-in} the local character IDs.
\end{itemize}

\subsection{Video to Language Challenge}\label{sec:challenge:vlc}
The Video to Language Challenge is one of the scenarios of the Microsoft Multimedia Challenge\footnote{Microsoft Multimedia Challenge website: \url{http://ms-multimedia-challenge.com/2017/challenge}}. 
Only two editions have been held until this moment, in 2016 and 2017. 
This challenge is based on the MSR-VTT dataset, but the participants can use other public or private datasets (images or videos) to train their models.
For the evaluation, a complete and natural sentence describing the content of each test set's video had to be submitted.
The performance was evaluated against sentences previously generated by a human during the evaluation stage.

The 2017 edition (hosted on ACMMM'17) was based on both automatic evaluation and human evaluation.
Specifically, two ranking lists of teams were produced by combining their scores on each evaluation metric.
For the automatic evaluation rank, the organizers proposed to combine its ranking positions in four ranking lists following this equation: 
\begin{equation}
    R = R_{\text{BLEU}-4} + R_{\text{METEOR}} + R_{\text{ROUGE}_L} + R_{\text{CIDEr}},
\end{equation}
where $R_{\text{BLEU}-4}$, $R_{\text{METEOR}}$, $R_{\text{ROUGE}_L}$, $R_{\text{CIDEr}}$ are the rank positions of the team for each metric. 
The smaller the final rank, the better performance.

Simultaneously, the human evaluation rank is based on three scores (on the scale of 1-5): \emph{coherence} $S_{c}$, \emph{relevance} $S_{r}$, and \emph{helpful for blind} $S_{hb}$.
The final human evaluation is given by:
\begin{equation}
    S = S_{c} + S_{r} + S_{hb},
\end{equation}
where the higher the final score, the better performance.

Then the final rank is divided into two lists, one in terms of R and one in terms of S.

\subsection{TREC Video Retrieval Evaluation (TRECVID)} \label{sec:challenge:trecvid}
The National Institute of Standards and Technology (NIST) and other US-\-gov-ern\-ment agencies sponsor the TREC conference series.
Over almost twenty years, the TRECVID conference of TREC has pushed the progress in content-based digital video exploitation by using open evaluation metrics.
This effort has allowed a better understanding of how the methods can effectively process videos and evaluate their performance.
Specifically, some of the tasks that have been part of TRECVID are: 
\begin{itemize}
    \item \textbf{Ad-hoc Video Search (AVS)}, for the video retrieval from a text query
    \item \textbf{Instance Search (INS)}, for retrieving specific instances of individual objects, persons, and locations
    \item \textbf{Video to Text Description (VTT)}
\end{itemize}

The VTT task was incorporated, as a pilot-task, in the TRECVID'16 conference~\citep{Awad2016TRECVIDHyperlinking}.
That year, given a development set of 2,000 Twitter Vine videos of Twitter and two sets of descriptions, the goal was to train models for resolving the following subtasks: 
\begin{itemize}
    \item \textbf{Matching \& Ranking:} Requires, for each video, submitting a ranked list of the most likely text descriptions for each set of descriptions.
    \item \textbf{Description Generation:} Requires submitting a generated text description for each video.
\end{itemize}

After that, the number of videos increased (7,485 videos in 2020), and new sources were used.
In Section~\ref{sec:dataset:trecvid-vtt}, we analyze how these datasets have been constructed.
Likewise, in the 2020 edition, five automatic evaluation metrics, \ie BLEU, METEOR, CIDEr, SPICE, and STS, in addition to human evaluation, were used for the Description Generation subtask. 

For the VTT task of TRECVID'21, a new multi-modality subtask, called \emph{Fill-in the Blanks}, has been announced.
For this subtask, the most appropriate word or words to fill in the blank and complete each video's sentence must be submitted.
The blank will represent a single concept but not necessarily a single word.
The scoring will be done using manual evaluation only.
Assessors will view the video and its associated sentence with the system-generated word to determine how well it fills in the blank.

\subsection{ActivityNet}\label{sec:challenge:activitynet}
A challenge with more annual participants than TRECVID is the International Challenge on Activity Recognition, also called ActivityNet Large-Scale Activity Recognition Challenge (before 2018), a CVPR Workshop from 2016.
It centers on recognizing daily life and detecting and captioning multiple events in a video.
The challenge has hosted six diverse tasks in these editions, aiming to drive semantic video understanding limits and link the visual content with human captions.
The organizers based three of the six tasks on the challenging ActivityNet Captions dataset~\citep{Krishna2017Dense-CaptioningVideos}, described in Section~\ref{sec:dataset:activitynet}.
These tasks focus on trace evidence of activities in time in the form of proposals, class labels, and captions. 
Specifically, its task, \emph{Dense-Captioning Events in Videos}\savefn{note:actnettask}{Dense-Captioning Events in Videos task of ActivityNet 2019 challenge website: \url{http://activity-net.org/challenges/2019/tasks/anet_captioning.html}}, involves both detecting and describing events in a video. 
The leader-boards of 2018\savefn{note:actnet18}{Captioning tab in ActivityNet 2018 evaluation website: \url{http://activity-net.org/challenges/2018/evaluation.html}}, 2019\savefn{note:actnet19}{Captioning tab in ActivityNet 2019 evaluation website: \url{http://activity-net.org/challenges/2019/evaluation.html}}, and 2020 have been obtained by server evaluation using the Avg. METEOR metric (see Section \ref{sec:metrics-dg}).

\subsection{VATEX Video Captioning Challenge}\label{sec:challenge:vatex}

The recent VATEX Video Captioning Challenge\footnote{VATEX Captioning Challenge website: \url{https://eric-xw.github.io/vatex-website/index.html}} aims to benchmark progress towards models that can describe the videos in various languages such as English and Chinese.
This challenge is based on the dataset with the same name, described in Section~\ref{sec:dataset:vatex}.
The first edition of the challenge was hosted at an ICCV'19 Workshop and the second edition at a CVPR'20 Workshop.
In 2020, the challenge had two tracks: \emph{Video Captioning} and \emph{Video-guided Machine Translation}.
For the Video Captioning track, the teams should submit generated captions for English or Chinese languages.
While, for the Video-guided Machine Translation track, the participants should submit English-to-Chinese translations using video information as the additional spatiotemporal context.

A drawback of this challenge competition is that for evaluating submissions, human evaluation is not performed.
The scoring is obtained by automatic evaluation only, computing for both tracks the $\text{BLEU-1}$, $\text{BLEU-2}$, $\text{BLEU-3}$, $\text{BLEU-4}$, $\text{METEOR}$, $\text{ROUGE-L}$, and $\text{CIDEr-D}$ metrics.
\section{Datasets for Video-to-Text}\label{sec:datasets}

As mentioned in the previous sections, VTT methods require datasets to train and validate the models.
In this sense, several datasets have been recently created to supports the training of neural VTT models.
They vary in terms of the number of reference descriptions for each video, the length of these references (number of words and sentences), domain specificity, and other aspects discussed in this section.

\begin{figure}[t!]
    \centering
    \includegraphics[width=.45\textwidth]{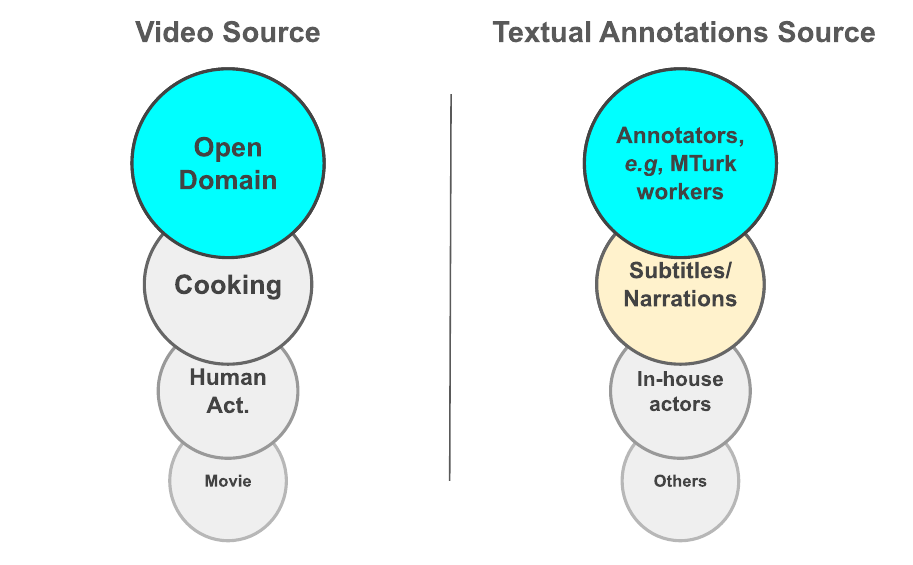}
    \caption{For training VTT models, there are more than twenty-five annotated datasets that we can group according to the video source and the different ways descriptions are obtained (textual annotation source).
    Larger circles represent categories with a more significant number of datasets.
    In blue, the most reported groups, and in wheat color, the group with the most extensive datasets.}
    \label{fig:datasets-groups}
\end{figure}

For training the VTT models, we have found more than twenty-five annotated datasets that we can group according to the video domain and different ways the descriptions are obtained (see Figure~\ref{fig:datasets-groups}).

On the one hand, according to the video domain, we basically have four types of video datasets:
\begin{itemize}
    \item \emph{Open domain}, with videos randomly extracted from online video platforms like \textit{YouTube}
    \item \emph{Cooking videos}, with videos related to cooking activities only
    \item \emph{Human Activities videos}, with videos that show people performing specific actions, like \textit{instructional} videos
    \item \emph{Movie videos}, for videos extracted from films
\end{itemize}

On the other hand, according to the source of the textual annotations, a big group of datasets utilizes a manual annotation process usually performed on the Amazon Mechanical Turk service.
This service is based on crowd workers who watch the video and generate a textual description of them.
However, the biggest dataset (HowTo100M~\cite{Miech2019HowTo100M:Clips}) has been constructed using automatic subtitles or narrations due to the low cost of this technique.

As shown in Section~\ref{sec:results}, extensive results have been reported on datasets with videos in the open domain and human annotations.
Specifically, the most-reported dataset in the literature is MSVD, followed by MSR-VTT, MPII-MD, M-VAD, and ActivityNet Captions.
The popularity of MSVD and MSR-VTT is because they are the first to use open-domain videos and many associated captions with each video (more than 40 and 20, respectively).
While MPII-MD, M-VAD, ActivityNet Captions, and TRECVID VTT datasets are popular for their domain specificity and their uses in respective competitions and Workshops (see Section~\ref{sec:competitions}).

\citet{Aafaq2019VideoMetrics} included a comparative table between almost used video-caption/description pairs datasets.
In contrast, in this section, we include other general characteristics in Table~\ref{tab:video-dataset-summarise}.
We detail the corpus's vocabulary composition\footnote{Categorizing and tagging words: \url{http://www.nltk.org/book/ch05.html}}, the average length of captions, and the percent of tokens that appear in the GloVe-6B\footnote{Global Vectors for Word Representation (GloVe) website: \url{https://nlp.stanford.edu/projects/glove/}} dictionary in Table~\ref{tab:dataset-vocab}.
We show the most reported splits sets and the total size of the datasets in Table~\ref{tab:dataset-splits}.
Additionally, all datasets' references and the tables presented in this section are publicly available in our GitHub repository\footnote{Datasets summary available at: \url{https://github.com/jssprz/video_captioning_datasets}}.

We organize the analysis of datasets for VTT in the rest of this section focusing on five VTT facets: 
open video description, movie description, cooking description, activity description, and dense video description.
However, if the reader is interested in datasets with text sequences of a specific length, we suggest following the order shown in Table~\ref{tab:dataset-vocab}. 
In this Table, we sorted the datasets according to the average length of captions.

\begin{table*}[t]
  \fontsize{4.5pt}{4.5pt}\selectfont
  \centering
  \begin{tabular}{lccccc}
    \toprule
    Dataset                                                     & Type                   & Caps. Source          & Localization  & Audio         & Long-Caps. \\
    \midrule        
    MSVD \citep{L.Chen2011CollectingEvaluation}                  & Open (YouTube)        & Annotators (MTurk)    &               &               & \\
    MPII Cooking Act. \citep{Rohrbach2012AActivities}            & Cooking               & in-house actors       &               &               & \\
    MPII Cooking C. Act. \citep{Rohrbach2012ScriptActivities}    & Cooking               & in-house actors       &               &               & \\
    YouCook \citep{Das2013AStitching}                            & Cooking               & Annotators (MTurk)    &               &               & \\
    TACoS \citep{Regneri2013GroundingVideos}                     & Cooking               & Annotators (MTurk)    & \checkmark    &               & \\
    TACoS-Multilevel \citep{Rohrbach2014CoherentDetail}          & Cooking               & Annotators (MTurk)    &               &               & \\
    M-VAD \citep{Torabi2015UsingResearch}                        & Movie                 & DVS                   &               & \checkmark    & \checkmark \\
    MPII-MD \citep{Rohrbach2015ADescription}                     & Movie                 & Script + DVS          &               & \checkmark    & \\
    LSDMC\footref{note:lsmdc}                                    & Movie                 & Script + DVS          &               & \checkmark    & \\
    MSR-VTT \citep{Xu2016MSR-VTT:Language}                       & Open                  & Annotators (MTurk)    &               & \checkmark    &\\
    MPII Cooking 2 \citep{Rohrbach2016RecognizingData}           & Cooking               & in-house actors       & \checkmark    &               &\\
    Charades \citep{Sigurdsson2016HollywoodUnderstanding}        & Daily indoor act.     & Annotators (MTurk)    & \checkmark    & \checkmark    & \checkmark \\
    VTW-full \citep{Zeng2016TitleVideos}                         & Open (YouTube)        & Owner/Editor          &               &               &\\
    TGIF \citep{Li2016TGIF:Description}                          & Open                  & crowdworkers          &               &               & \checkmark \\
    TRECVID-VTT'16 \citep{Awad2016TRECVIDHyperlinking}           & Open                  & Annotators            &               & \checkmark    & \checkmark \\
    Co-ref+Gender \citep{Rohrbach2017GeneratingPeople}           & Movie                 & DVS                   &               & \checkmark    & \\
    ActivityNet Caps. \citep{Krishna2017Dense-CaptioningVideos}  & Human act.            & Annotators (MTurk)    & \checkmark    & \checkmark    & \checkmark \\
    TRECVID-VTT'17 \citep{Awad2017TrecvidHyperlinking}           & Open (Twitter)        & Annotators            &               & \checkmark    &\\
    YouCook2 \citep{Zhou2018TowardsVideos}                       & Cooking               & viewer/annotator      & \checkmark    & \checkmark    &\\
    Charades-Ego \citep{Sigurdsson2018ActorVideos}               & Daily indoor act.     & Annotators (MTurk)    & \checkmark    & \checkmark    & \checkmark \\
    20BN-s-s V2 \citep{Mahdisoltani2018Fine-grainedCaptioning}   & Human-obj. interact.  & Annotators (MTurk)    &               &               &\\
    TRECVID-VTT'18 \citep{Awad2018TRECVIDSearch}                 & Open (Twitter)        & Annotators            &               & \checkmark    & \checkmark \\
    TRECVID-VTT'19 \citep{Awad2019TRECVIDRetrieval}              & Open (Twitter+Flirck) & Annotators            &               & \checkmark    & \checkmark \\
    \VATEX\footref{note:vatex} \citep{Wang2019VaTeXResearch}     & Open                  & Annotators (MTurk)    &               & \checkmark    & \checkmark \\
    HowTo100M \citep{Miech2019HowTo100M:Clips}                   & Instructional (YouTube)& Subtitles            & \checkmark    & \checkmark    & \checkmark \\
    TRECVID-VTT'20 \citep{Awad2020TRECVIDDomains}                & Open (Twitter+Flirck) & Annotators            &               & \checkmark    & \checkmark \\
    \bottomrule
  \end{tabular}
  \caption{
  Datasets of video-caption/description pairs for training video-to-text models.
  The cells of the last three columns are marked: (1)~if the dataset contains temporal-localization information related to each description, (2)~if the videos of the dataset contain audio, and (3)~if corpus' descriptions are more than eleven words long on average.}
  \label{tab:video-dataset-summarise}
\end{table*}

\begin{table}[t]
\tiny
\centering
  \begin{tabular}{lccccccc}
    \toprule
    Corpus                                                              & Avg.          & \multicolumn{5}{c}{Vocabulary}                         & \% of tokens in\\
                                                                        & cap. len.  & Tokens        & Nouns         & Verbs         & Adjectives    & Adverbs    & GloVe-6B \\
    \midrule
    HowTo100M \citep{Miech2019HowTo100M:Clips}                          & 4.16          & \bf 593,238   & \bf 491,488   & \bf 198,859   & \bf 219,719   & \bf 76,535 & 36.64    \\
    VTW-full \citep{Zeng2016TitleVideos}                                & 6.40          & 23,059        & 13,606        & 6,223         & 3,967         & 846        & -        \\
    20BN-s-s V2 \citep{Mahdisoltani2018Fine-grainedCaptioning}          & 6.76          & 7,433         & 6,087         & 1,874         & 1,889         & 361        & 74.51    \\
    MSVD \citep{L.Chen2011CollectingEvaluation}                         & 7.14          & 9,629         & 6,057         & 3,211         & 1,751         & 378        & 83.44    \\
    TACoS-Multilevel \citep{Rohrbach2014CoherentDetail}                 & 8.27          & 2,863         & 1,475         & 1,178         & 609           & 207        & 91.86    \\
    MSR-VTT, 2016\footref{note:msrvtt16}\citep{Xu2016MSR-VTT:Language}  & 9.27          & 23,527        & 19,703        & 8,862         & 7,329         & 1,195      & 80.74    \\
    LSDMC\footref{note:lsmdc}                                           & 10.66         & 24,267        & 15,095        & 7,726         & 7,078         & 1,545      & 88.57    \\
    MPII-MD \citep{Rohrbach2015ADescription}                            & 11.05         & 20,650        & 11,397        & 6,100         & 3,952         & 1,162      & 88.96    \\
    TGIF \citep{Li2016TGIF:Description}                                 & 11.28         & 10,646        & 6,658         & 3,993         & 2,496         & 626        & \bf 97.85    \\
    M-VAD \citep{Torabi2015UsingResearch}                               & 12.44         & 17,609        & 9,512         & 2,571         & 3,560         & 857        & \ul{94.99}    \\
    ActivityNet Caps. \citep{Krishna2017Dense-CaptioningVideos}         & 14.72         & 10,162        & 4,671         & 3,748         & 2,131         & 493        & 94.00    \\
    \VATEX-en\footref{note:vatex} \citep{Wang2019VaTeXResearch}         & 15.25         & \ul{28,634}   & \ul{23,288}   & \ul{12,796}   & \ul{10,639}   & \ul{1,924} & 76.84    \\
    TRECVID-VTT'20 \citep{Awad2020TRECVIDDomains}                       & 18.90         & 11,634        & 7,316         & 4,038         & 2,878         & 542        & 93.36    \\
    Charades \citep{Sigurdsson2016HollywoodUnderstanding}               & \ul{23.91}    & 4,010         & 2,199         & 1,780         & 651           & 265        & 90.00    \\
    Charades-Ego \citep{Sigurdsson2018ActorVideos}                      & \bf 26.30     & 2,681         & 1,460         & 1,179         & 358           & 190        & 91.12    \\
    \bottomrule
  \end{tabular}
  \caption{Corpus details.
  Average length of captions, vocabulary (unique words) composition of each dataset's corpus, and percent of tokens that appear in the GloVe-6B dictionary.
  The words categorization have been calculated by POS-tagging with \textit{universal tagset} mapping.
  We sorted the corpus according to the average length of captions.}\label{tab:dataset-vocab}
\end{table}

\begin{table*}[t]
\fontsize{4.7pt}{4.7pt}\selectfont
\centering
\begin{tabular}{lccccccccc}
\toprule
Dataset                                         & \multicolumn{2}{c}{Train set} & \multicolumn{2}{c}{Validation set} & \multicolumn{2}{c}{Testing set}  & \multicolumn{2}{c}{Total}\\
                                                                        & clips     & captions  & clips     & captions  & clips     & captions          & clips             & captions \\
\midrule            
MSVD \citep{L.Chen2011CollectingEvaluation}                             & 1,200     & 48,779    & 100       & 4,291     & 670       & 27,768            & 1,970             & 80,838\\
TACoS \citep{Regneri2013GroundingVideos}                                & -         & -         & -         & -         & -         & -                 & 7,206             & 18,227\\
Charades-Ego \citep{Sigurdsson2018ActorVideos}                          & 6,167     & 12,346    & 1,693     & 1,693     & -         & -                 & 7,860             & 14,039\\
TRECVID-VTT'20 \citep{Awad2020TRECVIDDomains}                           & 7,485     & 28,183    & -         & -         & 1,700     & -                 & 9,185             & 28,183\\
Charades \citep{Sigurdsson2016HollywoodUnderstanding}                   & 7,985     & 18,167    & 1,863     & 6,865     & -         & -                 & 9,848             & 25,032\\
MSR-VTT, 2017\footref{note:msrvtt17}                                    & 10,000    & 200,000   & -         & -         & 3,000     & 60,000            & 13,000            & 260,000\\
TACoS-Multilevel \citep{Rohrbach2014CoherentDetail}                     & -         & -         & -         & -         & -         & -                 & 14,105            & 52,593\\
ActivityNet Caps. \citep{Krishna2017Dense-CaptioningVideos}             & 10,024    & 36,587    & 4,926     & 17,979    & 5,044     & 18,410            & 19,994            & 72,976\\
\VATEX\footref{note:vatex} \citep{Wang2019VaTeXResearch}                & 25,991    & 259,910   & 3,000     & 30,000    & 6,000     & 60,000            & 34,991            & 349,910\\
M-VAD \citep{Torabi2015UsingResearch}                                   & 36,921    & 36,921    & 4,717     & 4,717     & 4,951     & 4,951             & 46,589            & 46,589\\
MPII-MD \citep{Rohrbach2015ADescription, Rohrbach2017GeneratingPeople}  & 56,822    & 56,861    & 4,927     & 4,930     & 6,578     & 6,584             & 68,327            & 68,375\\
LSDMC\footref{note:lsmdc}                                               & 91,908    & 91,941    & 6,542     & 6,542     & 10,053    & 10,053            & 108,503           & 108,536\\
TGIF \citep{Li2016TGIF:Description}                                     & 80,850    & 80850     & 10,831    & 10,831    & 34,101    & 34,101            & 125,782           & 125,781\\
20BN-s-s V2 \citep{Mahdisoltani2018Fine-grainedCaptioning}              & 168,913   & 1,689,913 & 24,777    & 24,777    & 27,157    & -                 & 220,847           & \ul{1,714,690}\\
MSR-VTT, 2016\footref{note:msrvtt16}                                    & 6,512     & 130,260   & 498       & 9,940     & 2,990     & 59,800            & \ul{507,502}      & 200,000\\
HowTo100M \citep{Miech2019HowTo100M:Clips}                              & -         & -         & -         & -         & -         & -                 & \bf 139,668,840   & \bf 139,668,840\\
\bottomrule
\end{tabular}
\caption{Number of video clips and captions in the standard splits of each video-caption/description pairs dataset.
We sorted the datasets according to the total number of clips.}
\label{tab:dataset-splits}
\end{table*}



\subsection{Datasets for Open Video Description}


\subsubsection{Microsoft Video Description Corpus (MSVD)}
This dataset, also called \textit{YoutubeVideos2Text}, was proposed by \citet{L.Chen2011CollectingEvaluation}, considering \textit{YouTube} video clips of short duration (from four to ten seconds) with a single event.
Several crowd workers of the Amazon Mechanical Turk\footnote{MTurk is an online framework that allows researchers to post annotation tasks, called HITs (Human Intelligence Task)} (AMT) service \citep{Rashtchian2010CollectingTurk} annotated each video clip.
Each annotator wrote a sentence describing the event or the video's main action in a language of their choice.
Initially, 2,089 clips and 85,550 descriptions were compiled in English, with 41 descriptions for each video on average.
However, due to the dynamism of \textit{YouTube}, the current public data consists of 1,970 clips from 1,464 different videos.
There are at least 27 annotations for more than 95\% of the videos, with an average number of words of around 7.1, which indicates those descriptions are relatively short and simple sentences.


\subsubsection{Microsoft Research-Video to Text (MSR-VTT) or MSR-VTT-10K}\label{sec:dataset:msr-vtt}
Proposed by \citet{Xu2016MSR-VTT:Language}, it collects videos from a commercial video search engine.
Different AMT workers produced 20 human-written reference captions for each video.
The videos are about general topics in our life. They are grouped into 20 categories (\eg music, people, TV show, and movie).
In total MSR-VTT contains 10,000 clips from 7,180 different videos and 200,000 sentences (20 per clip).
The duration of each video clip is from 10 to 30 seconds and contains audio information as well.
The average number of words per description is around 9.3, indicating more complicated sentences than other datasets like MSVD.
Moreover, this was the dataset for the first Video to Language Challenge\savefn{note:msrvtt16}{MSR-VTT-10K dataset website: \url{http://ms-multimedia-challenge.com/2016/dataset}}, as we mentioned in Section~\ref{sec:challenge:vlc}.

In the second MSR Video to Language Challenge in 2017\savefn{note:msrvtt17}{MSR-VTT 2017 dataset website: \url{http://ms-multimedia-challenge.com/2017/dataset}}, the organizers used a combination of the train, validation, and test sets as the new training data.
Moreover, they released an additional test set of 3,000 video clips.
Specifically, this dataset version contains about 50 hours and 260,000 clip-sentence pairs in total, covering the most comprehensive categories and diverse visual content.



\subsubsection{Video Titles in the Wild (VTW)}
\citet{Zeng2016TitleVideos} proposed this dataset for video title generation.
The descriptions of this dataset make up a very diverse vocabulary.
The data includes 18,100 video clips with one sentence per clip.
On average, each word appears only in two sentences, and each clip has a duration of 1.5 minutes.
Besides, the authors provided a sentence-only example per video.
Attending to the diverse vocabulary (see Table \ref{tab:dataset-vocab}), its sophisticated sentences were produced by editors instead of simple sentences produced by Turkers.
This vocabulary characteristic allows the dataset to be used for language-level understanding tasks, including video question answering~\citep{Aafaq2019VideoMetrics}.



\subsubsection{Tumblr GIF (TGIF)}
This dataset, proposed by \citet{Li2016TGIF:Description}, comprises Tumblr's\footnote{Tumblr website: \url{https://www.tumblr.com}} one-year GIF publications.
The interest in creating a dataset with GIF was because very few academic works dealt with them.
GiFs have a short duration and tell a visual story without audio.
To obtain fluent descriptions, 931 workers participated in the GIFs annotation, preferably speaking English as their native language.
Each worker described at least 800 GIFs, guaranteeing the descriptions' diversity.
TRECVID challenge participants have widely explored this dataset's use as an additional data source for training models, improving their methods' performance.


\subsubsection{\VATEX}\label{sec:dataset:vatex}
More recently, \citet{Wang2019VaTeXResearch} proposed the \VATEX\savefn{note:vatex}{\VATEX{} dataset website: \url{https://eric-xw.github.io/vatex-website/index.html}} dataset, a multilingual video captioning dataset.
This dataset covers 600 human activities and a variety of video domains, reusing a subset of the videos from the Kinetics-600\footnote{Kinetics dataset website: \url{https://deepmind.com/research/open-source/kinetics}} dataset \citep{Carreira2017QuoDataset}.
Twenty individual human annotators (AMT workers) paired each video with ten English and ten Chinese captions.
Each caption is from an individual worker and should describe all the important characters and actions shown in the video clips with ten or more words.
The dataset comprises over 41,250 unique videos and 825,000 unique captions (every caption is unique in the whole corpus).
Finally, in Table~\ref{tab:dataset-vocab}, we show the linguistic complexity (vocabulary distribution) of this dataset into the English (\VATEX-en) and the Chinese (\VATEX-zh) corpus, which evidences that \VATEX{} has greater difficulty than other datasets like MSR-VTT.


\subsubsection{TRECVID-VTT}\label{sec:dataset:trecvid-vtt}
Since 2016, the TRECVID organizers have proposed a new version of the development and test sets of the VTT task.
For all these versions, a rigorous and monitored annotation process has been performed. 
The annotators are asked to combine in one sentence four facets if applicable~\citep{Awad2016TRECVIDHyperlinking, Awad2017TrecvidHyperlinking, Awad2018TRECVIDSearch, Awad2019TRECVIDRetrieval, Awad2020TRECVIDDomains}:
\begin{itemize}
    \item Who is the video describing (such as objects, persons, animals)?
    \item What are the objects and beings doing (such as actions, states, events)?
    \item Where (such as locale, site, place, geographic)?
    \item When (such as time of day, season)?
\end{itemize}

To answer these questions and produce the video descriptions, the annotators receive training for the task.
Their work is monitored, and feedback is provided.
The NIST personnel is available for any questions or confusion during the process.
This annotation process differentiates the TRECVID datasets from other datasets, which produce arguably better/more detailed descriptions than crowd-sourced datasets.

Furthermore, for each of the videos, the annotators rated how difficult it was to describe them.
The composition of each one of these datasets is as follows:
\begin{itemize}
    \item \textit{TRECVID-VTT'16}~\citep{Awad2016TRECVIDHyperlinking} consists of 2.000 Twitter Vine videos randomly selected.
    Each one with a duration of approximately six seconds. Eight annotators made the textual descriptions.
    For each video, two annotations were made by two different annotators, for a total of 4.000 textual descriptions, divided into two sets.
    
    \item \textit{TRECVID-VTT'17}~\citep{Awad2017TrecvidHyperlinking} is composed of 1880 Twitter Vine videos.
    The videos were divided amongst ten annotators, with each video being annotated by at least two and at most five.
    
    \item \textit{TRECVID-VTT'18}~\citep{Awad2018TRECVIDSearch} is composed of 1903 Vine videos manually annotated by exactly five assessors.
    Each assessor rated how likely is it that other assessors will write similar descriptions into not likely, somewhat likely, and very likely.
    Additionally, the task organizers manually removed videos with unrelated segments that are hard to describe, animated videos, and unappropriated and offensive videos.
    
    \item \textit{TRECVID-VTT'19}~\citep{Awad2019TRECVIDRetrieval}, in contrast to the previous versions, was enriched by using two video sources.
    Specifically, \textit{TRECVID-VTT'19} contains 1,044 Vine videos used in the challenge since 2016 and 1,010 video clips extracted from 91 Flickr videos.
    The 2,054 videos were annotated following the same criteria as in 2018.
    
    \item \textit{TRECVID-VTT'20}~\citep{Awad2020TRECVIDDomains} contains short videos (ranging from 3 seconds to 10 seconds) from the 2016 to 2019 versions. There are 9185 videos with captions. Each video has between 2 and 5 captions written by dedicated annotators.
    The video sources have the following distribution: 6,475 from Twitter Vine, 1,010 from Flickr, and 1,700 from the V3C dataset.
\end{itemize}

\subsection{Datasets for Movie Description}

\subsubsection{Montreal Video Annotation Dataset (M-VAD)}\label{sec:dataset:mvad}
\citet{Torabi2015UsingResearch}, researchers of the Mila group of the University of Montreal, proposed this dataset to address the movie description task.
M-VAD is a large-scale dataset based on Descriptive Video Service\footnote{Descriptive Video Service (DVS) is a major specialized in audio description, which aims to describe the visual content in form of narration. These narrations are commonly placed during natural pauses in the original audio of the video, and sometimes during dialogues} (DVS).
The authors eliminated the sentences obtained from the first narrated three minutes and the four final minutes of the film because the DVS's narrator at that moment could read, for example, the credits of the film. With 46,523 video clips in total, M-VAD consists of 84.6 hours of videos from 92 Hollywood films.
Each video clip was annotated automatically with a single narrative, using the vocabulary detailed in Table~\ref{tab:dataset-vocab}.
The usual split consists of 36,921 train clips, 4,717 validation clips, and 4,951 test clips.


\subsubsection{Max Planck Institute for Informatics-Movie Description (MPII-MD)}\label{sec:dataset:mpii-md}
\citet{Rohrbach2015ADescription} proposed this dataset by including transcribed ADs (audio descriptions for the blind) with temporal alignment to full-length 55 HD movies.
For this, the authors proposed a semiautomatic approach of two phases to obtain ADs' transcriptions and make them correspond to the video.
They selected 39 movie scripts for the first phase and filtered the sentences with a reliable alignment score (the ratio of matched words in the near-by monologues) of at least 0.5.
The obtained sentences were then manually aligned to video, filtering all those sentences that did not correspond to the film's dialogues, such as the cast's sentences.
This approach's importance is that, until that moment, there were not many available transcriptions from ADs.
In total, the dataset\footnote{MPII-MD dataset website: \href{https://www.mpi-inf.mpg.de/departments/computer-vision-and-multimodal-computing/research/vision-and-language/mpii-movie-description-dataset/}{www.mpi-inf.mpg.de}} contains a smaller number of video clips (from 94 unique movies) than the sentences, 68,337 and 68,375, respectively, because, in some cases, a video clip is annotated with more than one sentence.
The vocabulary distribution of this corpus is shown in Table~\ref{tab:dataset-vocab}.


\subsubsection{LSMDC}
This dataset is considered one of the most challenging datasets due, to some extent, to the low performance and reported metrics scores of several models.
The dataset is a unified version of the M-VAD~\citep{Torabi2015UsingResearch} and MPII-MD~\citep{Rohrbach2015ADescription} datasets.
These datasets use Audio Descriptions (AD) / Descriptive Video Service (DVS) resources for the visually impaired, transcribed and aligned with the video.
Moreover, this is the competition dataset with the same name, as described in Section~\ref{sec:challenge:lsmdc}.


\subsubsection{MPII-MD Co-ref+Gender}
\citet{Rohrbach2017GeneratingPeople} introduced the dataset for Grounded and Co-Referenced Characters (MPII-MD Co-ref+Gender) dataset, including annotations on language and visual\footnote{MPII-MD Co-ref+Gender dataset website: \href{https://www.mpi-inf.mpg.de/departments/computer-vision-and-multimodal-computing/research/vision-and-language/grounded-and-co-referenced-characters/}{www.mpi-inf.mpg.de}} side for the goal to learn the visual co-reference resolution on the MPII-MD dataset~\citep{Rohrbach2015ADescription}.
On the language side, the authors included information to know when different mentions refer to the same person.
On the visual side, they relate the names (``aliases'') with the visual appearances.
In total they labeled 45,325 name mentions and 17,839 pronouns (``he'' and ``she'').
The character mentioned in the original MPII-MD descriptions were replaced with ``MaleCoref'' (``FemaleCoref''), otherwise with ``MaleName'' (``FemaleName'').
MPII-MD Co-ref+Gender dataset also includes bounding box annotations for character heads.
It keeps the same split sets of the MPII-MD dataset. 


\subsection{Datasets for Cooking Description}

\subsubsection{MP-II Cooking Activities}
\citet{Rohrbach2012AActivities} proposed this fine-grained cooking activities dataset.
These authors video recorded participants cooking different dishes, such as ``fruit salad'', ``pizza'', or ``cake'', and annotated the videos with activity categories on time intervals.
Moreover, the authors annotated a subset of frames with the human pose.
They recorded 12 participants performing 65 different cooking activities, such as ``cut slices'', ``pour'', or ``spice''.
Consequently, the dataset contains a total of 5,609 annotations\footnote{MP-II Cooking Activities dataset website: \href{https://www.mpi-inf.mpg.de/departments/computer-vision-and-multimodal-computing/research/human-activity-recognition/mpii-cooking-activities-dataset/}{www.mpi-inf.mpg.de}}.
Overall, they recorded 44 videos with a total length of more than 8 hours or 88,1755 frames. 



\subsubsection{MPII Cooking Composite Activities}
In this case, \citet{Rohrbach2012ScriptActivities} focused on composite activities and thus incorporated significantly more dishes/composites, which are slightly shorter and more straightforward than in the previous dataset.
Although, this dataset contains more videos\footnote{MPII Cooking Composite Activities dataset website: \href{https://www.mpi-inf.mpg.de/departments/computer-vision-and-multimodal-computing/research/human-activity-recognition/mpii-cooking-composite-activities/}{www.mpi-inf.mpg.de}} and activities than MP-II Cooking dataset.
In total, they recorded 22 participants in 212 videos and 41 activities.


\subsubsection{YouCook}
\citet{Das2013AStitching} collected the data from 88 videos\footnote{YouCook dataset website: \url{http://web.eecs.umich.edu/~jjcorso/r/youcook/}} downloaded from \textit{YouTube}, roughly uniformly split into six different cooking styles, such as baking and grilling.
The training consists of 49 videos with frame-by-frame object and action annotations (as well as several pre-computed low-level features).
The test set consists of 39 videos.
Each video has several human-provided natural language descriptions.
On average, there are 8 paragraph descriptions per video, each one has 67 words on average, and the average number of words per sentence is 10.


\subsubsection{Saarbrucken Corpus of Textually Annotated Cooking Scenes (TACoS)}
This dataset, proposed by \citet{Regneri2013GroundingVideos}, was constructed on top of the MP-II Cooking Composite Activities dataset~\citep{Rohrbach2012ScriptActivities}.
They extend it with multiple textual descriptions collected by crowd-sourcing via MTurk.
To provide coherent descriptions and to facilitate the alignment of sentences describing activities with their proper video segments, they also provide approximate time-stamps as a textual annotation.
Specifically, the dataset contains 26 fine-grained cooking activities in 127 videos.
It contains 20 different descriptions per video, 17,334 action descriptions with 11,796 sentences, and 146,771 words.



\subsubsection{Textually Annotated Cooking Scenes Multi-Level (TACoS-Multilevel)}
This dataset~\citep{Rohrbach2014CoherentDetail} addressed the challenging task of the coherent multi-sentence video description on the cooking domain.
Based on the TACoS dataset~\citep{Regneri2013GroundingVideos}, they also collected the annotations via MTurk.
For each one of the 127 videos\footnote{TACoS-Multilevel dataset website: \href{https://www.mpi-inf.mpg.de/departments/computer-vision-and-multimodal-computing/research/vision-and-language/tacos-multi-level-corpus/}{www.mpi-inf.mpg.de}} AMT workers wrote the textual descriptions for three levels of details for each video.
For the first level, they wrote a detailed description with a maximum of 15 sentences.
For the second level, they wrote a brief description of three to five sentences.
For the third level, they wrote a single sentence.
In general, each video has 20 triplets of descriptions, of which approximately 2,600 were collected from TACoS dataset videos.

\subsubsection{YouCook2}
\citet{Zhou2018TowardsVideos} proposed the largest instructional dataset27 until now, where all descriptions are imperative English sentences.
Unlike YouCook dataset~\citep{Das2013AStitching}, this dataset contains 2,000 \textit{YouTube} videos from 89 cooking recipes, with 5.26 minutes on average, for a total of 176 hours.
Like the ActivityNet Captions dataset (see Section~\ref{sec:dataset:activitynet}), the videos were manually annotated with \emph{temporal localization} and captions.
They are the two most extensive datasets used to evaluate the models for dense video description task.
On average, YouCook2 has 22 videos per recipe, each video has 7.7 temporal-localized segments, and each description has 8.8 words.


\subsection{Datasets for Activity Description}

\subsubsection{Charades}
\citet{Sigurdsson2016HollywoodUnderstanding} selected 9,848 videos\footnote{Charades dataset website: \url{https://allenai.org/plato/charades/}} of \emph{daily indoor activities} collected through 267 MTurk workers.
Each worker received scripts with the objects and the actions from a fixed vocabulary, and they recorded videos acting out the script.
The dataset contains 66,500 temporal annotations for 157 action classes, 41,104 labels for 46 object classes, and 27,847 textual descriptions of the videos.

\subsubsection{ActivityNet Captions}\label{sec:dataset:activitynet}
\citet{Krishna2017Dense-CaptioningVideos} proposed this dataset, which includes \emph{temporally annotated} sentence descriptions. Videos cover a wide range of complex human activities that interest people in their daily lives.
In the current version of the dataset\footnote{ActivityNet Captions dataset website: \url{https://cs.stanford.edu/people/ranjaykrishna/densevid/}}, each sentence covers a unique segment of the video and describes multiple events that can co-occur (overlap).
On average, each of the 20,000 videos (849 hours in total) contains 3.65 temporally localized sentences, with 15.25 words describing 36 seconds and 31\% of its respective video.
The ActivityNet Challenge uses this dataset for addressing the scenario where multiple events co-occur.
Section~\ref{sec:challenge:activitynet} describes this competition.

\subsubsection{Charades-Ego}
More recently, \citet{Sigurdsson2018ActorVideos}, based on scripts from the Charades dataset \citep{Sigurdsson2016HollywoodUnderstanding}, proposed a new dataset for addressing one of the biggest bottlenecks facing egocentric vision research, providing a link from first-person to abundant third-person data on the web~\citep{Sigurdsson2018ActorVideos}.
This dataset has 112 actors performing 157 different actions, with semantically paired first and third-person videos.
Although these two datasets were not proposed for the VTT problem, the human-written scripts and descriptions included for each video allow their use as additional data to train and validate VTT models~\citep{Hu2019HierarchicalCaptioning}.


\subsubsection{20BN-something-something Dataset V2}
\citet{Mahdisoltani2018Fine-grainedCaptioning} proposed this dataset as the second release of the 20BN-something-something dataset, which, unlike the first release, considers captioning in addition to classification.
Many crowd workers created the dataset.
This version has 220,847 videos\footnote{20BN-something-something dataset website: \url{https://20bn.com/datasets/something-something}} (vs. 108,499 in V1). In total, there are 318,572 annotations involving 30,408 unique objects.
The descriptions compose 174 template labels replacing each object mention (nouns) for the ``SOMETHING'' word.


\subsubsection{HowTo100M}
Recently, \citet{Miech2019HowTo100M:Clips} proposed the large-scale HowTo100M dataset, which contains 1.2 million instructional \textit{YouTube} videos from 12 categories such as home-and-garden, computers-and-electronics, and food-and-entertaining.
The dataset was constructed by querying YouTube for \emph{everyday activities} from \textit{WikiHow}\footnote{WikiHow is an online resource that contains 120,000 articles on \textit{How to \ldots}
for a variety of domains ranging from cooking to human relationships structured in a hierarchy: \url{https://www.wikihow.com/}} and filtering to videos that contained English subtitles, appeared in the top 200 search results, had more than 100 views, and was less than 2,000 seconds long.
Each video in the dataset is split into clips according to the video's YouTube subtitles' time intervals.
This process results in 136 million video clips.
For corpus construction, narrated captions have been processed.
Therefore, the authors removed many stop words that are not relevant for learning the text-video joint embedding~\citep{Miech2019HowTo100M:Clips}.
On average, each video produces 110 clip-caption pairs, with an average duration of 4 seconds per clip and 4.16 words (after excluding stop-words) per caption. 
Additionally, the highest and second-highest level task category from \textit{WikiHow} was included.

The major drawback of this corpus, and generally of all corpus constructed from automatic subtitles, is the large number of unknown tokens that occur.
In the HowTo100M's corpus, only 36.64\% of words in the vocabulary (217,361 of the 593,238 unique words) appear in the widely used GloVe-6B dictionary, which has 400,000 tokens.

\section{State-of-the-Art Results, an Analysis}\label{sec:results}

In this section, we present a comparison and analysis of the experimental results and the performance of various state-of-the-art techniques as reported.
Although, there is no standard evaluation methodology for evaluating the VTT methods.
A set of automatic evaluation metrics, introduced in Sections~\ref{sec:metrics-dg} and~\ref{sec:metrics-mr}, have been widely reported by researchers at ``convenience.'' 
Generally, experimental results of the works in the literature use the codes released on the Microsoft COCO evaluation server~\citep{Chen2015MicrosoftServer} to enable systematic evaluation and benchmarking, reporting the $\text{BLEU-N}$, $\text{METEOR}$, $\text{ROUGE-L}$, and $\text{CIDEr-D}$ automated metrics.

The most-reported metric in the literature is $\text{METEOR}$ (defined in Equation~\ref{eq:meteor}), followed by $\text{BLEU-4}$ (defined in Equation~\ref{eq:bleu}).
Simultaneously, the two most reported datasets are \textbf{MSVD} and \textbf{MSR-VTT}.
Additionally, unlike other datasets, almost all \textbf{MSR-VTT} experiments report all four metrics: $\text{BLEU-4}$, $\text{METEOR}$, $\text{ROUGE-L}$, and $\text{CIDEr-D}$.
This discrepancy in the number of times each metric is reported does hard to compare the methods' performance and select the best and most generalizable solution.
To deal with this lack, in this section, we rank state-of-the-art methods according to an overall score defined as follows.

Some recent works~\citep{Chen2020ASampling, Chen2020DelvingCaptioning} have proposed overall scores to evaluate the performance of the video captioning models.
Following this idea, we computed the $\text{S}_{\text{overall}}$ score for all state-of-the-art methods as an overall judgment of their automatic evaluation-based performance.
Let $\mathcal{M} = \{\text{BLEU-4},\text{ROUGE-L}, \\\text{METEOR}, \text{CIDEr-D}\}$ the set of metrics we consider for the overall score, and $\mathcal{M}_t \subseteq \mathcal{M}$ the set of metrics that method $t$ reported. 
We define $\text{S}_{\text{overall}}(t)$ as:
\begin{equation}\label{eq:s-overall}
    \text{S}_{\text{overall}}(t) = \sum_{\mu \in \mathcal{M}_t} \alpha_\mu \frac{\mu(t)}{\max_q(\mu(q))},
\end{equation}
\noindent where $\mu(t)$ is the score that method $t$ obtained for metric $\mu$, and $\max_q(\mu(q))$ represents the maximum score obtained by any of the state-of-the-art methods for metric $\mu$. 
We use uniform weights $\alpha_\mu = \frac{1}{|\mathcal{M}_t|}$ for almost all datasets, but they can be changed to pay more attention to some metrics than others.
For example, we can represent the human judgment correlation of each metric in each dataset.
Some competitions, such as TRECVID-VTT, annually report the metrics correlation with human evaluation \citep{Awad2019TRECVIDRetrieval}.

\subsection{State of the Art on Description Generation Methods}

Table~\ref{tab:results:dg:msvd} summarizes the benchmark results of state-of-the-art techniques on the MSVD dataset.
The recent SemSynAN~\citep{Perez-Martin2021ImprovingEmbedding} and VNS-GRU~\citep{Chen2020DelvingCaptioning} methods dominate for all metrics.
These two methods, along with SCN-LSTM + sampling~\citep{Chen2020ASampling} and AVSSN~\citep{Perez-Martin2020AttentiveCaptioning}, surpass the 1.00 score for the CIDEr metric.
These four methods are based on compositional RNN-based decoders that compose intermediate learned representations into the generation process. 
Likewise, SemSynAN~\citep{Perez-Martin2021ImprovingEmbedding} and VNS-GRU~\citep{Chen2020DelvingCaptioning} incorporate variational dropout strategies in the decoder model for regularization.
One of the differences between these two methods is that SemSynAN~\citep{Perez-Martin2021ImprovingEmbedding} pre-train a joint embedding for extracting syntactic representations~\citep{Hou2019JointCaptioning, Wang2019ControllableNetwork} from videos, while VNS-GRU~\citep{Chen2020DelvingCaptioning} train the model by adopting a professional learning strategy for enlarging the generated vocabulary.

\begin{table}[t]
    \centering
    \scriptsize
    \begin{tabular}{lccccc}
    \toprule
    Method                                                                & BLEU-4    & ROUGE-L   & METEOR    & CIDEr-D   & $\text{S}_{\text{overall}}$ \\
    \midrule                    
ReBiLSTM(shortcut) \citep{Bin2019DescribingLSTM}                    & 0.373     & -         & 0.303     & -         & 0.649     \\
BAE \citep{Baraldi2017HierarchicalCaptioning}                       & 0.425     & -         & 0.324     & 0.635     & 0.660     \\
LSTM-GAN+attn. \citep{Yang2018VideoLSTM}                            & 0.429     & -         & 0.304     & -         & 0.693     \\
CAM-RNN \citep{Zhao2019CAM-RNN:Captioning}                          & 0.424     & 0.694     & 0.334     & 0.543     & 0.699     \\
MHB \citep{Nguyen2017MultistreamCaptioning}                         & 0.430     & 0.687     & 0.332     & 0.711     & 0.734     \\
aLSTMs \citep{Gao2017VideoConsistency}                              & 0.508     & -         & 0.333     & 0.748     & 0.743     \\
STAT\_V \citep{Yan2020STAT:Captioning}                              & 0.520     & -         & 0.333     & 0.738     & 0.746     \\
LSTM-TSA$_{IV}$ \citep{Pan2017VideoAttributes}                      & 0.528     & -         & 0.335     & 0.740     & 0.752     \\
TDDF(VGG+C3D) \citep{Zhang2017Task-DrivenDescription}               & 0.458     & 0.697     & 0.333     & 0.730     & 0.753     \\
TSA-ED \citep{Wu2018InterpretableLocalization}                      & 0.517     & -         & 0.340     & 0.749     & 0.753     \\
hLSTMat \citep{Gao2019HierarchicalCaptioning}                       & 0.530     & -         & 0.336     & 0.738     & 0.753     \\
SCN-LSTM \citep{Gan2017SemanticCaptioning}                          & 0.511     & -         & 0.335     & 0.777     & 0.754     \\
MS-RNN(R) \citep{Song2019FromCaptioning}                            & 0.533     & -         & 0.338     & 0.748     & 0.759     \\
\citet{Wei2020ExploitingCaptioning}	                                & 0.468	    & -	        & 0.344	    & 0.857	    & 0.762     \\
TDConvED (R) \citep{Chen2019TemporalCaptioning}                     & 0.533     & -         & 0.338     & 0.764     & 0.764     \\
PickNet(V+L) \citep{Chen2018LessCaptioning}                         & 0.523     & 0.692     & 0.333     & 0.765     & 0.784     \\
MTLE+ResNet \citep{Nina2018MTLE:Description}                        & 0.530     & -         & 0.318     & -         & 0.788     \\
GRU-EVE \citep{Aafaq2019Spatio-TemporalCaptioning}                  & 0.479     & 0.715     & 0.350     & 0.781     & 0.788     \\
topic-guided \citep{Chen2019GeneratingGuidance}                     & 0.492     & 0.710     & 0.339     & 0.830     & 0.795     \\
Res-F2F \citep{Tang2019RichCaptioning}                              & 0.524     & -         & 0.357     & 0.843     & 0.797     \\
RecNet(RL)$_{g+l}$ \citep{Zhang2019ReconstructLearning}             & 0.529     & -         & 0.348     & 0.859     & 0.798     \\ 
RecNet$_{local}$(SA-LSTM) \citep{Wang2018ReconstructionCaptioning}  & 0.523     & 0.698     & 0.341     & 0.803     & 0.799     \\
MSAN$_{f+o+c}$ \citep{Sun2019MultimodalCaptioning}                  & 0.564     & -         & 0.353     & 0.796     & 0.801     \\
Non-local enc. \citep{Lee2019CapturingCaptioning}                   & 0.497     & 0.717     & 0.337     & 0.845     & 0.802     \\ 
ECO \citep{Zolfaghari2018ECO:Understanding}                         & 0.535     & -         & 0.350     & 0.858     & 0.802     \\
M$^3$-IC \citep{Wang2018M3:Captioning}                              & 0.528     & -         & 0.333     & -         & 0.804     \\
E2E(beam-search) \citep{Li2019End-to-EndLearning}                   & 0.503     & 0.708     & 0.341     & 0.875     & 0.810     \\
Attr-Attn(I+C+A+L+M) \citep{Xiao2019ACaptioning}                    & 0.565     & -         & 0.354     & 0.861     & 0.821     \\
Joint-VisualPOS \citep{Hou2019JointCaptioning}                       & 0.528     & 0.715     & 0.361     & 0.878     & 0.834     \\
GFN-POS\_RL(IR+M) \citep{Wang2019ControllableNetwork}                & 0.539     & 0.721     & 0.349     & 0.910     & 0.840     \\
ORG-TRL \citep{Zhang2020ObjectCaptioning}                           & 0.543     & 0.739     & 0.364     & 0.952     & 0.866     \\
SCN-LSTM+sampling \citep{Chen2020ASampling}                         & 0.618     & 0.768     & 0.378     & 1.030     & 0.929     \\
AVSSN \citep{Perez-Martin2020AttentiveCaptioning}                   & 0.623     & 0.783     & 0.392     & 1.077     & 0.954     \\
SemSynAN \citep{Perez-Martin2021ImprovingEmbedding}                 & \ul{0.644}& \bf 0.795 & \bf 0.419 & \ul{1.115}& \ul{0.990}\\
VNS-GRU \citep{Chen2020DelvingCaptioning}                           & \bf 0.649 & \ul{0.785}& \ul{0.411}& \bf 1.15  & \bf 0.992 \\
    \bottomrule
    \end{tabular}
    \caption{State-of-the-art for description generation on MSVD dataset.
    We consider the experiments results reported by works from 2017.
    We sorted the methods by 
    the $\text{S}_{\text{overall}}$ score, defined in Equation~\ref{eq:s-overall}.}
    \label{tab:results:dg:msvd}
\end{table}

Table~\ref{tab:results:dg:msr-vtt} shows that these two methods also achieve the highest $\text{S}_{\text{overall}}$ scores for the MSR-VTT dataset, but they are not the best for CIDEr.
The POS tagging-based GFN-POS\_RL(IR+M)~\citep{Wang2019ControllableNetwork} approach and the reviewing network-based REVnet$_{v3}$-RL~\citep{Li2019REVnet:Description} method show the impact of Reinforcement Learning.
They achieve the best CIDEr score by using RL to directly optimize it, explaining the increase in performance and the margin with the other metrics.

\begin{table}[t]
    \centering
    \fontsize{6pt}{6pt}\selectfont
    \begin{tabular}{lccccc}
    \toprule
    Method                                                                       & BLEU-4     & ROUGE-L     & METEOR     & CIDEr-D      & $\text{S}_{\text{overall}}$ \\
    \midrule                  
    ReBiLSTM (shortcut) \citep{Bin2019DescribingLSTM}                            & 0.339      & -           & 0.266      & -          & 0.786       \\
    LSTM-GAN+attn. \citep{Yang2018VideoLSTM}                                     & 0.360      & -           & 0.261      & -          & 0.800       \\
    aLSTMs \citep{Gao2017VideoConsistency}                                       & 0.380      & -           & 0.261      & -          & 0.822       \\
    M$^3$-VC \citep{Wang2018M3:Captioning}                                       & 0.381      & -           & 0.266      & -          & 0.831       \\
    Two-stream \citep{Gao2019HierarchicalCaptioning}                             & 0.397      & -           & 0.270      & 0.421      & 0.833       \\
    \citet{Wei2020ExploitingCaptioning}                                          & 0.385	  & -           & 0.269      & 0.437	  & 0.834       \\
    TDConvED (R) \citep{Chen2019TemporalCaptioning}                              & 0.395      & -           & 0.275      & 0.428      & 0.841       \\
    MS-RNN(R) \citep{Song2019FromCaptioning}                                     & 0.398      & 0.593       & 0.261      & 0.409      & 0.842       \\
    STAT\_L \citep{Yan2020STAT:Captioning}                                       & 0.393      & -           & 0.271      & 0.439      & 0.843       \\
    MTLE+ResNet \citep{Nina2018MTLE:Description}                                 & 0.392      & 0.593       & 0.266      & 0.421      & 0.848       \\ 
    RecNet$_{local}$ \citep{Wang2018ReconstructionCaptioning}                    & 0.391      & 0.593       & 0.266      & 0.427      & 0.850       \\
    PickNet(V+L) \citep{Chen2018LessCaptioning}                                  & 0.389      & 0.595       & 0.272      & 0.421      & 0.852       \\
    TDDF(VGG+C3D) \citep{Zhang2017Task-DrivenDescription}                        & 0.373      & 0.592       & 0.278      & 0.438      & 0.855       \\
    Attr-Attn(I+C+A+L+M) \citep{Xiao2019ACaptioning}                             & 0.401      & -           & 0.272      & 0.455      & 0.859       \\
    RecNet(RL)$_{g+l}$ \citep{Zhang2019ReconstructLearning}                      & 0.393      & -           & 0.277      & 0.495      & 0.884       \\
    GRU-EVE \citep{Aafaq2019Spatio-TemporalCaptioning}                           & 0.383      & 0.607       & 0.284      & 0.481      & 0.891       \\
    E2E \citep{Li2019End-to-EndLearning}                                         & 0.404      & 0.610       & 0.270      & 0.483      & 0.893       \\
    DenseVidCap \citep{Shen2017WeaklyCaptioning}                                 & 0.414      & 0.611       & 0.283      & 0.489      & 0.912       \\
    HRL \citep{Wang2018VideoLearning}                                            & 0.413      & 0.617       & 0.287      & 0.480      & 0.913       \\
    Res-F2F \citep{Tang2019RichCaptioning}                                       & 0.414      & 0.613       & 0.290      & 0.489      & 0.919       \\
    Joint-VisualPOS \citep{Hou2019JointCaptioning}                                & 0.423      & 0.628       & 0.297      & 0.491      & 0.935       \\
    HACA \citep{Wang2018WatchCaptioning}                                         & 0.434      & 0.618       & 0.295      & 0.497      & 0.939       \\
    GFN-POS\_RL(IR+M) \citep{Wang2019ControllableNetwork}                         & 0.413      & 0.621       & 0.287      & \bf 0.534  & 0.939       \\
    REVnet$_{v3}$-RL \citep{Li2019REVnet:Description}                            & 0.424      & 0.623       & 0.281      & \ul{0.532} & 0.941       \\
    ORG-TRL \citep{Zhang2020ObjectCaptioning}                                    & 0.436      & 0.621       & 0.288      & 0.509      & 0.941       \\
    CST\_GT\_None \citep{Phan2017Consensus-basedCaptioning}                      & 0.441      & 0.624       & 0.291      & 0.497      & 0.942       \\
    SCN-LSTM+sampling \citep{Chen2020ASampling}                                  & 0.438      & 0.624       & 0.289      & 0.514      & 0.947       \\
    topic-guided \citep{Chen2019GeneratingGuidance}                              & 0.449      & 0.628       & 0.296      & 0.518      & 0.962       \\
    VNS-GRU \citep{Chen2020DelvingCaptioning}                                    & 0.460      & 0.633       & 0.295      & 0.520      & 0.970       \\
    MSAN$_{f+o+c}$ \citep{Sun2019MultimodalCaptioning}                           & \bf 0.468  & -           & 0.295      & 0.524      & 0.975       \\
    AVSSN \citep{Perez-Martin2020AttentiveCaptioning}                            & 0.455      & \ul{0.643}  & \bf 0.314  & 0.506      & \ul{0.979}  \\ 
    SemSynAN \citep{Perez-Martin2021ImprovingEmbedding}                          & \ul{0.464} & \bf 0.647   & \ul{0.304} & 0.519      & \bf 0.984   \\
    \bottomrule
    \end{tabular}
    \caption{State-of-the-art for description generation on MSR-VTT dataset.
    We consider the experiments results reported by works from 2017.
    We sorted the methods by 
    the $\text{S}_{\text{overall}}$ score, defined in Equation~\ref{eq:s-overall}.}
    \label{tab:results:dg:msr-vtt}
\end{table}

In Table~\ref{tab:results:charades}, we summarize the results on the Charades dataset.
One of the first reported results on this dataset is for the HRL-16\savefn{note:hrl16}{They achieved the results on the Charades Caption dataset, which was obtained by pre-processing the raw Charades dataset.}~\citep{Wang2018VideoLearning} method.
Although this model achieves the best $\text{S}_{\text{overall}}$ score, the more recent CAM-RNN(GoogleNet)~\citep{Zhao2019CAM-RNN:Captioning} method surpassed HRL-16 for METEOR by an absolute margin of 0.002.
The method proposed by \citet{Wei2020ExploitingCaptioning} reports the pour performance.
This method is based on a recurrent encoder, which has not shown robustness for recognizing fine-grained activities and generating detailed descriptions.

\begin{table}[t]
    \centering
    \scriptsize
    \begin{tabular}{lccccc}
    \toprule
    Method                                                          & BLEU-4     & ROUGE     & METEOR    & CIDEr-D   & $\text{S}_{\text{overall}}$ \\
    \midrule                
    \citet{Wei2020ExploitingCaptioning}                             & 0.127      & -         & 0.172     & \ul{0.216}& 0.827     \\
    CAM-RNN(GoogleNet) \citep{Zhao2019CAM-RNN:Captioning}           & 0.129      & -         & \bf 0.197 & 0.188     & 0.832     \\
    TSA-ED \citep{Wu2018InterpretableLocalization}                  & \ul{0.135} & -         & 0.178     & 0.208     & \ul{0.839}\\
    HRL-16\reffn{note:hrl16} \citep{Wang2018VideoLearning}          & \bf 0.188  & \bf 0.414 & \ul{0.195}& \bf 0.232 & \bf 0.997 \\
    \bottomrule
    \end{tabular}
    \caption{State-of-the-art for description generation on Charades dataset. 
    We sorted the methods by 
    the $\text{S}_{\text{overall}}$ score, defined in Equation~\ref{eq:s-overall}.}
    \label{tab:results:charades}
\end{table}

Regarding the benchmarking for movie description, results reported on the complex M-VAD dataset are inferior in general.
For the three datasets showed in Table~\ref{tab:results:dg:lsmdc}, all papers report METEOR.
On M-VAD, SF-SSAG-LSTM~\citep{Xu2019Semantic-filteredFeature}, followed by BAE~\citep{Baraldi2017HierarchicalCaptioning}, achieves the best METEOR score.
While, only SA~\citep{Yao2015DescribingStructure}, HRNE~\citep{Pan2016HierarchicalCaptioning}, and MHB~\citep{Sah2020UnderstandingCaptioning} report results for BLEU-4, being  MHB~\citep{Sah2020UnderstandingCaptioning} the best one.
In contrast, on the MPII-MD dataset, only BAE~\citep{Baraldi2017HierarchicalCaptioning} reports the four metrics, LSTM-TSA$_{IV}$ \citep{Pan2017VideoAttributes} obtains the best-reported result for METEOR, and no experiments results have been reported since 2017.

\begin{table}[t]
    \centering
    \scriptsize
    \begin{threeparttable}
    \begin{tabular}{lccccc}
        \toprule
        Method                                                  & BLEU-4     & ROUGE     & METEOR    & CIDEr-D   & $\text{S}_{\text{overall}}$ \\
        \midrule            
        \multicolumn{6}{c}{M-VAD dataset}\\         
        \midrule            
        HRNE \citep{Pan2016HierarchicalCaptioning}              & \ul{0.007} & -         & 0.068     & -         & 0.760      \\
        Visual-Labels \citep{Rohrbach2015TheDescription}        & -          & -         & 0.064     & -         & 0.771      \\
        SA \citep{Yao2015DescribingStructure}                   & \ul{0.007} & -         & 0.057     & \bf 0.061 & 0.796      \\
        S2VT \citep{Venugopalan2015SequenceText}                & -          & -         & 0.067     & -         & 0.807      \\
        LSTM-E \citep{Pan2016JointlyLanguage}                   & -          & -         & 0.067     & -         & 0.807      \\
        Glove+DeepFusion \citep{Venugopalan2016ImprovingText}   & -          & -         & 0.068     & -         & 0.819      \\
        LSTM-TSA$_{IV}$ \citep{Pan2017VideoAttributes}          & -          & -         & 0.072     & -         & 0.867      \\
        BAE \citep{Baraldi2017HierarchicalCaptioning}           & -          & -         & \ul{0.073}& -         & 0.880      \\
        MHB \citep{Sah2020UnderstandingCaptioning}              & \bf 0.010  & -         & 0.069     & -         & \ul{0.916} \\
        SF-SSAG-LSTM \citep{Xu2019Semantic-filteredFeature}     & -          & -         & \bf 0.083 &           & \bf 1.000  \\ 
        \midrule    
        \multicolumn{6}{c}{MPII-MD dataset}\\   
        \midrule    
        SMT \citep{Rohrbach2015ADescription}                    & -          & -         & 0.056     & -         & 0.700      \\
        Glove+DeepFusion \citep{Venugopalan2016ImprovingText}   & -          & -         & 0.068     & -         & 0.850      \\
        Visual-Labels \citep{Rohrbach2015TheDescription}        & -          & -         & 0.070     & -         & 0.875      \\
        S2VT \citep{Venugopalan2015SequenceText}                & -          & -         & 0.071     & -         & 0.888      \\
        LSTM-E \citep{Pan2016JointlyLanguage}                   & -          & -         & \ul{0.073}& -         & 0.913      \\
        BAE \citep{Baraldi2017HierarchicalCaptioning}           & \bf 0.008  & \bf 0.167 & 0.070     & \bf 0.108 & \ul{0.969} \\
        LSTM-TSA$_{IV}$ \citep{Pan2017VideoAttributes}          & -          & -         & \bf 0.080 & -         & \bf 1.000  \\
        \midrule
        \multicolumn{6}{c}{LSMDC dataset}\\
        \midrule
        MTLE+ResNet \citep{Nina2018MTLE:Description}            & 0.005      & -         & 0.055     & 0.087     & 0.753      \\ 
        GEAN+GNet+C3D+Scene \citep{Yu2017SupervisingData}       & -*         & \ul{0.156}& \bf 0.072 & \ul{0.093}& \ul{0.970} \\
        CT-SAN \citep{Yu2017EndAnswering}                & \bf 0.008  & \bf 0.159 & \ul{0.071}& \bf 0.100 & \bf 0.997  \\
        \bottomrule
    \end{tabular}
    \begin{tablenotes}
        \scriptsize
        \item *did not report $\text{BLEU-4}$, but reported $\text{BLEU-3}=0.021$.
    \end{tablenotes}
    \end{threeparttable}
    \caption{State-of-the-art for description generation on Movie Description datasets.
    In this table we considered the results reported in each paper on the M-VAD, MPII-MD and LSMDC datasets.
    We sorted the methods by 
    the $\text{S}_{\text{overall}}$ score, defined in Equation~\ref{eq:s-overall}.}
    \label{tab:results:dg:lsmdc}
\end{table}

Table~\ref{tab:results:cook} shows the state-of-the-art results on two cooking-videos datasets, \ie TACoS-Multilevel and YouCook2.
For the TACoS-Multilevel dataset, h-RNN~\citep{Yu2016VideoNetworks} obtained the best result for BLEU, METEOR, and CIDEr.
For the challenging YouCook2 dataset, the researchers report results using the validation set's ground-truth proposals (localization), MART~\citep{Lei2020MART:Captioning} is shown as the best method, achieving the best score for all metrics they reported.

\begin{table}[t]
    \centering
    \scriptsize
    \begin{threeparttable}
    \begin{tabular}{lccccc}
    \toprule
    Method                                                  & BLEU-4     & ROUGE     & METEOR    & CIDEr-D   & $\text{S}_{\text{overall}}$ \\
    \midrule
    \multicolumn{6}{c}{TACoS-Multilevel}\\
    \midrule
    JEDDi-Net \citep{Xu2019JointStreams}                    & \ul{0.181} & \bf 0.509 & \ul{0.239}& \ul{1.040}& \ul{0.769}\\
    h-RNN \citep{Yu2016VideoNetworks}                       & \bf 0.305  & -         & \bf 0.287 & \bf 1.602 & \bf 1.000 \\
    \midrule
    \multicolumn{6}{c}{YouCook2 (validation set, using ground-truth proposals*)}\\
    \midrule
    M. Transformer \citep{Zhou2018End-to-EndTransformer}    & 0.142      & -         & 0.112     & -         & 0.441     \\
    VideoBERT+S3D \citep{Sun2019VideoBERT:Learning}         & \ul{0.433} & \bf 0.288 & \ul{0.119}& \ul{0.006}& \ul{0.577}\\
    MART \citep{Lei2020MART:Captioning}                     & \bf 0.800  & -         & \bf 0.159 & \bf 0.357 & \bf 1.000 \\
    \bottomrule
    \end{tabular}
    \begin{tablenotes}
        \scriptsize
        \item *YouCook2 Captions dataset includes temporal localization of video segments, and captions related to segments. For this, to evaluate the video captioning task only, it is needed to use the ground-truth video segments information.
    \end{tablenotes}
    \end{threeparttable}
    \caption{State-of-the-art for description generation on Cooking datasets.
    In this table we considered the results reported in each paper on the TACoS-Multilevel and YouCook2 datasets.
    We sorted the methods by 
    the $\text{S}_{\text{overall}}$ score, defined in Equation~\ref{eq:s-overall}.}
    \label{tab:results:cook}
\end{table}

As we mentioned in Section~\ref{sec:dataset:activitynet}, the ActivityNet Captions dataset also includes temporal events localization and reference descriptions for these events.
To exploit this information to its full extent, the researchers have evaluated their captioning methods using the ground-truth localization (like in YouCook2 dataset) or trying to predict them.
The results of these experiments are reported on the validation set.
For this dataset, the testing videos' ground-truth is not revealed and is evaluated by the official test server\reffn{note:actnettask} as part of the ActivityNet Challenge (see Section~\ref{sec:challenge:activitynet}).

The dense video captioning problem using the ground-truth localization of events is similar to the video description generation problem.
We do not need to predict event proposals before captioning the video.
As we can see in Table~\ref{tab:results:dg:activitynet}, for these experiments, COOT (video+clip)~\citep{Ging2020COOT:Learning} achieved the best performance for BLEU-4, ROUGE, and METEOR.
In comparison, RUC+CMU\reffn{note:actnet19}~\citep{Chen2019ActivitynetVideos} achieved the best performance for CIDEr and the second-best one for BLEU-4.

\begin{table}[t]
\centering
\fontsize{4.3pt}{4.3pt}\selectfont
\begin{threeparttable}
\begin{tabular}{l|cccc|cccc|c}
\toprule
Method                                                         & \multicolumn{8}{c}{Validation set}                                                                 & \multicolumn{1}{|c}{Test} \\
                                                               & \multicolumn{4}{c}{predicting event proposals}   & \multicolumn{4}{|c|}{using GT proposals*}       & Server\reffn{note:actnettask}\\
                                                               & BLEU-4       & ROUGE     & METEOR    & CIDEr-D   & BLEU-4      & ROUGE     & METEOR    & CIDEr-D   & METEOR      \\
\midrule            
DEM \citep{Krishna2017Dense-CaptioningVideos}                  & -            & -         & -         & -         & -           & -         & -         & -         & 0.048       \\
LSTM-A$_3$ \citep{Yao2017MSRVideos}                            & \ul{0.031}   & 0.143     & 0.087     & 0.148     & -           & -         & -         & -         & -           \\
RUC+CMU\reffn{note:actnet18} \citep{Chen2018RUC+CMU:Videos}    & \bf 0.040    & -         & \bf 0.124 & -         & \ul{0.040}  & -         & 0.138     & \bf 0.565 & 0.085       \\
M. Transformer \citep{Zhou2018End-to-EndTransformer}           & 0.022        & -         & 0.096     & -         & 0.027       & -         & 0.111     & -         & \bf 0.101   \\
Bi-SST \citep{Wang2018BidirectionalCaptioning}                 & 0.023        & 0.191     & 0.096     & 0.127     & -           & -         & -         & -         & 0.097       \\
SDVC \citep{Mun2019StreamlinedCaptioning}                      & 0.009        & -         & 0.088     & \bf 0.306 & 0.013       & -         & 0.131     & \ul{0.435}& 0.082       \\
JEDDi-Net \citep{Xu2019JointStreams}                           & 0.016        & \ul{0.196}& 0.086     & \ul{0.199}& -           & -         & -         & -         & 0.088       \\
DaS \citep{Zhang2019ShowSummarization}                         & 0.021        & \bf 0.212 & \ul{0.103}& 0.129     & 0.016       & 0.229     & 0.107     & 0.314     & -           \\
RecNet(RL)$_{local}$ \citep{Zhang2019ReconstructLearning}      & -            & -         & -         & -         & 0.017       & \ul{0.235}& 0.105     & 0.384     & -           \\
RUC+CMU\reffn{note:actnet19} \citep{Chen2019ActivitynetVideos} & -            & -         & -         & -         & -           & -         & \ul{0.143}&  -        & 0.099       \\
COOT (video+clip) \citep{Ging2020COOT:Learning}                & -            & -         & -         & -         & \bf 0.109   & \bf 0.315 & \bf 0.160 & 0.282     & -           \\         
\bottomrule
\end{tabular}
\begin{tablenotes}
    \tiny
    \item *ActivityNet Captions dataset includes temporal action proposal, temporal action localization, and captions related to each action (video segment). For this, to evaluate the video captioning task only, it is needed to use the ground-truth video segments information.
\end{tablenotes}
\end{threeparttable}
\caption{State-of-the-art for description generation on ActivityNet Captions dataset.
In this table we considered the results reported in each paper on the ActivityNet Captions dataset and the evaluation servers of 2018\reffn{note:actnet18} and 2019\reffn{note:actnet19} editions of the ActivityNet Challenge.
We sorted the methods by year and those of the same year by the METEOR score reported by Test server.}
\label{tab:results:dg:activitynet}
\end{table}

Likewise, the authors of almost these methods have participated in the annual ActivityNet Challenge, evaluated on the official test server's test set.
This server reports the Avg-METEOR score, being M. Transformer~\citep{Zhou2018End-to-EndTransformer} and RUC+CMU\reffn{note:actnet19}~\citep{Chen2019ActivitynetVideos} the best methods of 2018 and 2019, respectively.

Table~\ref{tab:results:dg:trecvid} shows the state-of-the-art results on the TRECVID-VTT datasets for the description generation subtask.
In the TRECVID Challenges, the STS metric is also reported (see Section~\ref{sec:challenge:trecvid}).
In 2019, scores increased for all metrics compared with 2018.
Likewise, in 2020, although videos from a new video source were incorporated, scores also increased compared to the 2019 results, except BLEU-4.

\begin{table}[t]
    \centering
    \tiny
    \begin{tabular}{l|c|ccccccc}
        \toprule
        Method                                          & Dataset               & BLEU-4 & METEOR & CIDEr & CIDEr-D  & SPICE & STS   \\
        \midrule
        MTLE+ResNet \citep{Nina2018MTLE:Description}    & TRECVID-VTT'2016      & 0.122  & 0.374  & 0.423 & -        & -     & 0.462 \\
        RUC\_CMU \citep{Chen2017Informedia2017}         & TRECVID-VTT'2017      & 0.023  & 0.198  & 0.408 & -        & -     & -     \\
        INF \citep{Chen2018InformediaTRECVID2018}       & TRECVID-VTT'2018      & 0.024  & 0.231  & 0.416 & 0.585    & -     & 0.433 \\
        RUC\_AIM3 \citep{Song2019RUC_AIM3Text}          & TRECVID-VTT'2019      & 0.064  & 0.306  & 0.154 & 0.332    & -     & 0.484 \\
        RUC\_AIM3 \citep{Zhao2020RUC_AIM3Description}   & TRECVID-VTT'2020      & 0.056  & 0.310  & 0.303 & -        & 0.110 & -     \\
        \bottomrule
    \end{tabular}
    \caption{State-of-the-art for description generation on TRECVID-VTT datasets.}
    \label{tab:results:dg:trecvid}
\end{table}

Finally, Table~\ref{tab:results:dg:vatex} lists the results on the recent \VATEX{} dataset.
The Baseline results for this dataset were reported by \citet{Wang2019VaTeXResearch} when the dataset was proposed.
The methods ORG-TRL~\citep{Zhang2020ObjectCaptioning}, Top-down + X-LAN~\citep{Lin2020Multi-modal2020}, and X-Linear+Transformer~\citep{Guo2020SequenceCaptioning} participated in the \VATEX{} video captioning challenge 2020 (see Section~\ref{sec:challenge:vatex}) and report their results.
The use of multiple features such as I3D, ECO, and audio, and the hybrid reward strategy proposed by X-Linear+Transformer~\citep{Guo2020SequenceCaptioning} reports the highest result on the \VATEX{} dataset.
They were the winners of the English video captioning competition.

\begin{table}[t]
    \centering
    \scriptsize
    \begin{tabular}{lccccc}
    \toprule
    Method                                                  & BLEU-4     & ROUGE     & METEOR    & CIDEr-D   & $\text{S}_{\text{overall}}$ \\
    \midrule    
    NITS-VC \citep{Singh2020NITS-VC2020}                    & 0.220      & 0.430     & 0.180     & 0.270     & 0.593 \\
    Baseline \citep{Wang2019VaTeXResearch}                  & 0.285      & 0.470     & 0.216     & 0.451     & 0.742 \\
    ORG-TRL \citep{Zhang2020ObjectCaptioning}               & 0.321      & 0.489     & 0.222     & 0.497     & 0.793 \\
    Top-down + X-LAN \citep{Lin2020Multi-modal2020}         & \ul{0.392} & \ul{0.527}& \ul{0.250}& \ul{0.760}& \ul{0.962} \\
    X-Linear+Transformer \citep{Guo2020SequenceCaptioning}  & \bf 0.407  & \bf 0.537 & \bf 0.258 & \bf 0.814 & \bf 1.000 \\
    \bottomrule
    \end{tabular}
    \caption{State-of-the-art for description generation on \VATEX{} dataset.
    In this table we considered the results reported in each paper on the \VATEX{} dataset.
    We sorted the methods by 
    the $\text{S}_{\text{overall}}$ score, defined in Equation~\ref{eq:s-overall}.}
    \label{tab:results:dg:vatex}
\end{table}

\subsection{State of the Art on Retrieval Methods}

In this study, we also compare the state-of-the-art VTT methods based on retrieval strategy.
As we study in Section~\ref{sec:matching-ranking}, these works are generally based on learning a joint video-text embedding, which can perform the retrieval in both directions.
For that, some papers report performances for both \emph{video retrieval} and \emph{description retrieval} facets. 

Table~\ref{tab:results:mr:msvd} shows the results on the MSVD dataset.
\citet{Mithun2018LearningRetrieval} reported three versions of their model, and their Fusion version reports the best Sum of Recalls result.
Table~\ref{tab:results:mr:msr-vtt} lists the retrieval-based results on the MSR-VTT dataset.
For this dataset, \citet{Mithun2018LearningRetrieval} also achieved good performance, but Dual Encoding~\citep{Dong2019DualRetrieval} model achieves the best Sum of Recalls performance of the methods that reported description retrieval experiments.
Table~\ref{tab:results:mr:activitynet} shows the experimental results for retrieval-based methods on the ActivityNet Captions dataset.
The recent COOT~\citep{Ging2020COOT:Learning} approach achieves the best results for all metrics, obtaining a median rank for retrieval in both directions of one.
Compared to the other methods, this is the first approach based on Transformers that reports retrieval experiments on this dataset.
Finally, for TRECVID-VTT datasets, the description retrieval methods are evaluated by MIR score.
In 2018, \citet{Li2018RenminRetrieval} obtained the highest result with 0.516.
While in 2019, \citet{Song2019RUC_AIM3Text} improved that result, obtaining 0.727.

\begin{table}[t]
    \centering
    \tiny
    \begin{tabular}{l|ccccc|ccccc|c}
\toprule
Method                                      & \multicolumn{5}{c|}{Description retrieval}    & \multicolumn{5}{c|}{Video retrieval}           & Avg. Sums of \\
                                            & R@1     & R@5     & R@10    & MR   & mAP      & R@1     & R@5      & R@10    & MR   & mAP      & Recalls\\
\midrule                                             
W2VV \citep{Dong2018PredictingRetrieval}    &18.5     &36.7     &45.1     &-     &-         &-        &-         &-        &-     &0.230     & 100.3*\\
i3D \citep{Mithun2018LearningRetrieval}     &21.3     &43.7     &\ul{53.3}&9     &\ul{0.722}&15.4     &39.2      &51.4     &10    &\bf 0.432 & 112.2\\
ResNet \citep{Mithun2018LearningRetrieval}  &\ul{23.4}&\ul{45.4}&53.0     &\ul{8}&\bf 0.752 &\ul{16.1}&\ul{41.1} &\ul{53.5}&\ul{9}&\ul{0.427}& \ul{116.3}\\
Fusion \citep{Mithun2018LearningRetrieval}  &\bf 31.5 &\bf 51.0 &\bf 61.5 &\bf 5 &0.417     &\bf 20.3 &\bf 47.8  &\bf 61.1 &\bf 6 &0.283     & \bf 136.6\\
Dual Encoding \citep{Dong2019DualRetrieval} &-        &-        &-        &-     &-         &-        &-         &-        &-     &0.232     & - \\
\bottomrule
    \end{tabular}
    \caption{State-of-the-art for Matching-and-Ranking on MSVD dataset. In this table we considered the results reported in each paper on the MSVD dataset for both facets of the task: description retrieval from videos (Description retrieval) and video retrieval from text descriptions (Video retrieval). 
    Higher recall at K (R@1, R@5, R@10) and mAP, and smaller Median Rank (MR) indicates better performance.
    Methods are sorted by year. 
    For overall comparison, we compute the average between the sums of recalls of each facet. 
    We sorted the results by the Avg. Sum. of Recalls column.}
    \label{tab:results:mr:msvd}
\end{table}

\begin{table}[t]
    \centering
    \tiny
    \begin{tabular}{l|ccccc|ccccc|c}
        \toprule
        Method                                          & \multicolumn{5}{c|}{Description retrieval}             & \multicolumn{5}{c|}{Video retrieval}                       & Avg. Sums of \\
                                                        & R@1        & R@5       & R@10      & MR    & mAP       & R@1       & R@5       & R@10      & MR        & mAP        & Recalls \\
        \midrule                                        
        ResNet \citep{Mithun2018LearningRetrieval}      & 10.5       & 26.7      & 35.9      & 25    & \bf 0.267 & 5.8       & 17.6      & 25.2      & 61        & \bf 0.297  & 60.9   \\
        i3D+Audio \citep{Mithun2018LearningRetrieval}   & 9.3        & 27.8      & 38.0      & 22    & \ul{0.162}& 5.7       & 18.4      & 26.8      & 48        & \ul{0.243} & 63.0   \\
        W2VV \citep{Dong2018PredictingRetrieval}        & 11.8       & 28.9      & 39.1      & 21    & 0.058     & 6.1       & 18.7      & 27.5      & 45        & 0.131      & 66.1   \\
        Fusion \citep{Mithun2018LearningRetrieval}      & \ul{12.5}  & \bf 32.1  & \ul{42.4} & 16    & 0.134     & 7.0       & 20.9      & 29.7      & 38        & 0.214      & 72.3   \\
        Dual Encoding \citep{Dong2019DualRetrieval}     & \bf 13.0   & \ul{30.8} & \bf 43.3  & \bf 15& 0.065     & 7.7       & 22.0      & 31.8      & 32        & 0.155      & 74.3   \\
        HGR \citep{Chen2020Fine-grainedReasoning}       & 9.2        & 26.2      & 36.5      & 24    & -         & \bf 15.0  & \ul{36.7} & \ul{48.8} & \ul{11}   & -          & 86.2   \\
        S3D \citep{Miech2020End-to-EndVideos}           & -          & -         & -         & -     & -         & 9.9       & 24.0      & 32.4      & 29.5      & -          & 95.8*  \\
        HowTo100M \citep{Miech2019HowTo100M:Clips}      & -          & -         & -         & -     & -         & \ul{14.9} & \bf 40.2  & \bf 52.8  & \bf 9     & -          & 116.9* \\
        \bottomrule
    \end{tabular}
    \caption{State-of-the-art for Matching-and-Ranking on MSR-VTT dataset.
    In this table we considered the results reported in each paper on the MSR-VTT dataset for both facets of the task: description retrieval from videos (Description retrieval) and video retrieval from text descriptions (Video retrieval). 
    Higher recall at K (R@1, R@5, R@10) and mAP, and smaller Median Rank (MR) indicates better performance. 
    For overall comparison, we compute the average between the sums of recalls of each facet. 
    We sorted the results by the Avg. Sum. of Recalls column.}
    \label{tab:results:mr:msr-vtt}
\end{table}

\begin{table}[t]
    \centering
    \fontsize{5pt}{5pt}\selectfont
    \begin{tabular}{l|ccccc|ccccc|c}
        \toprule
        Method                                          & \multicolumn{5}{c|}{Description retrieval}             & \multicolumn{5}{c|}{Video retrieval}                  & Avg. Sums of \\
                                                        & R@1        & R@5       & R@10 & R@50       & MR        & R@1       & R@5       & R@10 & R@50       & MR        & Recalls\\
        \midrule                                        
        LSTM-YT \citep{Venugopalan2015SequenceText}     & 0.0        & 7.0       & -    & 38.0       & 98        & 0.0       & 4.0       & -    & 24.0       & 102       & 36.5   \\
        DENSE \citep{Krishna2017Dense-CaptioningVideos} & 18.0       & 36.0      & -    & 74.0       & 32        & 14.0      & 32.0      & -    & 65.0       & 34        & 119.5  \\
        HSE \citep{Zhang2018Cross-ModalText}            & \ul{44.2}  & \ul{76.7} & -    & \ul{97.0}  & \ul{2}    & \ul{44.4} & \ul{76.7} & -    & \ul{97.1}  & \ul{2}    & \ul{218.1}  \\
        COOT \citep{Ging2020COOT:Learning}              & \bf 60.9   & \bf 87.4  & -    & \bf 98.6   & \bf 1     & \bf 60.8  & \bf 86.6  & -    & \bf 98.6   & \bf 1     & \bf 246.3  \\
        \bottomrule
    \end{tabular}
    \caption{State-of-the-art for Matching-and-Ranking on ActivityNet Captions dataset (val-1 split). 
    In this table we considered the results reported in each paper on the ActivityNet Captions dataset for both facets of the task: description retrieval from videos (Description retrieval) and video retrieval from text descriptions (Video retrieval). 
    Higher recall at K (R@1, R@5, R@10, R@50), and smaller Median Rank (MR) indicates better performance.
    For overall comparison, we compute the average between the sums of recalls of each facet. 
    We sorted the results by the Avg. Sum. of Recalls column.}
    \label{tab:results:mr:activitynet}
\end{table}

\section{Discussion and Conclusions}\label{sec:limitations}

In this review, we have analyzed the state-of-the-art techniques for developing VTT solutions.
The techniques based on Deep Learning have achieved promising results for both description generation and retrieval-based methods.
Despite the significant progress in descriptive text generation and retrieval tasks for several benchmark datasets, we can say that the state-of-the-art methods fail to extract/capture all the complex spatiotemporal information present in videos.
There is still much work to do for understanding the diversity regarding the visual content in videos and the structure of associated textual descriptions.

As a text generation task, video captioning requires predicting a semantic and syntactically correct sequence of words given some context video.
Early works followed the strategy of detecting Subject, Verb, and Object, forming an SVO triplet, and then a sentence.
This approach requires the models to recognize the subjects and objects that participate in the action we want to describe, achieving their best results in specific environments, such as sports or cooking.
It occurs because, in this kind of video clip, the number of objects and actions is limited, and the duration is generally short.

Recent works clearly show that videos contain implicit dimensions with valuable information about the possible descriptions in addition to the appearance and motion.
Directly from visual information, we can extract semantic and syntactic information to guide the text generation process (see Sections~\ref{sec:dg-semantic-based} and ~\ref{sec:dg-syntactic-based}).
However, having a strong dependence on only one of them can harms the models' performance on standard datasets, producing semantic gaps or syntactically incorrect sentences.
It is essential to determine how to combine these information channels adaptively.
Here, learning to ensemble both retrieval and generation techniques has been shown as a promising strategy.

The VTT field is strongly related to other tasks like \emph{action recognition}.
Many of the challenges that emerge in solving these tasks will also appear addressing video-text translation.
For instance, the video-text translation approaches also need to generalize over variations within one target class, distinguish between different classes in the input, and accurately depict the critical activities in a video clip. 
Here, the representation of informative semantics by learning to ensemble visual perception models plays a crucial role in video captioning.
Some state-of-the-art methods report that learning high-level discriminative features~\citep{Pan2017VideoAttributes, Gan2017SemanticCaptioning, Snoek2017SearchingVideo} and incorporating visual classifiers in the encoding process~\citep{Rohrbach2015TheDescription, Xu2015JointlyFramework, Otani2016LearningSearch, Pan2016JointlyLanguage} are useful approaches to deal with these variations.
These methods show the benefits of describing the videos according to dynamic visual and semantic information.
However, these models' performance has a strong dependence on the quality of semantic concept detection models. 

To accurately distinguish between different classes from visual information, the models must be trained on high-quality, diverse captions that describe a wide variety of videos at scale.
Creating large-scale datasets for VTT requires a significant and expensive human effort for their annotation because collecting a large number of references can be time-consuming and difficult for less common languages.
To deal with this limitation, some researchers aimed to preserve the models' generalization properties by collecting their training data from interesting combinations of several existing datasets~\citep{Chen2018RUC+CMU:Videos, Snoek2017SearchingVideo, Li2018UTS_CETC_D2DCRCTask}.
In contrast, other works aimed to develop more robust models~\citep{Gan2017SemanticCaptioning, Baraldi2017HierarchicalCaptioning, Pan2017VideoAttributes}, but these models do not preserve the generalization properties well.

Recent works have shown the benefits of pre-training the models for multi-modal vision and language tasks and fine-tuning on the specific downstream tasks.
For example, we can pre-train a joint embedding for multi-modal tasks such as visual question answering or cross-modal retrieval and then fine-tuning its visual encoding on the video captioning task.
This technique could require more data for learning an accurate generic representation in the embedding space.
Generally, this data is automatically obtained from the subtitles and narrations provided by the online video platforms.
However, a major drawback of this kind of corpus, is the large number of unknown tokens that occur.
The HowTo100M's corpus is an excellent example of this issue, in which only 36.64\% of words in the vocabulary appear in the widely used GloVe-6B dictionary.
This high ``noise'' in the captions is an interesting aspect of the training process we must learn to take advantage of. 


Several researchers employ one of the evaluation metrics to select the best checkpoint model for testing, such as BLEU, CIDEr, or METEOR. 
However, some works~\citep{Vedantam2015CIDEr:Evaluation, Hodosh2015FramingAbstract, Chen2020DelvingCaptioning} have shown that a single measure cannot reflect the video description/caption generation methods' overall performance and generalization properties.
There is no standard evaluation method for video-text translation models, and current metrics have not shown sufficient robustness~\citep{Aafaq2019VideoMetrics}.
Most VTT metrics have been adopted from machine translation or image captioning and do not measure the quality of the generated descriptions from videos well.
These metrics give very different performance measures for the same method and are not perfectly aligned with human judgments~\citep{Aafaq2019VideoMetrics, Awad2018TRECVIDSearch, Vedantam2015CIDEr:Evaluation}. 
Some of the annual competitions provide a good contribution to deal with these evaluation issues.
In addition to automatic evaluation, these competitions carry out a human evaluation process of the participants' solutions. 

Another contribution in this sense can be the use of explicit semantic match-based metrics, such as SPICE.
However, even though SPICE correlates well with human evaluations, it depends on its internal parser quality and ignores the generated captions' fluency.
Section~\ref{sec:eval-limitations} summarizes some other limitations that carry the current evaluation metrics for video captioning.

Further research is still needed on dense captioning in long videos with multiple events that co-occur.
ActivityNet Competition started in 2017 to include this challenging subtask on the large-scale ActivityNet Captions dataset, a very representative dataset for training the models.
Future research could also be processing the audio information of videos, which has not been widely explored in the state-of-art approaches.
In the video caption/description generation task, the audio can be used for multi-modal research as extra knowledge to refine generated descriptions.
Some datasets like MSR-VTT include audio information, and several techniques for audio processing can be fine-tuned in the experiments.

In this comprehensive review, we have categorized and analyzed the most important approaches for the Video-to-Text problem.
We have also reviewed and compared the automatic evaluation metrics and the loss functions used in the optimization process.
Moreover, we have reviewed the popular benchmark datasets and the most related competitions commonly used for training and testing the models.
Lastly, we have summarized and analyzed the state-of-the-art results on each one of the principal datasets.

\bibliographystyle{myspbasic}      
\bibliography{reduced-references}   

\begin{thebibliography}{247}
\providecommand{\natexlab}[1]{#1}
\providecommand{\url}[1]{{#1}}
\providecommand{\urlprefix}{URL }
\expandafter\ifx\csname urlstyle\endcsname\relax
  \providecommand{\doi}[1]{DOI~\discretionary{}{}{}#1}\else
  \providecommand{\doi}{DOI~\discretionary{}{}{}\begingroup
  \urlstyle{rm}\Url}\fi
\providecommand{\eprint}[2][]{\url{#2}}

\bibitem[{Aafaq et~al.(2019{\natexlab{a}})Aafaq, Akhtar, Syed, Gilani, and
  Mian}]{Aafaq2019Spatio-TemporalCaptioning}
Aafaq N, Akhtar N, Syed WL, Gilani Z, Mian A (2019{\natexlab{a}})
  {Spatio-Temporal Dynamics and Semantic Attribute Enriched Visual Encoding for
  Video Captioning}. In: \emph{IEEE CVPR}, pp. 12487--12496

\bibitem[{Aafaq et~al.(2019{\natexlab{b}})Aafaq, Mian, Liu, Zulqarnain~Gilani,
  Mian, Liu, Gilani, and Shah}]{Aafaq2019VideoMetrics}
Aafaq N, Mian A, Liu W, Zulqarnain~Gilani S, Mian A, Liu W, Gilani SZ, Shah M
  (2019{\natexlab{b}}) {Video Description: A Survey of Methods, Datasets, and
  Evaluation Metrics}. \emph{ACM Computing Surveys} 52(6)

\bibitem[{Abbas et~al.(2019)Abbas, Ibrahim, and Jaffar}]{Abbas2019ASystems}
Abbas Q, Ibrahim ME, Jaffar MA (2019) {A comprehensive review of recent
  advances on deep vision systems}. \emph{Artificial Intelligence Review}
  52(1), 39--76

\bibitem[{Anderson et~al.(2016)Anderson, Fernando, Johnson, and
  Gould}]{Anderson2016SPICE:Evaluation}
Anderson P, Fernando B, Johnson M, Gould S (2016) {SPICE: Semantic
  Propositional Image Caption Evaluation}. In: \emph{ECCV}, Springer, Springer
  Nature, pp. 382--398

\bibitem[{Awad et~al.(2016)Awad, Fiscus, Joy, Michel, Smeaton, Kraaij,
  Eskevich, Aly, Ordelman, Jones, Huet, and
  Larson}]{Awad2016TRECVIDHyperlinking}
Awad G, Fiscus J, Joy D, Michel M, Smeaton AF, Kraaij W, Eskevich M, Aly R,
  Ordelman R, Jones GJF, Huet B, Larson M (2016) {TRECVID 2016: Evaluating vdeo
  search, video event detection, localization, and hyperlinking}. In:
  \emph{TRECVID}, Gaithersburg, Ma, US

\bibitem[{Awad et~al.(2017)Awad, Butt, Fiscus, Joy, Delgado, Michel, Smeaton,
  Graham, Kraaij, Qu{\'{e}}not, Eskevich, Ordelman, Jones, and
  Huet}]{Awad2017TrecvidHyperlinking}
Awad G, Butt A, Fiscus J, Joy D, Delgado A, Michel M, Smeaton A, Graham Y,
  Kraaij W, Qu{\'{e}}not G, Eskevich M, Ordelman R, Jones GJ, Huet B (2017)
  {Trecvid 2017: Evaluating ad-hoc and instance video search, events detection,
  video captioning and hyperlinking}. In: \emph{TRECVID}, Gaithersburg, Ma, US

\bibitem[{Awad et~al.(2018)Awad, Butt, Curtis, Yooyoung, Fiscus, Godil, Joy,
  Delgado, Smeaton, Graham, Kraaij, Qu{\'{e}}not, Magalhaes, Semedo, and
  Blasi}]{Awad2018TRECVIDSearch}
Awad G, Butt AA, Curtis K, Yooyoung L, Fiscus J, Godil A, Joy D, Delgado A,
  Smeaton AF, Graham Y, Kraaij W, Qu{\'{e}}not G, Magalhaes J, Semedo D, Blasi
  S (2018) {TRECVID 2018: Benchmarking Video Activity Detection, Video
  Captioning and Matching, Video Storytelling Linking and Video Search}. In:
  \emph{TRECVID}, NIST, Gaithersburg, Ma, US

\bibitem[{Awad et~al.(2019)Awad, Butt, Curtis, Lee, Fiscus, Godil, Delgado,
  Zhang, Godard, Diduch, Smeaton, Graham, Kraaij, and
  Qu{\'{e}}not}]{Awad2019TRECVIDRetrieval}
Awad G, Butt AA, Curtis K, Lee Y, Fiscus J, Godil A, Delgado A, Zhang J, Godard
  E, Diduch L, Smeaton AF, Graham Y, Kraaij W, Qu{\'{e}}not G (2019) {TRECVID
  2019: An evaluation campaign to benchmark Video Activity Detection, Video
  Captioning and Matching, and Video Search {\&} retrieval}. In:
  \emph{TRECVID}, Gaithersburg, Ma, US

\bibitem[{Awad et~al.(2020)Awad, Butt, Curtis, Lee, Fiscus, Godil, Delgado,
  Zhang, Godard, Diduch, Liu, Smeaton, Graham, Jones, Kraaij, and
  Qu{\'{e}}not}]{Awad2020TRECVIDDomains}
Awad G, Butt AA, Curtis K, Lee Y, Fiscus J, Godil A, Delgado A, Zhang J, Godard
  E, Diduch L, Liu J, Smeaton AF, Graham Y, Jones GJF, Kraaij W, Qu{\'{e}}not G
  (2020) {TRECVID 2020: comprehensive campaign for evaluating video retrieval
  tasks across multiple application domains}. In: \emph{TRECVID}, NIST, US

\bibitem[{Bahdanau et~al.(2015)Bahdanau, Cho, and
  Bengio}]{Bahdanau2015NeuralTranslate}
Bahdanau D, Cho K, Bengio Y (2015) {Neural Machine Translation by Jointly
  Learning to Align and Translate}. In: Bengio Y, LeCun Y (eds) \emph{ICLR}

\bibitem[{Banerjee and Lavie(2005)}]{Banerjee2005METEOR:Judgments}
Banerjee S, Lavie A (2005) {METEOR: An Automatic Metric for MT Evaluation with
  Improved Correlation with Human Judgments}. In: \emph{ACL Workshop on
  Intrinsic and Extrinsic Evaluation Measures for Machine Translation and/or
  Summarization}, pp. 65--72

\bibitem[{Baraldi et~al.(2017)Baraldi, Grana, and
  Cucchiara}]{Baraldi2017HierarchicalCaptioning}
Baraldi L, Grana C, Cucchiara R (2017) {Hierarchical Boundary-Aware Neural
  Encoder for Video Captioning}. In: \emph{IEEE CVPR}, IEEE, pp. 3185--3194

\bibitem[{Barbu et~al.(2012)Barbu, Bridge, Burchill, Coroian, Dickinson,
  Fidler, Michaux, Mussman, Narayanaswamy, Salvi, Schmidt, Shangguan, Siskind,
  Waggoner, Wang, Wei, Yin, and Zhang}]{Barbu2012VideoOut}
Barbu A, Bridge A, Burchill Z, Coroian D, Dickinson S, Fidler S, Michaux A,
  Mussman S, Narayanaswamy S, Salvi D, Schmidt L, Shangguan J, Siskind JM,
  Waggoner J, Wang S, Wei J, Yin Y, Zhang Z (2012) {Video In Sentences Out}.
  \url{http://arxiv.org/abs/1204.2742}

\bibitem[{Bin et~al.(2019)Bin, Yang, Shen, Xie, Shen, and
  Li}]{Bin2019DescribingLSTM}
Bin Y, Yang Y, Shen F, Xie N, Shen HT, Li X (2019) {Describing Video With
  Attention-Based Bidirectional LSTM}. \emph{IEEE Transactions on Cybernetics}
  49(7), 2631--2641

\bibitem[{Bojanowski et~al.(2015)Bojanowski, Lajugie, Grave, Bach, Laptev,
  Ponce, and Schmid}]{Bojanowski2015WeaklyText}
Bojanowski P, Lajugie R, Grave E, Bach F, Laptev I, Ponce J, Schmid C (2015)
  {Weakly-Supervised Alignment of Video with Text}. In: \emph{IEEE ICCV}, IEEE,
  pp. 4462--4470

\bibitem[{Buch et~al.(2017)Buch, Escorcia, Shen, Ghanem, and
  Niebles}]{Buch2017SST:Proposals}
Buch S, Escorcia V, Shen C, Ghanem B, Niebles JC (2017) {SST: Single-Stream
  Temporal Action Proposals}. In: \emph{IEEE CVPR}, IEEE, vol 2017-January, pp.
  6373--6382

\bibitem[{Carreira and Zisserman(2017)}]{Carreira2017QuoDataset}
Carreira J, Zisserman A (2017) {Quo Vadis, Action Recognition? A New Model and
  the Kinetics Dataset}. In: \emph{IEEE CVPR}, IEEE, pp. 4724--4733

\bibitem[{Caruana(1998)}]{Caruana1998MultitaskLearning}
Caruana R (1998) {Multitask Learning}. In: Thrun S, Pratt L (eds)
  \emph{Learning to Learn}, Springer US, Boston, MA, pp. 95--133

\bibitem[{Celikyilmaz et~al.(2020)Celikyilmaz, Clark, and
  Gao}]{Celikyilmaz2020EvaluationSurvey}
Celikyilmaz A, Clark E, Gao J (2020) {Evaluation of Text Generation: A Survey}

\bibitem[{Chen et~al.(2020{\natexlab{a}})Chen, Li, and
  Hu}]{Chen2020DelvingCaptioning}
Chen H, Li J, Hu X (2020{\natexlab{a}}) {Delving Deeper into the Decoder for
  Video Captioning}. \emph{CoRR}

\bibitem[{Chen et~al.(2020{\natexlab{b}})Chen, Lin, Maye, Li, and
  Hu}]{Chen2020ASampling}
Chen H, Lin K, Maye A, Li J, Hu X (2020{\natexlab{b}}) {A Semantics-Assisted
  Video Captioning Model Trained with Scheduled Sampling}. \emph{Frontiers in
  Robotic and AI} 7

\bibitem[{Chen et~al.(2017{\natexlab{a}})Chen, Liang, Liu, Chen, Gao, Jin, and
  Hauptmann}]{Chen2017Informedia2017}
Chen J, Liang J, Liu J, Chen S, Gao C, Jin Q, Hauptmann A (2017{\natexlab{a}})
  {Informedia @ TRECVID 2017}. In: \emph{TRECVID}

\bibitem[{Chen et~al.(2018{\natexlab{a}})Chen, Chen, Jin, and
  Hauptmann}]{Chen2018InformediaTRECVID2018}
Chen J, Chen S, Jin Q, Hauptmann A (2018{\natexlab{a}}) {Informedia@TRECVID
  2018}. In: \emph{TRECVID}

\bibitem[{Chen et~al.(2019{\natexlab{a}})Chen, Pan, Li, Yao, Chao, and
  Mei}]{Chen2019TemporalCaptioning}
Chen J, Pan Y, Li Y, Yao T, Chao H, Mei T (2019{\natexlab{a}}) {Temporal
  Deformable Convolutional Encoder-Decoder Networks for Video Captioning}. In:
  \emph{AAAI}, AAAI, vol~33, pp. 8167--8174

\bibitem[{Chen et~al.(2017{\natexlab{b}})Chen, Chen, Jin, and
  Hauptmann}]{Chen2017VideoTopics}
Chen S, Chen J, Jin Q, Hauptmann A (2017{\natexlab{b}}) {Video Captioning with
  Guidance of Multimodal Latent Topics}. In: \emph{ACM MM}, ACM Press, New
  York, New York, USA, pp. 1838--1846

\bibitem[{Chen et~al.(2018{\natexlab{b}})Chen, Song, Zhao, Qiu, Jin, and
  Hauptmann}]{Chen2018RUC+CMU:Videos}
Chen S, Song Y, Zhao Y, Qiu J, Jin Q, Hauptmann A (2018{\natexlab{b}})
  {RUC+CMU: System Report for Dense Captioning Events in Videos}. \emph{CoRR}
  abs/1806.0

\bibitem[{Chen et~al.(2019{\natexlab{b}})Chen, Jin, Chen, and
  Hauptmann}]{Chen2019GeneratingGuidance}
Chen S, Jin Q, Chen J, Hauptmann A (2019{\natexlab{b}}) {Generating Video
  Descriptions with Latent Topic Guidance}. \emph{IEEE Transactions on
  Multimedia} 21, 2407--2418

\bibitem[{Chen et~al.(2019{\natexlab{c}})Chen, Song, Zhao, Jin, Zeng, Liu, Fu,
  and Hauptmann}]{Chen2019ActivitynetVideos}
Chen S, Song Y, Zhao Y, Jin Q, Zeng Z, Liu B, Fu J, Hauptmann A
  (2019{\natexlab{c}}) {Activitynet 2019 Task 3: Exploring Contexts for Dense
  Captioning Events in Videos}. \url{http://arxiv.org/abs/1907.05092}

\bibitem[{Chen et~al.(2020{\natexlab{c}})Chen, Zhao, Jin, and
  Wu}]{Chen2020Fine-grainedReasoning}
Chen S, Zhao Y, Jin Q, Wu Q (2020{\natexlab{c}}) {Fine-grained Video-Text
  Retrieval with Hierarchical Graph Reasoning}. In: \emph{IEEE/CVF CVPR}

\bibitem[{Chen and Zitnick(2015)}]{Chen2015MindsGeneration}
Chen X, Zitnick CL (2015) {Mind's eye: A recurrent visual representation for
  image caption generation}. In: \emph{IEEE CVPR}, IEEE, vol 07-12-June, pp.
  2422--2431

\bibitem[{Chen et~al.(2015)Chen, Fang, Lin, Vedantam, Gupta, Dollar, and
  Zitnick}]{Chen2015MicrosoftServer}
Chen X, Fang H, Lin TY, Vedantam R, Gupta S, Dollar P, Zitnick CL (2015)
  {Microsoft COCO Captions: Data Collection and Evaluation Server}. \emph{CoRR}
  abs/1504.0

\bibitem[{Chen et~al.(2019{\natexlab{d}})Chen, Rohrbach, and
  Parikh}]{Chen2019Cycle-ConsistencyAnswering}
Chen X, Rohrbach M, Parikh D (2019{\natexlab{d}}) {Cycle-Consistency for Robust
  Visual Question Answering}. In: \emph{IEEE/CVF CVPR}, pp. 6649--6658

\bibitem[{Chen et~al.(2018{\natexlab{c}})Chen, Wang, Zhang, and
  Huang}]{Chen2018LessCaptioning}
Chen Y, Wang S, Zhang W, Huang Q (2018{\natexlab{c}}) {Less Is More: Picking
  Informative Frames for Video Captioning}. In: \emph{ECCV}, Springer
  International Publishing, pp. 367--384

\bibitem[{Cho et~al.(2014)Cho, van Merrienboer, Gulcehre, Bahdanau, Bougares,
  Schwenk, and Bengio}]{Cho2014LearningTranslation}
Cho K, van Merrienboer B, Gulcehre C, Bahdanau D, Bougares F, Schwenk H, Bengio
  Y (2014) {Learning Phrase Representations using RNN Encoder–Decoder for
  Statistical Machine Translation}. In: \emph{EMNLP}, ACL, Stroudsburg, PA,
  USA, pp. 1724--1734

\bibitem[{Craswell(2009)}]{Craswell2009MeanRank}
Craswell N (2009) {Mean Reciprocal Rank}. In: LIU L (ed) \emph{Encyclopedia of
  Database Systems}, Springer US, Boston, MA, pp. 1703--1703

\bibitem[{Dai et~al.(2016)Dai, Li, He, and Sun}]{Dai2016R-FCN:Networks}
Dai J, Li Y, He K, Sun J (2016) {R-FCN: Object Detection via Region-based Fully
  Convolutional Networks}. In: \emph{NIPS}, Barcelona, Spain, NIPS’16

\bibitem[{Dalal and Triggs(2005)}]{Dalal2005HistogramsDetection}
Dalal N, Triggs B (2005) {Histograms of oriented gradients for human
  detection}. In: \emph{IEEE Computer Society CVPR}, vol~I, pp. 886--893

\bibitem[{Das et~al.(2013)Das, Xu, Doell, and Corso}]{Das2013AStitching}
Das P, Xu C, Doell RF, Corso JJ (2013) {A thousand frames in just a few words:
  Lingual description of videos through latent topics and sparse object
  stitching}. In: \emph{IEEE Computer Society CVPR}, IEEE, Portland, OR, USA,
  pp. 2634--2641

\bibitem[{Davis and Mermelstein(1980)}]{Davis1980ComparisonSentences}
Davis S, Mermelstein P (1980) {Comparison of parametric representations for
  monosyllabic word recognition in continuously spoken sentences}. \emph{IEEE
  Transactions on Acoustics, Speech, and Signal Processing} 28(4), 357--366

\bibitem[{Deshpande et~al.(2019)Deshpande, Aneja, Wang, Schwing, and
  Forsyth}]{Deshpande2019FastPart-Of-Speech}
Deshpande A, Aneja J, Wang L, Schwing AG, Forsyth D (2019) {Fast, Diverse and
  Accurate Image Captioning Guided by Part-Of-Speech}. In: \emph{IEEE/CVF
  CVPR}, IEEE, pp. 10687--10696

\bibitem[{Dollar et~al.(2005)Dollar, Rabaud, Cottrell, and
  Belongie}]{Dollar2005BehaviorFeatures}
Dollar P, Rabaud V, Cottrell G, Belongie S (2005) {Behavior Recognition via
  Sparse Spatio-Temporal Features}. In: \emph{IEEE International Workshop on
  Visual Surveillance and Performance Evaluation of Tracking and Surveillance},
  IEEE, pp. 65--72

\bibitem[{Donahue et~al.(2014)Donahue, Jia, Vinyals, Hoffman, Zhang, Tzeng, and
  Darrell}]{Donahue2014DeCAF:Recognition}
Donahue J, Jia Y, Vinyals O, Hoffman J, Zhang N, Tzeng E, Darrell T (2014)
  {DeCAF: A Deep Convolutional Activation Feature for Generic Visual
  Recognition}. In: \emph{ICML}, JMLR.org, Beijing, China

\bibitem[{Donahue et~al.(2015)Donahue, Hendricks, Rohrbach, Venugopalan,
  Guadarrama, Saenko, and Darrell}]{Donahue2015Long-TermDescription}
Donahue J, Hendricks LA, Rohrbach M, Venugopalan S, Guadarrama S, Saenko K,
  Darrell T (2015) {Long-Term Recurrent Convolutional Networks for Visual
  Recognition and Description}. \emph{IEEE Transactions on Pattern Analysis and
  Machine Intelligence} 39(4), 677--691

\bibitem[{Dong et~al.(2016)Dong, Li, and
  Snoek}]{Dong2016Word2VisualVec:Prediction}
Dong J, Li X, Snoek CGM (2016) {Word2VisualVec: Image and Video to Sentence
  Matching by Visual Feature Prediction}. \emph{CoRR} abs/1604.0

\bibitem[{Dong et~al.(2018)Dong, Li, and Snoek}]{Dong2018PredictingRetrieval}
Dong J, Li X, Snoek CGM (2018) {Predicting Visual Features From Text for Image
  and Video Caption Retrieval}. \emph{IEEE Transactions on Multimedia} 20(12),
  3377--3388

\bibitem[{Dong et~al.(2019)Dong, Li, Xu, Ji, He, Yang, and
  Wang}]{Dong2019DualRetrieval}
Dong J, Li X, Xu C, Ji S, He Y, Yang G, Wang X (2019) {Dual Encoding for
  Zero-Example Video Retrieval}. In: \emph{IEEE/CVF CVPR}, IEEE, pp. 9338--9347

\bibitem[{Dwibedi et~al.(2019)Dwibedi, Aytar, Tompson, Sermanet, and
  Zisserman}]{Dwibedi2019TemporalLearning}
Dwibedi D, Aytar Y, Tompson J, Sermanet P, Zisserman A (2019) {Temporal
  cycle-consistency learning}. In: \emph{IEEE/CVF CVPR}, IEEE Computer Society,
  vol 2019-June, pp. 1801--1810

\bibitem[{Eisenstein(2019)}]{Eisenstein2019IntroductionProcessing}
Eisenstein J (2019) \emph{{Introduction to Natural Language Processing}}. MIT
  Press

\bibitem[{Elhamifar et~al.(2016)Elhamifar, Sapiro, and
  Sastry}]{Elhamifar2016Dissimilarity-basedSelection}
Elhamifar E, Sapiro G, Sastry SS (2016) {Dissimilarity-based sparse subset
  selection}. \emph{IEEE Transactions on Pattern Analysis and Machine
  Intelligence} 38(11), 2182--2197

\bibitem[{Faghri et~al.(2018)Faghri, Fleet, Kiros, and
  Fidler}]{Faghri2018VSE++:Negatives}
Faghri F, Fleet DJ, Kiros JR, Fidler S (2018) {VSE++: Improving Visual-Semantic
  Embeddings with Hard Negatives}. In: \emph{BMVC}

\bibitem[{Fang et~al.(2015)Fang, Gupta, Iandola, Srivastava, Deng, Dollar, Gao,
  He, Mitchell, Platt, Zitnick, and Zweig}]{Fang2015FromBack}
Fang H, Gupta S, Iandola F, Srivastava RK, Deng L, Dollar P, Gao J, He X,
  Mitchell M, Platt JC, Zitnick CL, Zweig G (2015) {From captions to visual
  concepts and back}. In: \emph{IEEE CVPR}, IEEE, vol 07-12-June, pp.
  1473--1482

\bibitem[{Feichtenhofer et~al.(2017)Feichtenhofer, Pinz, and
  Wildes}]{Feichtenhofer2017SpatiotemporalRecognition}
Feichtenhofer C, Pinz A, Wildes RP (2017) {Spatiotemporal Multiplier Networks
  for Video Action Recognition}. In: \emph{IEEE CVPR}, IEEE, pp. 7445--7454

\bibitem[{Gan et~al.(2017{\natexlab{a}})Gan, Gan, He, Gao, and
  Deng}]{Gan2017StyleNet:Styles}
Gan C, Gan Z, He X, Gao J, Deng L (2017{\natexlab{a}}) {StyleNet: Generating
  Attractive Visual Captions with Styles}. In: \emph{IEEE CVPR}, IEEE, pp.
  955--964

\bibitem[{Gan et~al.(2017{\natexlab{b}})Gan, Gan, He, Pu, Tran, Gao, Carin, and
  Deng}]{Gan2017SemanticCaptioning}
Gan Z, Gan C, He X, Pu Y, Tran K, Gao J, Carin L, Deng L (2017{\natexlab{b}})
  {Semantic Compositional Networks for Visual Captioning}. In: \emph{IEEE
  CVPR}, IEEE, vol 2017-Janua, pp. 1141--1150

\bibitem[{Gao et~al.(2017)Gao, Guo, Zhang, Xu, and
  Shen}]{Gao2017VideoConsistency}
Gao L, Guo Z, Zhang H, Xu X, Shen HT (2017) {Video Captioning with
  Attention-Based LSTM and Semantic Consistency}. \emph{IEEE Transactions on
  Multimedia} 19(9)

\bibitem[{Gao et~al.(2019)Gao, Li, Song, and
  Shen}]{Gao2019HierarchicalCaptioning}
Gao L, Li X, Song J, Shen HT (2019) {Hierarchical LSTMs with Adaptive Attention
  for Visual Captioning}. \emph{IEEE Transactions on Pattern Analysis and
  Machine Intelligence} pp. 1--19

\bibitem[{Gatt and Krahmer(2018)}]{Gatt2018SurveyEvaluation}
Gatt A, Krahmer E (2018) {Survey of the State of the Art in Natural Language
  Generation: Core tasks, applications and evaluation}. \emph{Journal of
  Artificial Intelligence Research} 61, 65--170

\bibitem[{Ging et~al.(2020)Ging, Zolfaghari, Pirsiavash, and
  Brox}]{Ging2020COOT:Learning}
Ging S, Zolfaghari M, Pirsiavash H, Brox T (2020) {COOT: Cooperative
  Hierarchical Transformer for Video-Text Representation Learning}. In:
  \emph{NIPS}

\bibitem[{Girshick(2015)}]{Girshick2015FastR-CNN}
Girshick R (2015) {Fast R-CNN}. In: \emph{IEEE ICCV}, IEEE, pp. 1440--1448

\bibitem[{Girshick et~al.(2014)Girshick, Donahue, Darrell, and
  Malik}]{Girshick2014RichSegmentation}
Girshick R, Donahue J, Darrell T, Malik J (2014) {Rich Feature Hierarchies for
  Accurate Object Detection and Semantic Segmentation}. In: \emph{IEEE CVPR},
  IEEE, pp. 580--587

\bibitem[{Goodfellow et~al.(2016)Goodfellow, Bengio, and
  Courville}]{Goodfellow2016DeepLearning}
Goodfellow I, Bengio Y, Courville A (2016) \emph{{Deep Learning}}. The MIT
  Press

\bibitem[{Graham et~al.(2018)Graham, Awad, and
  Smeaton}]{Graham2018EvaluationAssessment}
Graham Y, Awad G, Smeaton A (2018) {Evaluation of automatic video captioning
  using direct assessment}. \emph{PLOS ONE} 13(9), e0202789

\bibitem[{Graves et~al.(2013)Graves, Mohamed, and
  Hinton}]{Graves2013SpeechNetworks}
Graves A, Mohamed Ar, Hinton G (2013) {Speech recognition with deep recurrent
  neural networks}. In: \emph{IEEE ICASSP}, IEEE, pp. 6645--6649

\bibitem[{Guadarrama et~al.(2013)Guadarrama, Krishnamoorthy, Malkarnenkar,
  Venugopalan, Mooney, Darrell, and
  Saenko}]{Guadarrama2013Youtube2text:Recognition}
Guadarrama S, Krishnamoorthy N, Malkarnenkar G, Venugopalan S, Mooney R,
  Darrell T, Saenko K (2013) {YouTube2Text: Recognizing and Describing
  Arbitrary Activities Using Semantic Hierarchies and Zero-Shot Recognition}.
  In: \emph{IEEE ICCV}, IEEE, vol~1, pp. 2712--2719

\bibitem[{Guo et~al.(2020)Guo, Yao, and Liu}]{Guo2020SequenceCaptioning}
Guo Y, Yao B, Liu Y (2020) {Sequence to Sequence Model for Video Captioning}.
  \emph{Pattern Recognition Letters} pp. 327--334

\bibitem[{Han et~al.(2013)Han, Kashyap, Finin, Mayfield, and
  Weese}]{Han2013UMBC_EBIQUITY-CORE:Systems}
Han L, Kashyap AL, Finin T, Mayfield J, Weese J (2013) {UMBC{\_}EBIQUITY-CORE:
  Semantic Textual Similarity Systems}. In: \emph{Second Joint Conference on
  Lexical and Computational Semantics}

\bibitem[{Hara et~al.(2018)Hara, Kataoka, and Satoh}]{Hara2018CanImageNet}
Hara K, Kataoka H, Satoh Y (2018) {Can Spatiotemporal 3D CNNs Retrace the
  History of 2D CNNs and ImageNet?} In: \emph{IEEE CVPR}

\bibitem[{He et~al.(2015)He, Zhang, Ren, and Sun}]{He2015DelvingClassification}
He K, Zhang X, Ren S, Sun J (2015) {Delving Deep into Rectifiers: Surpassing
  Human-Level Performance on ImageNet Classification}. In: \emph{IEEE ICCV},
  IEEE, pp. 1026--1034

\bibitem[{He et~al.(2016)He, Zhang, Ren, and Sun}]{He2016DeepRecognition}
He K, Zhang X, Ren S, Sun J (2016) {Deep Residual Learning for Image
  Recognition}. In: \emph{IEEE CVPR}, IEEE, vol 2016-Decem, pp. 770--778

\bibitem[{He et~al.(2019)He, Shi, Bai, Xia, Zhang, and
  Dong}]{He2019ImageGuidance}
He X, Shi B, Bai X, Xia GS, Zhang Z, Dong W (2019) {Image Caption Generation
  with Part of Speech Guidance}. \emph{Pattern Recognition Letters} 119,
  229--237

\bibitem[{Heilbron et~al.(2015)Heilbron, Escorcia, Ghanem, and
  Niebles}]{Heilbron2015ActivityNet:Understanding}
Heilbron FC, Escorcia V, Ghanem B, Niebles JC (2015) {ActivityNet: A
  large-scale video benchmark for human activity understanding}. In: \emph{IEEE
  CVPR}, IEEE, pp. 961--970

\bibitem[{Hemalatha and
  Chandra~Sekhar(2020)}]{Hemalatha2020Domain-SpecificCaptioning}
Hemalatha M, Chandra~Sekhar C (2020) {Domain-Specific Semantics Guided Approach
  to Video Captioning}. In: \emph{IEEE WACV}, pp. 1587--1596

\bibitem[{Hendricks et~al.(2017)Hendricks, Wang, Shechtman, Sivic, Darrell, and
  Russell}]{Hendricks2017LocalizingLanguage}
Hendricks LA, Wang O, Shechtman E, Sivic J, Darrell T, Russell B (2017)
  {Localizing Moments in Video with Natural Language}. In: \emph{2017 IEEE
  International Conference on Computer Vision (ICCV)}, IEEE, pp. 5804--5813

\bibitem[{Hochreiter and Schmidhuber(1997)}]{Hochreiter1997LongMemory}
Hochreiter S, Schmidhuber J (1997) {Long Short-Term Memory}. \emph{Neural
  Computation} 9(8), 1735--1780

\bibitem[{Hodosh et~al.(2015)Hodosh, Young, and
  Hockenmaier}]{Hodosh2015FramingAbstract}
Hodosh M, Young P, Hockenmaier J (2015) {Framing Image Description as a Ranking
  Task Data, Models and Evaluation Metrics Extended Abstract}. In:
  \emph{IJCAI}, pp. 4188--4192

\bibitem[{Hou et~al.(2019)Hou, Wu, Zhao, Luo, and Jia}]{Hou2019JointCaptioning}
Hou J, Wu X, Zhao W, Luo J, Jia Y (2019) {Joint Syntax Representation Learning
  and Visual Cue Translation for Video Captioning}. In: \emph{IEEE ICCV}

\bibitem[{Hu et~al.(2019)Hu, Chen, Zha, and Wu}]{Hu2019HierarchicalCaptioning}
Hu Y, Chen Z, Zha ZJ, Wu F (2019) {Hierarchical Global-Local Temporal Modeling
  for Video Captioning}. In: \emph{ACM MM}, ACM, New York, NY, USA, pp.
  774--783

\bibitem[{Ilg et~al.(2017)Ilg, Mayer, Saikia, Keuper, Dosovitskiy, and
  Brox}]{Ilg2017FlowNetNetworks}
Ilg E, Mayer N, Saikia T, Keuper M, Dosovitskiy A, Brox T (2017) {FlowNet 2.0:
  Evolution of Optical Flow Estimation With Deep Networks}. In: \emph{IEEE
  CVPR}, pp. 2462--2470

\bibitem[{Ji et~al.(2013)Ji, Xu, Yang, and Yu}]{Ji20133DRecognition}
Ji S, Xu W, Yang M, Yu K (2013) {3D Convolutional Neural Networks for Human
  Action Recognition}. \emph{IEEE Transactions on Pattern Analysis and Machine
  Intelligence} 35(1), 221--231

\bibitem[{Karpathy and Fei-Fei(2015)}]{Karpathy2015DeepDescriptions}
Karpathy A, Fei-Fei L (2015) {Deep visual-semantic alignments for generating
  image descriptions}. In: \emph{IEEE CVPR}, IEEE, pp. 3128--3137

\bibitem[{Karpathy et~al.(2014)Karpathy, Toderici, Shetty, Leung, Sukthankar,
  and Fei-Fei}]{Karpathy2014Large-ScaleNetworks}
Karpathy A, Toderici G, Shetty S, Leung T, Sukthankar R, Fei-Fei L (2014)
  {Large-Scale Video Classification with Convolutional Neural Networks}. In:
  \emph{IEEE CVPR}, IEEE, Columbus, OH, US, pp. 1725--1732

\bibitem[{Kipf and Welling(2017)}]{Kipf2017Semi-SupervisedNetworks}
Kipf TN, Welling M (2017) {Semi-Supervised Classification with Graph
  Convolutional Networks}. In: \emph{ICLR}, Neptune, Toulon, France

\bibitem[{Kiros et~al.(2014)Kiros, Salakhutdinov, and
  Zemel}]{Kiros2014UnifyingModels}
Kiros R, Salakhutdinov R, Zemel RS (2014) {Unifying visual-semantic embeddings
  with multimodal neural language models}. \url{http://arxiv.org/abs/1411.2539}

\bibitem[{Kojima et~al.(2002)Kojima, Tamura, and
  Fukunaga}]{Kojima2002NaturalActions}
Kojima A, Tamura T, Fukunaga K (2002) {Natural Language Description of Human
  Activities from Video Images Based on Concept Hierarchy of Actions}.
  \emph{International Journal of Computer Vision} 50(2), 171--184

\bibitem[{Kong and Fu(2018)}]{Kong2018HumanSurvey}
Kong Y, Fu Y (2018) {Human Action Recognition and Prediction: A Survey}.
  \url{https://arxiv.org/abs/1806.11230}

\bibitem[{Krishna et~al.(2017{\natexlab{a}})Krishna, Hata, Ren, Fei-Fei, and
  Niebles}]{Krishna2017Dense-CaptioningVideos}
Krishna R, Hata K, Ren F, Fei-Fei L, Niebles JC (2017{\natexlab{a}})
  {Dense-Captioning Events in Videos}. In: \emph{IEEE ICCV}, IEEE, vol
  2017-Octob, pp. 706--715

\bibitem[{Krishna et~al.(2017{\natexlab{b}})Krishna, Zhu, Groth, Johnson, Hata,
  Kravitz, Chen, Kalantidis, Li, Shamma, Bernstein, and
  Fei-Fei}]{Krishna2017VisualAnnotations}
Krishna R, Zhu Y, Groth O, Johnson J, Hata K, Kravitz J, Chen S, Kalantidis Y,
  Li LJ, Shamma DA, Bernstein MS, Fei-Fei L (2017{\natexlab{b}}) {Visual
  Genome: Connecting Language and Vision Using Crowdsourced Dense Image
  Annotations}. \emph{International Journal of Computer Vision} 123(1), 32--73

\bibitem[{Krishnamoorthy et~al.(2013)Krishnamoorthy, Malkarnenkar, Mooney,
  Saenko, and Guadarrama}]{Krishnamoorthy2013GeneratingKnowledge}
Krishnamoorthy N, Malkarnenkar G, Mooney R, Saenko K, Guadarrama S (2013)
  {Generating Natural-Language Video Descriptions Using Text-Mined Knowledge}.
  \emph{NAACL HLT Workshop on Vision and Language} pp. 10--19

\bibitem[{Krizhevsky et~al.(2012)Krizhevsky, Sutskever, and
  Hinton}]{Krizhevsky2012ImageNetNetworks}
Krizhevsky A, Sutskever I, Hinton GE (2012) {ImageNet classification with deep
  convolutional neural networks}. In: \emph{NIPS}, Curran Associates Inc., Lake
  Tahoe, Nevada, vol~1, pp. 1097--1105

\bibitem[{Kuznetsova et~al.(2014)Kuznetsova, Ordonez, Berg, and
  Choi}]{Kuznetsova2014TREETALK:Descriptions}
Kuznetsova P, Ordonez V, Berg T, Choi Y (2014) {TREETALK: Composition and
  Compression of Trees for Image Descriptions}. \emph{Transactions of the ACL}
  2(1), 351--362

\bibitem[{L.~Chen and B.~Dolan(2011)}]{L.Chen2011CollectingEvaluation}
L~Chen D, B~Dolan W (2011) {Collecting highly parallel data for paraphrase
  evaluation}. In: \emph{Annual Meeting of the ACL: Human Language
  Technologies}, ACL, vol~1, pp. 190--200

\bibitem[{Laptev(2005)}]{Laptev2005OnPoints}
Laptev I (2005) {On Space-Time Interest Points}. \emph{International Journal of
  Computer Vision} 64(2-3), 107--123

\bibitem[{Laptev et~al.(2008)Laptev, Marszalek, Schmid, and
  Rozenfeld}]{Laptev2008LearningMovies}
Laptev I, Marszalek M, Schmid C, Rozenfeld B (2008) {Learning realistic human
  actions from movies}. In: \emph{IEEE CVPR}, IEEE, pp. 1--8

\bibitem[{Le et~al.(2016)Le, Phan, Nguyen, Renoust, Nguyen, Hoang, Duc~Ngo,
  Tran, Watanabe, Klinkigt, Hiroike, Duong, Miyao, and
  Ichi~Satoh}]{Le2016NII-HITACHI-UIT2016}
Le DD, Phan S, Nguyen VT, Renoust B, Nguyen TA, Hoang VN, Duc~Ngo T, Tran MT,
  Watanabe Y, Klinkigt M, Hiroike A, Duong DA, Miyao Y, Ichi~Satoh S (2016)
  {NII-HITACHI-UIT at TRECVID 2016}. In: \emph{TRECVID}, p~25

\bibitem[{Le and Mikolov(2014)}]{Le2014DistributedDocuments}
Le Q, Mikolov T (2014) {Distributed Representations of Sentences and
  Documents}. In: \emph{ICML}, JMLR.org, Beijing, China, vol~32, pp. 1188--1196

\bibitem[{Lee et~al.(2019)Lee, Lee, Seong, Kim, Kim, and
  Kim}]{Lee2019CapturingCaptioning}
Lee J, Lee Y, Seong S, Kim K, Kim S, Kim J (2019) {Capturing Long-Range
  Dependencies in Video Captioning}. In: \emph{IEEE ICIP}, IEEE, pp. 1880--1884

\bibitem[{Lei et~al.(2020)Lei, Wang, Shen, Yu, Berg, and
  Bansal}]{Lei2020MART:Captioning}
Lei J, Wang L, Shen Y, Yu D, Berg TL, Bansal M (2020) {MART: Memory-Augmented
  Recurrent Transformer for Coherent Video Paragraph Captioning}. In:
  \emph{Annual Meeting of the ACL}, pp. 2603--2614

\bibitem[{Li et~al.(2018{\natexlab{a}})Li, Pan, and
  Yang}]{Li2018UTS_CETC_D2DCRCTask}
Li G, Pan P, Yang Y (2018{\natexlab{a}}) {UTS{\_}CETC{\_}D2DCRC Submission at
  the TRECVID 2018 Video to Text Description Task}. In: \emph{TRECVID}

\bibitem[{Li et~al.(2019)Li, Song, Liao, and Peng}]{Li2019REVnet:Description}
Li H, Song D, Liao L, Peng C (2019) {REVnet: Bring Reviewing Into Video
  Captioning for a Better Description}. In: \emph{IEEE ICME}, IEEE, pp.
  1312--1317

\bibitem[{Li and Gong(2019)}]{Li2019End-to-EndLearning}
Li L, Gong B (2019) {End-to-End Video Captioning With Multitask Reinforcement
  Learning}. In: \emph{IEEE WACV}, IEEE, pp. 339--348

\bibitem[{Li et~al.(2015)Li, Liao, Lan, Du, and
  Yang}]{Li2015Zero-shotEmbedding}
Li X, Liao S, Lan W, Du X, Yang G (2015) {Zero-shot image tagging by
  Hierarchical semantic embedding}. In: \emph{SIGIR}, ACM, pp. 879--882

\bibitem[{Li et~al.(2018{\natexlab{b}})Li, Dong, Xu, Cao, Wang, and
  Yang}]{Li2018RenminRetrieval}
Li X, Dong J, Xu C, Cao J, Wang X, Yang G (2018{\natexlab{b}}) {Renmin
  University of China and Zhejiang Gongshang University at TRECVID 2018: Deep
  Cross-Modal Embeddings for Video-Text Retrieval}. In: \emph{TRECVID}

\bibitem[{Li et~al.(2016)Li, Song, Cao, Tetreault, Goldberg, Jaimes, and
  Luo}]{Li2016TGIF:Description}
Li Y, Song Y, Cao L, Tetreault J, Goldberg L, Jaimes A, Luo J (2016) {TGIF: A
  New Dataset and Benchmark on Animated GIF Description}. In: \emph{IEEE CVPR},
  IEEE, vol 2016-Decem, pp. 4641--4650

\bibitem[{Li et~al.(2017)Li, Min, Shen, Carlson, and Carin}]{Li2017VideoText}
Li Y, Min MR, Shen D, Carlson D, Carin L (2017) {Video Generation From Text}.
  \url{http://arxiv.org/abs/1710.00421}

\bibitem[{Lin(2004)}]{Lin2004Rouge:Summaries}
Lin Cy (2004) {Rouge: a package for automatic evaluation of summaries}. In:
  \emph{ACL Post-Conference Workshop}, Barcelona, Spain, pp. 25--26

\bibitem[{Lin et~al.(2020)Lin, Gan, and Wang}]{Lin2020Multi-modal2020}
Lin K, Gan Z, Wang L (2020) {Multi-modal Feature Fusion with Feature Attention
  for VATEX Captioning Challenge 2020}. \emph{arXiv}

\bibitem[{Liu et~al.(2018)Liu, Ren, and Yuan}]{Liu2018SibNet:Captioning}
Liu S, Ren Z, Yuan J (2018) {SibNet: Sibling Convolutional Encoder for Video
  Captioning}. In: \emph{ACM MM}, ACM, New York, NY, USA, pp. 1425--1434

\bibitem[{Liu et~al.(2016)Liu, Anguelov, Erhan, Szegedy, Reed, Fu, and
  Berg}]{Liu2016SSD:Detector}
Liu W, Anguelov D, Erhan D, Szegedy C, Reed S, Fu CY, Berg AC (2016) {SSD:
  Single Shot MultiBox Detector}. In: \emph{ECCV}, pp. 21--37

\bibitem[{Long et~al.(2018)Long, Gan, and De~Melo}]{Long2018VideoAttention}
Long X, Gan C, De~Melo G (2018) {Video Captioning with Multi-Faceted
  Attention}. In: \emph{Transactions of the ACL}, pp. 173--184

\bibitem[{Lu et~al.(2020)Lu, Goswami, Rohrbach, Parikh, and
  Lee}]{Lu202012-in-1:Learning}
Lu J, Goswami V, Rohrbach M, Parikh D, Lee S (2020) {12-in-1: Multi-Task Vision
  and Language Representation Learning}. In: \emph{IEEE/CVF CVPR}

\bibitem[{Mahdisoltani et~al.(2018{\natexlab{a}})Mahdisoltani, Berger,
  Gharbieh, Fleet, and Memisevic}]{Mahdisoltani2018Fine-grainedCaptioning}
Mahdisoltani F, Berger G, Gharbieh W, Fleet D, Memisevic R (2018{\natexlab{a}})
  {Fine-grained Video Classification and Captioning}. \emph{CoRR} abs/1804.0

\bibitem[{Mahdisoltani et~al.(2018{\natexlab{b}})Mahdisoltani, Berger,
  Gharbieh, Fleet, and Memisevic}]{Mahdisoltani2018OnLearning}
Mahdisoltani F, Berger G, Gharbieh W, Fleet D, Memisevic R (2018{\natexlab{b}})
  {On the effectiveness of task granularity for transfer learning}. \emph{CoRR}
  abs/1804.0

\bibitem[{Manmadhan and Kovoor(2020)}]{Manmadhan2020VisualReview}
Manmadhan S, Kovoor BC (2020) {Visual question answering: a state-of-the-art
  review}. \emph{Artificial Intelligence Review} 53(8), 5705--5745

\bibitem[{Mao et~al.(2014)Mao, Xu, Yang, Wang, Huang, and
  Yuille}]{Mao2014Deepm-RNN}
Mao J, Xu W, Yang Y, Wang J, Huang Z, Yuille A (2014) {Deep Captioning with
  Multimodal Recurrent Neural Networks (m-RNN)}. \emph{CoRR} abs/1412.6

\bibitem[{Markatopoulou et~al.(2016)Markatopoulou, Moumtzidou, Galanopoulos,
  Mironidis, Kaltsa, Ioannidou, Symeonidis, Avgerinakis, Andreadis,
  Gialampoukidis, Vrochidis, Briassouli, Mezaris, Kompatsiaris, and
  Patras}]{Markatopoulou2016ITI-CERTH2016}
Markatopoulou F, Moumtzidou A, Galanopoulos D, Mironidis T, Kaltsa V, Ioannidou
  A, Symeonidis S, Avgerinakis K, Andreadis S, Gialampoukidis I, Vrochidis S,
  Briassouli A, Mezaris V, Kompatsiaris I, Patras I (2016) {ITI-CERTH
  participation in TRECVID 2016}. In: \emph{TRECVID}

\bibitem[{Marsden et~al.(2016)Marsden, Mohedano, Mcguinness, Calafell,
  Gir{\'{o}}-I-Nieto, O'connor, Zhou, Azevedo, Daudert, Davis, H{\"{u}}rlimann,
  Afli, Du, Ganguly, Li, Way, and Smeaton}]{Marsden2016Dublin2016}
Marsden M, Mohedano E, Mcguinness K, Calafell A, Gir{\'{o}}-I-Nieto X, O'connor
  NE, Zhou J, Azevedo L, Daudert T, Davis B, H{\"{u}}rlimann M, Afli H, Du J,
  Ganguly D, Li W, Way A, Smeaton AF (2016) {Dublin City University and
  Partners' Participation in the INS and VTT Tracks at TRECVid 2016}. In:
  \emph{TRECVID}

\bibitem[{Meister et~al.(2018)Meister, Hur, and Roth}]{Meister2018UnFlow:Loss}
Meister S, Hur J, Roth S (2018) {UnFlow: Unsupervised Learning of Optical Flow
  with a Bidirectional Census Loss}. In: \emph{AAAI}

\bibitem[{Miech et~al.(2018)Miech, Laptev, and Sivic}]{Miech2018LearningData}
Miech A, Laptev I, Sivic J (2018) {Learning a Text-Video Embedding from
  Incomplete and Heterogeneous Data}. \emph{CoRR} abs/1804.0

\bibitem[{Miech et~al.(2019)Miech, Zhukov, Alayrac, Tapaswi, Laptev, and
  Sivic}]{Miech2019HowTo100M:Clips}
Miech A, Zhukov D, Alayrac JB, Tapaswi M, Laptev I, Sivic J (2019) {HowTo100M:
  Learning a Text-Video Embedding by Watching Hundred Million Narrated Video
  Clips}. In: \emph{IEEE/CVF ICCV}, IEEE, pp. 2630--2640

\bibitem[{Miech et~al.(2020)Miech, Alayrac, Smaira, Laptev, Sivic, and
  Zisserman}]{Miech2020End-to-EndVideos}
Miech A, Alayrac JB, Smaira L, Laptev I, Sivic J, Zisserman A (2020)
  {End-to-End Learning of Visual Representations from Uncurated Instructional
  Videos}. In: \emph{IEEE/CVF CVPR}

\bibitem[{Mikolov et~al.(2013)Mikolov, Sutskever, Chen, Corrado, and
  Dean}]{Mikolov2013DistributedCompositionality}
Mikolov T, Sutskever I, Chen K, Corrado G, Dean J (2013) {Distributed
  Representations of Words and Phrases and Their Compositionality}. In:
  \emph{NIPS}, Curran Associates Inc., vol~2, p 3111–3119

\bibitem[{Mithun et~al.(2017)Mithun, Li, Metze, Roy-Chowdhury, Das, and
  Bosch}]{Mithun2017CMU-UCR-BOSCHRETRIEVAL}
Mithun NC, Li JB, Metze F, Roy-Chowdhury AK, Das S, Bosch R (2017)
  {CMU-UCR-BOSCH @ TRECVID 2017: VIDEO TO TEXT RETRIEVAL}. In: \emph{TRECVID}

\bibitem[{Mithun et~al.(2018)Mithun, Li, Metze, and
  Roy-Chowdhury}]{Mithun2018LearningRetrieval}
Mithun NC, Li J, Metze F, Roy-Chowdhury AK (2018) {Learning Joint Embedding
  with Multimodal Cues for Cross-Modal Video-Text Retrieval}. In: \emph{ACM
  ICMR}, ACM, New York, NY, USA, pp. 19--27

\bibitem[{Mithun et~al.(2019)Mithun, Li, Metze, and
  Roy-Chowdhury}]{Mithun2019JointRetrieval}
Mithun NC, Li J, Metze F, Roy-Chowdhury AK (2019) {Joint embeddings with
  multimodal cues for video-text retrieval}. \emph{International Journal of
  Multimedia Information Retrieval} 8(1), 3--18

\bibitem[{Mun et~al.(2019)Mun, Yang, Ren, Xu, and
  Han}]{Mun2019StreamlinedCaptioning}
Mun J, Yang L, Ren Z, Xu N, Han B (2019) {Streamlined Dense Video Captioning}.
  In: \emph{IEEE CVPR}

\bibitem[{Nguyen et~al.(2017{\natexlab{a}})Nguyen, Li, Cheng, Lu, Zhang, Wu,
  and Ngo}]{Nguyen2017VIREOHyperlinking}
Nguyen PA, Li Q, Cheng ZQ, Lu YJ, Zhang H, Wu X, Ngo CW (2017{\natexlab{a}})
  {VIREO @ TRECVID 2017: Video-to-Text, Ad-hoc Video Search and Video
  Hyperlinking}. In: \emph{TRECVID}

\bibitem[{Nguyen et~al.(2017{\natexlab{b}})Nguyen, Sah, and
  Ptucha}]{Nguyen2017MultistreamCaptioning}
Nguyen T, Sah S, Ptucha R (2017{\natexlab{b}}) {Multistream hierarchical
  boundary network for video captioning}. In: \emph{IEEE WNYISPW}, IEEE, pp.
  1--5

\bibitem[{Nina et~al.(2018)Nina, Garcia, Clouse, and
  Yilmaz}]{Nina2018MTLE:Description}
Nina O, Garcia W, Clouse S, Yilmaz A (2018) {MTLE: A Multitask Learning Encoder
  of Visual Feature Representations for Video and Movie Description}.
  \emph{CoRR} abs/1809.0

\bibitem[{Otani et~al.(2016)Otani, Nakashima, Rahtu, Heikkil{\"{a}}, and
  Yokoya}]{Otani2016LearningSearch}
Otani M, Nakashima Y, Rahtu E, Heikkil{\"{a}} J, Yokoya N (2016) {Learning
  Joint Representations of Videos and Sentences with Web Image Search}. In:
  \emph{ECCV}, Springer International Publishing, pp. 651--667

\bibitem[{Pan et~al.(2020)Pan, Cai, Huang, Lee, Gaidon, Adeli, and
  Niebles}]{Pan2020Spatio-TemporalDistillation}
Pan B, Cai H, Huang DA, Lee KH, Gaidon A, Adeli E, Niebles JC (2020)
  {Spatio-Temporal Graph for Video Captioning with Knowledge Distillation}. In:
  \emph{IEEE/CVF CVPR}, pp. 10870--10879

\bibitem[{Pan et~al.(2016{\natexlab{a}})Pan, Xu, Yang, Wu, and
  Zhuang}]{Pan2016HierarchicalCaptioning}
Pan P, Xu Z, Yang Y, Wu F, Zhuang Y (2016{\natexlab{a}}) {Hierarchical
  Recurrent Neural Encoder for Video Representation with Application to
  Captioning}. In: \emph{IEEE CVPR}, pp. 1029--1038

\bibitem[{Pan et~al.(2016{\natexlab{b}})Pan, Mei, Yao, Li, and
  Rui}]{Pan2016JointlyLanguage}
Pan Y, Mei T, Yao T, Li H, Rui Y (2016{\natexlab{b}}) {Jointly Modeling
  Embedding and Translation to Bridge Video and Language}. In: \emph{IEEE
  CVPR}, IEEE, pp. 4594--4602

\bibitem[{Pan et~al.(2017)Pan, Yao, Li, and Mei}]{Pan2017VideoAttributes}
Pan Y, Yao T, Li H, Mei T (2017) {Video Captioning with Transferred Semantic
  Attributes}. In: \emph{IEEE CVPR}, IEEE, vol 2017-Janua, pp. 984--992

\bibitem[{Papineni et~al.(2002)Papineni, Roukos, Ward, and
  Zhu}]{Papineni2002BLEU:Translation}
Papineni K, Roukos S, Ward T, Zhu WJ (2002) {BLEU: a method for automatic
  evaluation of machine translation}. In: \emph{ACL}, ACL, Morristown, NJ, USA,
  no. July in ACL ’02, p 311

\bibitem[{Parkhi et~al.(2015)Parkhi, Vedaldi, and
  Zisserman}]{Parkhi2015DeepRecognition}
Parkhi OM, Vedaldi A, Zisserman A (2015) {Deep Face Recognition}. In:
  \emph{BMVC}, British Machine Vision Association, pp. 1--41

\bibitem[{Pasunuru and Bansal(2017)}]{Pasunuru2017ReinforcedRewards}
Pasunuru R, Bansal M (2017) {Reinforced Video Captioning with Entailment
  Rewards}. In: \emph{EMNLP}, ACL, Stroudsburg, PA, USA, pp. 979--985

\bibitem[{Perez-Martin et~al.(2020{\natexlab{a}})Perez-Martin, Bustos, and
  P{\'{e}}rez}]{Perez-Martin2020AttentiveCaptioning}
Perez-Martin J, Bustos B, P{\'{e}}rez J (2020{\natexlab{a}}) {Attentive Visual
  Semantic Specialized Network for Video Captioning}. In: \emph{ICPR}

\bibitem[{Perez-Martin et~al.(2020{\natexlab{b}})Perez-Martin, Bustos,
  P{\'{e}}rez, and Barrios}]{Perez-Martin2020IMFD-IMPRESEEEmbedding}
Perez-Martin J, Bustos B, P{\'{e}}rez J, Barrios JM (2020{\natexlab{b}})
  {IMFD-IMPRESEE at TRECVID 2020: Description Generation by Visual-Syntactic
  Embedding}. In: \emph{TRECVID}

\bibitem[{Perez-Martin et~al.(2021)Perez-Martin, Bustos, and
  P{\'{e}}rez}]{Perez-Martin2021ImprovingEmbedding}
Perez-Martin J, Bustos B, P{\'{e}}rez J (2021) {Improving Video Captioning with
  Temporal Composition of a Visual-Syntactic Embedding}. In: \emph{IEEE/CVF
  WACV}

\bibitem[{Phan et~al.(2017{\natexlab{a}})Phan, Henter, Miyao, and
  Satoh}]{Phan2017Consensus-basedCaptioning}
Phan S, Henter GE, Miyao Y, Satoh S (2017{\natexlab{a}}) {Consensus-based
  Sequence Training for Video Captioning}. \emph{CoRR} abs/1712.0

\bibitem[{Phan et~al.(2017{\natexlab{b}})Phan, Klinkigt, Nguyen, Mai,
  Xalabarder, Hinami, Renoust, Duc~Ngo, Tran, Watanabe, Hiroike, Duong, Le,
  Miyao, and Ichi~Satoh}]{Phan2017NII-Hitachi-UIT2017}
Phan S, Klinkigt M, Nguyen VT, Mai TD, Xalabarder AG, Hinami R, Renoust B,
  Duc~Ngo T, Tran MT, Watanabe Y, Hiroike A, Duong DA, Le DD, Miyao Y,
  Ichi~Satoh S (2017{\natexlab{b}}) {NII-Hitachi-UIT at TRECVID 2017}. In:
  \emph{TRECVID}, p~18

\bibitem[{Plummer et~al.(2017)Plummer, Brown, and
  Lazebnik}]{Plummer2017EnhancingEmbedding}
Plummer BA, Brown M, Lazebnik S (2017) {Enhancing Video Summarization via
  Vision-Language Embedding}. In: \emph{IEEE CVPR}, IEEE, pp. 1052--1060

\bibitem[{Ranzato et~al.(2016)Ranzato, Chopra, Auli, and
  Zaremba}]{Ranzato2016SequenceNetworks}
Ranzato M, Chopra S, Auli M, Zaremba W (2016) {Sequence Level Training with
  Recurrent Neural Networks}. In: \emph{ICLR}

\bibitem[{Rashtchian et~al.(2010)Rashtchian, Young, Hodosh, and
  Hockenmaier}]{Rashtchian2010CollectingTurk}
Rashtchian C, Young P, Hodosh M, Hockenmaier J (2010) {Collecting Image
  Annotations Using Amazon's Mechanical Turk}. In: \emph{NAACL HLT 2010
  Workshop on Creating Speech and Language Data with Amazon's Mechanical Turk},
  ACL, Los Angeles, California, pp. 139--147

\bibitem[{Redmon et~al.(2016)Redmon, Divvala, Girshick, and
  Farhadi}]{Redmon2016YouDetection}
Redmon J, Divvala S, Girshick R, Farhadi A (2016) {You Only Look Once: Unified,
  Real-Time Object Detection}. In: \emph{IEEE CVPR}, IEEE, pp. 779--788

\bibitem[{Regneri et~al.(2013)Regneri, Rohrbach, Wetzel, Thater, Schiele, and
  Pinkal}]{Regneri2013GroundingVideos}
Regneri M, Rohrbach M, Wetzel D, Thater S, Schiele B, Pinkal M (2013)
  {Grounding Action Descriptions in Videos}. \emph{Transactions of the ACL} 1,
  25--36

\bibitem[{Reiter(2018)}]{Reiter2018ABLEU}
Reiter E (2018) {A structured review of the validity of BLEU}.
  \emph{Computational Linguistics} 44(3), 393--401

\bibitem[{Reiter and Dale(2000)}]{Reiter2000BuildingSystems}
Reiter E, Dale R (2000) \emph{{Building natural language generation systems}}.
  Cambridge University Press

\bibitem[{Ren et~al.(2017)Ren, He, Girshick, and Sun}]{Ren2017FasterNetworks}
Ren S, He K, Girshick R, Sun J (2017) {Faster R-CNN: Towards Real-Time Object
  Detection with Region Proposal Networks}. \emph{IEEE Transactions on Pattern
  Analysis and Machine Intelligence} 39(6), 1137--1149

\bibitem[{Rijsbergen(1979)}]{Rijsbergen1979InformationRetrieval}
Rijsbergen CJV (1979) \emph{{Information Retrieval}}. Butterworth-Heinemann313
  Washington Street Newton, MAUnited States

\bibitem[{Rohrbach et~al.(2014)Rohrbach, Rohrbach, Qiu, Friedrich, Pinkal, and
  Schiele}]{Rohrbach2014CoherentDetail}
Rohrbach A, Rohrbach M, Qiu W, Friedrich A, Pinkal M, Schiele B (2014)
  {Coherent multi-sentence video description with variable level of detail}.
  In: \emph{Pattern Recognition}, Springer International Publishing, pp.
  184--195

\bibitem[{Rohrbach et~al.(2015{\natexlab{a}})Rohrbach, Rohrbach, and
  Schiele}]{Rohrbach2015TheDescription}
Rohrbach A, Rohrbach M, Schiele B (2015{\natexlab{a}}) {The Long-Short Story of
  Movie Description}. In: \emph{Pattern Recognition}, Springer International
  Publishing, pp. 209--221

\bibitem[{Rohrbach et~al.(2015{\natexlab{b}})Rohrbach, Rohrbach, Tandon, and
  Schiele}]{Rohrbach2015ADescription}
Rohrbach A, Rohrbach M, Tandon N, Schiele B (2015{\natexlab{b}}) {A dataset for
  Movie Description}. In: \emph{IEEE CVPR}, IEEE, vol 07-12-June, pp.
  3202--3212

\bibitem[{Rohrbach et~al.(2017)Rohrbach, Rohrbach, Tang, Oh, and
  Schiele}]{Rohrbach2017GeneratingPeople}
Rohrbach A, Rohrbach M, Tang S, Oh SJ, Schiele B (2017) {Generating
  Descriptions with Grounded and Co-Referenced People}. In: \emph{IEEE CVPR}

\bibitem[{Rohrbach et~al.(2012{\natexlab{a}})Rohrbach, Amin, Andriluka, and
  Schiele}]{Rohrbach2012AActivities}
Rohrbach M, Amin S, Andriluka M, Schiele B (2012{\natexlab{a}}) {A database for
  fine grained activity detection of cooking activities}. In: \emph{IEEE CVPR},
  IEEE, pp. 1194--1201

\bibitem[{Rohrbach et~al.(2012{\natexlab{b}})Rohrbach, Regneri, Andriluka,
  Amin, Pinkal, and Schiele}]{Rohrbach2012ScriptActivities}
Rohrbach M, Regneri M, Andriluka M, Amin S, Pinkal M, Schiele B
  (2012{\natexlab{b}}) {Script Data for Attribute-Based Recognition of
  Composite Activities}. In: \emph{ECCV}, Springer Berlin Heidelberg, Berlin,
  Heidelberg, pp. 144--157

\bibitem[{Rohrbach et~al.(2013)Rohrbach, Qiu, Titov, Thater, Pinkal, and
  Schiele}]{Rohrbach2013TranslatingDescriptions}
Rohrbach M, Qiu W, Titov I, Thater S, Pinkal M, Schiele B (2013) {Translating
  Video Content to Natural Language Descriptions}. In: \emph{IEEE ICCV}, IEEE,
  December, pp. 433--440

\bibitem[{Rohrbach et~al.(2016)Rohrbach, Rohrbach, Regneri, Amin, Andriluka,
  Pinkal, and Schiele}]{Rohrbach2016RecognizingData}
Rohrbach M, Rohrbach A, Regneri M, Amin S, Andriluka M, Pinkal M, Schiele B
  (2016) {Recognizing Fine-Grained and Composite Activities Using Hand-Centric
  Features and Script Data}. \emph{International Journal of Computer Vision}
  119(3), 346--373

\bibitem[{Rumelhart et~al.(1986)Rumelhart, Hinton, and
  Williams}]{Rumelhart1986LearningErrors}
Rumelhart DE, Hinton GE, Williams RJ (1986) {Learning representations by
  back-propagating errors}. \emph{Nature} 323(6088), 533--536

\bibitem[{Sah et~al.(2019)Sah, Nguyen, and
  Ptucha}]{Sah2019UnderstandingCaptioning}
Sah S, Nguyen T, Ptucha R (2019) {Understanding temporal structure for video
  captioning}. \emph{Pattern Analysis and Applications}

\bibitem[{Sah et~al.(2020)Sah, Nguyen, and
  Ptucha}]{Sah2020UnderstandingCaptioning}
Sah S, Nguyen T, Ptucha R (2020) {Understanding temporal structure for video
  captioning}. \emph{Pattern Analysis and Applications} 23(1), 147--159

\bibitem[{Saha et~al.(2017)Saha, Joty, and Al~Hasan}]{Saha2017Con-S2V:Sen2Vec}
Saha TK, Joty S, Al~Hasan M (2017) {Con-S2V: A Generic Framework for
  Incorporating Extra-Sentential Context into Sen2Vec}. In: \emph{Machine
  Learning and Knowledge Discovery in Databases}, Springer International
  Publishing, pp. 753--769

\bibitem[{Schluter(2017)}]{Schluter2017TheROUGE}
Schluter N (2017) {The limits of automatic summarisation according to ROUGE}.
  In: \emph{Conference of the European Chapter of the ACL}, ACL, vol~2, pp.
  41--45

\bibitem[{Schroff et~al.(2015)Schroff, Kalenichenko, and
  Philbin}]{Schroff2015FaceNet:Clustering}
Schroff F, Kalenichenko D, Philbin J (2015) {FaceNet: A unified embedding for
  face recognition and clustering}. In: \emph{IEEE CVPR}, IEEE, vol 07-12-June,
  pp. 815--823

\bibitem[{Sermanet et~al.(2013)Sermanet, Eigen, Zhang, Mathieu, Fergus, and
  LeCun}]{Sermanet2013OverFeat:Networks}
Sermanet P, Eigen D, Zhang X, Mathieu M, Fergus R, LeCun Y (2013) {OverFeat:
  Integrated Recognition, Localization and Detection using Convolutional
  Networks}. \url{https://arxiv.org/abs/1312.6229}

\bibitem[{Shao et~al.(2015)Shao, Kang, Loy, and
  Wang}]{Shao2015DeeplyUnderstanding}
Shao J, Kang K, Loy CC, Wang X (2015) {Deeply learned attributes for crowded
  scene understanding}. In: \emph{IEEE CVPR}, IEEE, vol 07-12-June, pp.
  4657--4666

\bibitem[{Sharif et~al.(2018)Sharif, White, Bennamoun, and
  Shah}]{Sharif2018Learning-basedEvaluation}
Sharif N, White L, Bennamoun M, Shah SAA (2018) {Learning-based composite
  metrics for improved caption evaluation}. In: \emph{ACL, Student Research
  Workshop}, ACL, pp. 14--20

\bibitem[{Shen et~al.(2017)Shen, Li, Su, Li, Chen, Jiang, and
  Xue}]{Shen2017WeaklyCaptioning}
Shen Z, Li J, Su Z, Li M, Chen Y, Jiang YG, Xue X (2017) {Weakly Supervised
  Dense Video Captioning}. In: \emph{IEEE CVPR}, pp. 1916--1924

\bibitem[{Shetty and Laaksonen(2016)}]{Shetty2016Frame-Generation}
Shetty R, Laaksonen J (2016) {Frame- and Segment-Level Features and Candidate
  Pool Evaluation for Video Caption Generation}. In: \emph{ACM MM}, ACM, New
  York, NY, USA, pp. 1073--1076

\bibitem[{Sigurdsson et~al.(2016)Sigurdsson, Varol, Wang, Farhadi, Laptev, and
  Gupta}]{Sigurdsson2016HollywoodUnderstanding}
Sigurdsson GA, Varol G, Wang X, Farhadi A, Laptev I, Gupta A (2016) {Hollywood
  in Homes: Crowdsourcing Data Collection for Activity Understanding}. In:
  Leibe B, Matas J, Sebe N, Welling M (eds) \emph{ECCV}, Springer International
  Publishing, Amsterdam, The Netherlands, pp. 510--526

\bibitem[{Sigurdsson et~al.(2018)Sigurdsson, Gupta, Schmid, Farhadi, and
  Alahari}]{Sigurdsson2018ActorVideos}
Sigurdsson GA, Gupta A, Schmid C, Farhadi A, Alahari K (2018) {Actor and
  Observer: Joint Modeling of First and Third-Person Videos}. In: \emph{IEEE
  CVPR}, pp. 7396--7404

\bibitem[{Simonyan and Zisserman(2015)}]{Simonyan2015VeryRecognition}
Simonyan K, Zisserman A (2015) {Very Deep Convolutional Networks for
  Large-Scale Image Recognition}. In: \emph{ICLR}, San Diego, CA, USA

\bibitem[{Singh et~al.(2020)Singh, Singh, and
  Bandyopadhyay}]{Singh2020NITS-VC2020}
Singh A, Singh TD, Bandyopadhyay S (2020) {NITS-VC System for VATEX Video
  Captioning Challenge 2020}. \emph{arXiv}

\bibitem[{Snoek et~al.(2016)Snoek, Dong, Li, Wang, Wei, Lan, Gavves, Hussein,
  Koelma, and M~Smeulders}]{Snoek2016UniversityVideo}
Snoek CGM, Dong J, Li X, Wang X, Wei Q, Lan W, Gavves E, Hussein N, Koelma DC,
  M~Smeulders AW (2016) {University of Amsterdam and Renmin University at
  TRECVID 2016: Searching Video, Detecting Events and Describing Video}. In:
  \emph{TRECVID}, p~5

\bibitem[{Snoek et~al.(2017{\natexlab{a}})Snoek, Li, Xu, and
  Koelma}]{Snoek2017SearchingVideo}
Snoek CGM, Li X, Xu C, Koelma DC (2017{\natexlab{a}}) {Searching Video,
  Detecting Events and Describing Video}. In: \emph{TRECVID}

\bibitem[{Snoek et~al.(2017{\natexlab{b}})Snoek, Li, Xu, and
  Koelma}]{Snoek2017UniversityVideo}
Snoek CGM, Li X, Xu C, Koelma DC (2017{\natexlab{b}}) {University of Amsterdam
  and Renmin University at TRECVID 2017: Searching Video, Detecting Events and
  Describing Video}. In: \emph{TRECVID}

\bibitem[{Song et~al.(2019{\natexlab{a}})Song, Guo, Gao, Li, Hanjalic, and
  Shen}]{Song2019FromCaptioning}
Song J, Guo Y, Gao L, Li X, Hanjalic A, Shen HT (2019{\natexlab{a}}) {From
  Deterministic to Generative: Multimodal Stochastic RNNs for Video
  Captioning}. \emph{IEEE Transactions on Neural Networks and Learning Systems}
  30(10), 3047--3058

\bibitem[{Song et~al.(2019{\natexlab{b}})Song, Zhao, Chen, and
  Jin}]{Song2019RUC_AIM3Text}
Song Y, Zhao Y, Chen S, Jin Q (2019{\natexlab{b}}) {RUC{\_}AIM3 at TRECVID
  2019: Video to Text}. In: \emph{TRECVID}

\bibitem[{Srivastava et~al.(2015)Srivastava, Mansimov, and
  Salakhutdinov}]{Srivastava2015UnsupervisedLSTMs}
Srivastava N, Mansimov E, Salakhutdinov R (2015) {Unsupervised Learning of
  Video Representations using LSTMs}. In: \emph{ICML}, JMLR.org, Lille, France,
  ICML '15, vol~37, p 843–852

\bibitem[{Srivastava et~al.(2019)Srivastava, Murali, Dubey, and
  Mukherjee}]{Srivastava2019VisualAnalysis}
Srivastava Y, Murali V, Dubey SR, Mukherjee S (2019) {Visual Question Answering
  using Deep Learning: A Survey and Performance Analysis}. \emph{CoRR}
  abs/1909.0

\bibitem[{Sun et~al.(2019{\natexlab{a}})Sun, Myers, Vondrick, Murphy, Schmid,
  and Research}]{Sun2019VideoBERT:Learning}
Sun C, Myers A, Vondrick C, Murphy K, Schmid C, Research G (2019{\natexlab{a}})
  {VideoBERT: A Joint Model for Video and Language Representation Learning}.
  In: \emph{IEEE ICCV}, pp. 7464--7473

\bibitem[{Sun et~al.(2019{\natexlab{b}})Sun, Li, Yuan, Zha, and
  Hu}]{Sun2019MultimodalCaptioning}
Sun L, Li B, Yuan C, Zha Z, Hu W (2019{\natexlab{b}}) {Multimodal Semantic
  Attention Network for Video Captioning}. In: \emph{IEEE ICME}, IEEE, pp.
  1300--1305

\bibitem[{Szegedy et~al.(2015)Szegedy, {Wei Liu}, {Yangqing Jia}, Sermanet,
  Reed, Anguelov, Erhan, Vanhoucke, and
  Rabinovich}]{Szegedy2015GoingConvolutions}
Szegedy C, {Wei Liu}, {Yangqing Jia}, Sermanet P, Reed S, Anguelov D, Erhan D,
  Vanhoucke V, Rabinovich A (2015) {Going deeper with convolutions}. In:
  \emph{IEEE CVPR}, IEEE, vol 07-12-June, pp. 1--9

\bibitem[{Tang et~al.(2019)Tang, Wang, and Li}]{Tang2019RichCaptioning}
Tang P, Wang H, Li Q (2019) {Rich Visual and Language Representation with
  Complementary Semantics for Video Captioning}. \emph{ACM Transactions on
  Multimedia Computing, Communications, and Applications} 15(2), 1--23

\bibitem[{Tapaswi et~al.(2016)Tapaswi, Zhu, Stiefelhagen, Torralba, Urtasun,
  and Fidler}]{Tapaswi2016MovieQA}
Tapaswi M, Zhu Y, Stiefelhagen R, Torralba A, Urtasun R, Fidler S (2016)
  {MovieQA: Understanding Stories in Movies through Question-Answering}. In:
  \emph{IEEE CVPR}, IEEE, pp. 4631--4640

\bibitem[{Thomason et~al.(2014)Thomason, Venugopalan, Guadarrama, Saenko, and
  Mooney}]{Thomason2014IntegratingWild}
Thomason J, Venugopalan S, Guadarrama S, Saenko K, Mooney R (2014) {Integrating
  Language and Vision to Generate Natural Language Descriptions of Videos in
  the Wild}. In: \emph{COLING}, Dublin, Ireland, pp. 1218--1227

\bibitem[{Torabi et~al.(2015)Torabi, Pal, Larochelle, and
  Courville}]{Torabi2015UsingResearch}
Torabi A, Pal C, Larochelle H, Courville A (2015) {Using Descriptive Video
  Services to Create a Large Data Source for Video Annotation Research}.
  \emph{CoRR} abs/1503.0

\bibitem[{Tran et~al.(2018)Tran, Wang, Torresani, Ray, LeCun, and
  Paluri}]{Tran2018ARecognition}
Tran D, Wang H, Torresani L, Ray J, LeCun Y, Paluri M (2018) {A Closer Look at
  Spatiotemporal Convolutions for Action Recognition}. In: \emph{IEEE/CVF
  CVPR}, IEEE, pp. 6450--6459

\bibitem[{Varol et~al.(2018)Varol, Laptev, and
  Schmid}]{Varol2018Long-TermRecognition}
Varol G, Laptev I, Schmid C (2018) {Long-Term Temporal Convolutions for Action
  Recognition}. \emph{IEEE Transactions on Pattern Analysis and Machine
  Intelligence} 40(6), 1510--1517

\bibitem[{Vaswani et~al.(2017)Vaswani, Shazeer, Parmar, Uszkoreit, Jones,
  Gomez, Kaiser, and Polosukhim}]{Vaswani2017AttentionNeed}
Vaswani A, Shazeer N, Parmar N, Uszkoreit J, Jones L, Gomez AN, Kaiser L,
  Polosukhim I (2017) {Attention is all you need}. In: \emph{NIPS}, Curran
  Associates Inc., Long Beach, California, USA, pp. 6000--6010

\bibitem[{Vedantam et~al.(2015)Vedantam, Zitnick, and
  Parikh}]{Vedantam2015CIDEr:Evaluation}
Vedantam R, Zitnick CL, Parikh D (2015) {CIDEr: Consensus-based image
  description evaluation}. In: \emph{IEEE CVPR}, IEEE, pp. 4566--4575

\bibitem[{Venugopalan et~al.(2015{\natexlab{a}})Venugopalan, Rohrbach, Donahue,
  Mooney, Darrell, and Saenko}]{Venugopalan2015SequenceText}
Venugopalan S, Rohrbach M, Donahue J, Mooney R, Darrell T, Saenko K
  (2015{\natexlab{a}}) {Sequence to Sequence -- Video to Text}. In: \emph{IEEE
  ICCV}, IEEE, vol 2015 Inter, pp. 4534--4542

\bibitem[{Venugopalan et~al.(2015{\natexlab{b}})Venugopalan, Xu, Donahue,
  Rohrbach, Mooney, and Saenko}]{Venugopalan2015TranslatingNetworks}
Venugopalan S, Xu H, Donahue J, Rohrbach M, Mooney R, Saenko K
  (2015{\natexlab{b}}) {Translating Videos to Natural Language Using Deep
  Recurrent Neural Networks}. In: \emph{Conference of the North American
  Chapter of the ACL: Human Language Technologies}, ACL, Stroudsburg, PA, USA,
  June, pp. 1494--1504

\bibitem[{Venugopalan et~al.(2016)Venugopalan, Hendricks, Mooney, and
  Saenko}]{Venugopalan2016ImprovingText}
Venugopalan S, Hendricks LA, Mooney R, Saenko K (2016) {Improving LSTM-based
  Video Description with Linguistic Knowledge Mined from Text}. In:
  \emph{EMNLP}, ACL, Stroudsburg, PA, USA, pp. 1961--1966

\bibitem[{Vinyals et~al.(2015)Vinyals, Toshev, Bengio, and
  Erhan}]{Vinyals2015ShowGenerator}
Vinyals O, Toshev A, Bengio S, Erhan D (2015) {Show and tell: A neural image
  caption generator}. In: \emph{IEEE CVPR}, IEEE, vol 07-12-June, pp.
  3156--3164

\bibitem[{Wang et~al.(2018{\natexlab{a}})Wang, Ma, Zhang, and
  Liu}]{Wang2018ReconstructionCaptioning}
Wang B, Ma L, Zhang W, Liu W (2018{\natexlab{a}}) {Reconstruction Network for
  Video Captioning}. In: \emph{IEEE CVPR}, pp. 7622--7631

\bibitem[{Wang et~al.(2019{\natexlab{a}})Wang, Ma, Zhang, Jiang, Wang, and
  Liu}]{Wang2019ControllableNetwork}
Wang B, Ma L, Zhang W, Jiang W, Wang J, Liu W (2019{\natexlab{a}})
  {Controllable Video Captioning with POS Sequence Guidance Based on Gated
  Fusion Network}. In: \emph{IEEE ICCV}

\bibitem[{Wang and Schmid(2013)}]{Wang2013ActionTrajectories}
Wang H, Schmid C (2013) {Action Recognition with Improved Trajectories}. In:
  \emph{IEEE ICCV}, IEEE, pp. 3551--3558

\bibitem[{Wang et~al.(2003)Wang, Divakaran, Vetro, Chang, and
  Sun}]{Wang2003SurveyAnalysis}
Wang H, Divakaran A, Vetro A, Chang SF, Sun H (2003) {Survey of
  compressed-domain features used in audio-visual indexing and analysis}.
  \emph{Journal of Visual Communication and Image Representation} 14, 150--183

\bibitem[{Wang et~al.(2009)Wang, Ullah, Kl{\"{a}}ser, Laptev, and
  Schmid}]{Wang2009EvaluationRecognition}
Wang H, Ullah MM, Kl{\"{a}}ser A, Laptev I, Schmid C (2009) {Evaluation of
  local spatio-temporal features for action recognition}. In: \emph{BMVC},
  British Machine Vision Association, BMVA

\bibitem[{Wang et~al.(2011)Wang, Klaser, Schmid, and
  Liu}]{Wang2011ActionTrajectories}
Wang H, Klaser A, Schmid C, Liu CL (2011) {Action recognition by Dense
  Trajectories}. In: \emph{IEEE CVPR}, IEEE, pp. 3169--3176

\bibitem[{Wang et~al.(2018{\natexlab{b}})Wang, Jiang, Ma, Liu, and
  Xu}]{Wang2018BidirectionalCaptioning}
Wang J, Jiang W, Ma L, Liu W, Xu Y (2018{\natexlab{b}}) {Bidirectional
  Attentive Fusion with Context Gating for Dense Video Captioning}. In:
  \emph{IEEE/CVF CVPR}, IEEE, pp. 7190--7198

\bibitem[{Wang et~al.(2018{\natexlab{c}})Wang, Wang, Huang, Wang, and
  Tan}]{Wang2018M3:Captioning}
Wang J, Wang W, Huang Y, Wang L, Tan T (2018{\natexlab{c}}) {M3: Multimodal
  Memory Modelling for Video Captioning}. In: \emph{IEEE/CVF CVPR}, IEEE, pp.
  7512--7520

\bibitem[{Wang et~al.(2018{\natexlab{d}})Wang, Chen, Wu, Wang, and
  Wang}]{Wang2018VideoLearning}
Wang X, Chen W, Wu J, Wang YF, Wang WY (2018{\natexlab{d}}) {Video Captioning
  via Hierarchical Reinforcement Learning}. In: \emph{IEEE/CVF CVPR}, IEEE, pp.
  4213--4222

\bibitem[{Wang et~al.(2018{\natexlab{e}})Wang, Wang, and
  Wang}]{Wang2018WatchCaptioning}
Wang X, Wang YF, Wang WY (2018{\natexlab{e}}) {Watch, Listen, and Describe:
  Globally and Locally Aligned Cross-Modal Attentions for Video Captioning}.
  In: \emph{Conference of the North American Chapter of the ACL: Human Language
  Technologies}, ACL, Stroudsburg, PA, USA, vol~2, pp. 795--801

\bibitem[{Wang et~al.(2019{\natexlab{b}})Wang, Jabri, and
  Efros}]{Wang2019LearningTime}
Wang X, Jabri A, Efros AA (2019{\natexlab{b}}) {Learning Correspondence from
  the Cycle-consistency of Time}. In: \emph{IEEE/CVF CVPR}, pp. 2566--2576

\bibitem[{Wang et~al.(2019{\natexlab{c}})Wang, Wu, Chen, Li, Wang, and
  Wang}]{Wang2019VaTeXResearch}
Wang X, Wu J, Chen J, Li L, Wang YF, Wang WY (2019{\natexlab{c}}) {VATEX: A
  Large-Scale, High-Quality Multilingual Dataset for Video-and-Language
  Research}. In: \emph{IEEE ICCV}, pp. 4581--4591

\bibitem[{Wei et~al.(2020)Wei, Mi, Hu, and Chen}]{Wei2020ExploitingCaptioning}
Wei R, Mi L, Hu Y, Chen Z (2020) {Exploiting the local temporal information for
  video captioning}. \emph{Journal of Visual Communication and Image
  Representation} 67, 102751

\bibitem[{Weinberger et~al.(2005)Weinberger, Blitzer, and
  Lawrence~K.}]{Weinberger2005DistanceClassification}
Weinberger KQ, Blitzer J, Lawrence~K S (2005) {Distance metric learning for
  large margin nearest neighbor classification}. In: \emph{NIPS}, pp.
  1473--1480

\bibitem[{Wray et~al.(2019)Wray, Csurka, Larlus, and
  Damen}]{Wray2019Fine-GrainedEmbeddings}
Wray M, Csurka G, Larlus D, Damen D (2019) {Fine-Grained Action Retrieval
  Through Multiple Parts-of-Speech Embeddings}. In: \emph{IEEE ICCV}, IEEE, pp.
  450--459

\bibitem[{Wu et~al.(2018)Wu, Li, Cao, Ji, and
  Lin}]{Wu2018InterpretableLocalization}
Wu X, Li G, Cao Q, Ji Q, Lin L (2018) {Interpretable Video Captioning via
  Trajectory Structured Localization}. In: \emph{IEEE/CVF CVPR}, IEEE, pp.
  6829--6837

\bibitem[{Xiao and Shi(2019)}]{Xiao2019ACaptioning}
Xiao H, Shi J (2019) {A Novel Attribute Selection Mechanism for Video
  Captioning}. In: \emph{IEEE ICIP}, IEEE, pp. 619--623

\bibitem[{Xie et~al.(2017)Xie, Girshick, Dollar, Tu, and
  He}]{Xie2017AggregatedNetworks}
Xie S, Girshick R, Dollar P, Tu Z, He K (2017) {Aggregated Residual
  Transformations for Deep Neural Networks}. In: \emph{IEEE CVPR}, IEEE, vol
  2017-Janua, pp. 5987--5995

\bibitem[{Xie et~al.(2018)Xie, Sun, Huang, Tu, and
  Murphy}]{Xie2018RethinkingClassification}
Xie S, Sun C, Huang J, Tu Z, Murphy K (2018) {Rethinking Spatiotemporal Feature
  Learning: Speed-Accuracy Trade-offs in Video Classification}. In:
  \emph{ECCV}, pp. 305--321

\bibitem[{Xu et~al.(2015{\natexlab{a}})Xu, Venugopalan, Ramanishka, Rohrbach,
  and Saenko}]{Xu2015ANetwork}
Xu H, Venugopalan S, Ramanishka V, Rohrbach M, Saenko K (2015{\natexlab{a}}) {A
  Multi-scale Multiple Instance Video Description Network}.
  \url{http://arxiv.org/abs/1505.05914}

\bibitem[{Xu et~al.(2019{\natexlab{a}})Xu, Li, Ramanishka, Sigal, and
  Saenko}]{Xu2019JointStreams}
Xu H, Li B, Ramanishka V, Sigal L, Saenko K (2019{\natexlab{a}}) {Joint Event
  Detection and Description in Continuous Video Streams}. In: \emph{IEEE
  WACVW}, IEEE, pp. 25--26

\bibitem[{Xu et~al.(2016)Xu, Mei, Yao, and Rui}]{Xu2016MSR-VTT:Language}
Xu J, Mei T, Yao T, Rui Y (2016) {MSR-VTT: A Large Video Description Dataset
  for Bridging Video and Language}. \emph{2016 IEEE CVPR} pp. 5288--5296

\bibitem[{Xu et~al.(2015{\natexlab{b}})Xu, Xiong, Chen, and
  Corso}]{Xu2015JointlyFramework}
Xu R, Xiong C, Chen W, Corso JJ (2015{\natexlab{b}}) {Jointly Modeling Deep
  Video and Compositional Text to Bridge Vision and Language in a Unified
  Framework}. In: \emph{AAAI}, pp. 2346--2352

\bibitem[{Xu et~al.(2019{\natexlab{b}})Xu, Yang, and
  Mao}]{Xu2019Semantic-filteredFeature}
Xu Y, Yang J, Mao K (2019{\natexlab{b}}) {Semantic-filtered Soft-Split-Aware
  video captioning with audio-augmented feature}. \emph{Neurocomputing} 357,
  24--35

\bibitem[{Yan et~al.(2020)Yan, Tu, Wang, Zhang, Hao, Zhang, and
  Dai}]{Yan2020STAT:Captioning}
Yan C, Tu Y, Wang X, Zhang Y, Hao X, Zhang Y, Dai Q (2020) {STAT:
  Spatial-Temporal Attention Mechanism for Video Captioning}. \emph{IEEE
  Transactions on Multimedia} 22(1), 229--241

\bibitem[{Yang et~al.(2016)Yang, Zhang, and Xu}]{Yang2016SemanticUnderstanding}
Yang X, Zhang T, Xu C (2016) {Semantic Feature Mining for Video Event
  Understanding}. \emph{ACM Transactions on Multimedia Computing,
  Communications, and Applications} 12(4), 1--22

\bibitem[{Yang et~al.(2018)Yang, Zhou, Ai, Bin, Hanjalic, Shen, and
  Ji}]{Yang2018VideoLSTM}
Yang Y, Zhou J, Ai J, Bin Y, Hanjalic A, Shen HT, Ji Y (2018) {Video Captioning
  by Adversarial LSTM}. \emph{IEEE Transactions on Image Processing} 27(11),
  5600--5611

\bibitem[{Yao et~al.(2015)Yao, Torabi, Cho, Ballas, Pal, Larochelle, and
  Courville}]{Yao2015DescribingStructure}
Yao L, Torabi A, Cho K, Ballas N, Pal C, Larochelle H, Courville A (2015)
  {Describing Videos by Exploiting Temporal Structure}. In: \emph{IEEE ICCV},
  IEEE, pp. 4507--4515

\bibitem[{Yao et~al.(2017)Yao, Li, Qiu, Long, Pan, Li, and
  Mei}]{Yao2017MSRVideos}
Yao T, Li Y, Qiu Z, Long F, Pan Y, Li D, Mei T (2017) {MSR Asia MSM at
  ActivityNet Challenge 2017: Trimmed Action Recognition, Temporal Action
  Proposals and Dense-Captioning Events in Videos}. Tech. rep., Microsoft

\bibitem[{Yosinski et~al.(2014)Yosinski, Clune, Bengio, and
  Lipson}]{Yosinski2014HowNetworks}
Yosinski J, Clune J, Bengio Y, Lipson H (2014) {How Transferable Are Features
  in Deep Neural Networks?} In: \emph{NIPS}, MIT Press, p 3320–3328

\bibitem[{Yu et~al.(2017{\natexlab{a}})Yu, Gao, Li, Dong, and
  Sun}]{Yu2017Shandong2017}
Yu E, Gao M, Li Y, Dong X, Sun J (2017{\natexlab{a}}) {Shandong Normal
  University in the VTT Tasks at TRECVID 2017}. In: \emph{TRECVID}

\bibitem[{Yu et~al.(2015{\natexlab{a}})Yu, Siskind, and
  Lafayette}]{Yu2015LearningInformation}
Yu H, Siskind JM, Lafayette W (2015{\natexlab{a}}) {Learning to Describe Video
  with Weak Supervision by Exploiting Negative Sentential Information}. In:
  \emph{AAAI}, AAAI Press, Austin, Texas, pp. 3855--3863

\bibitem[{Yu et~al.(2016)Yu, Wang, Huang, Yang, and Xu}]{Yu2016VideoNetworks}
Yu H, Wang J, Huang Z, Yang Y, Xu W (2016) {Video Paragraph Captioning Using
  Hierarchical Recurrent Neural Networks}. In: \emph{IEEE CVPR}, IEEE, pp.
  4584--4593

\bibitem[{Yu et~al.(2015{\natexlab{b}})Yu, Park, Berg, and
  Berg}]{Yu2015VisualAnswering}
Yu L, Park E, Berg AC, Berg TL (2015{\natexlab{b}}) {Visual Madlibs: Fill in
  the Blank Description Generation and Question Answering}. In: \emph{IEEE
  ICCV}, IEEE, vol 2015 Inter, pp. 2461--2469

\bibitem[{Yu et~al.(2017{\natexlab{b}})Yu, Choi, Kim, Yoo, Lee, and
  Kim}]{Yu2017SupervisingData}
Yu Y, Choi J, Kim Y, Yoo K, Lee SH, Kim G (2017{\natexlab{b}}) {Supervising
  Neural Attention Models for Video Captioning by Human Gaze Data}. In:
  \emph{IEEE CVPR}, IEEE, pp. 6119--6127

\bibitem[{Yu et~al.(2017{\natexlab{c}})Yu, Ko, Choi, and
  Kim}]{Yu2017EndAnswering}
Yu Y, Ko H, Choi J, Kim G (2017{\natexlab{c}}) {End-to-End Concept Word
  Detection for Video Captioning, Retrieval, and Question Answering}. In:
  \emph{IEEE CVPR}, IEEE, pp. 3261--3269

\bibitem[{Yuan et~al.(2018)Yuan, Tian, Zhang, Ding, and
  Wei}]{Yuan2018VideoGuiding}
Yuan J, Tian C, Zhang X, Ding Y, Wei W (2018) {Video Captioning with Semantic
  Guiding}. In: \emph{IEEE BigMM}, IEEE, pp. 1--5

\bibitem[{Zeng et~al.(2016)Zeng, Chen, Niebles, and Sun}]{Zeng2016TitleVideos}
Zeng KH, Chen TH, Niebles JC, Sun M (2016) {Title Generation for User Generated
  Videos}. In: \emph{ECCV}, Springer International Publishing, pp. 609--625

\bibitem[{Zhang et~al.(2018)Zhang, Hu, and Sha}]{Zhang2018Cross-ModalText}
Zhang B, Hu H, Sha F (2018) {Cross-Modal and Hierarchical Modeling of Video and
  Text}. In: Ferrari V, Hebert M, Sminchisescu C, Weiss Y (eds) \emph{ECCV},
  Springer International Publishing, Cham, pp. 385--401

\bibitem[{Zhang et~al.(2016)Zhang, Pang, Lu, and
  Ngo}]{Zhang2016VIREODescription}
Zhang H, Pang L, Lu YJ, Ngo CW (2016) {VIREO @ TRECVID 2016: Multimedia Event
  Detection, Ad-hoc Video Search, Video-to-Text Description}. In:
  \emph{TRECVID}

\bibitem[{Zhang et~al.(2019{\natexlab{a}})Zhang, Wang, Ma, and
  Liu}]{Zhang2019ReconstructLearning}
Zhang W, Wang B, Ma L, Liu W (2019{\natexlab{a}}) {Reconstruct and Represent
  Video Contents for Captioning via Reinforcement Learning}. \emph{IEEE
  Transactions on Pattern Analysis and Machine Intelligence}

\bibitem[{Zhang et~al.(2017)Zhang, Zhang, Zhang, Li, and
  Qi~Tian}]{Zhang2017Task-DrivenDescription}
Zhang X, Zhang Y, Zhang D, Li J, Qi~Tian A (2017) {Task-Driven Dynamic Fusion:
  Reducing Ambiguity in Video Description}. In: \emph{IEEE CVPR}, IEEE, pp.
  6250--6258

\bibitem[{Zhang et~al.(2019{\natexlab{b}})Zhang, Xu, Ouyang, and
  Tan}]{Zhang2019ShowSummarization}
Zhang Z, Xu D, Ouyang W, Tan C (2019{\natexlab{b}}) {Show, Tell and Summarize:
  Dense Video Captioning Using Visual Cue Aided Sentence Summarization}.
  \emph{IEEE Transactions on Circuits and Systems for Video Technology}

\bibitem[{Zhang et~al.(2020)Zhang, Shi, Yuan, Li, Wang, Hu, and
  Zha}]{Zhang2020ObjectCaptioning}
Zhang Z, Shi Y, Yuan C, Li B, Wang P, Hu W, Zha Z (2020) {Object Relational
  Graph with Teacher-Recommended Learning for Video Captioning}. In:
  \emph{IEEE/CVF CVPR}, pp. 13278--13288

\bibitem[{Zhao et~al.(2019)Zhao, Li, and Lu}]{Zhao2019CAM-RNN:Captioning}
Zhao B, Li X, Lu X (2019) {CAM-RNN: Co-Attention Model based RNN for Video
  Captioning}. \emph{IEEE Transactions on Image Processing} 28, 5552--5565

\bibitem[{Zhao et~al.(2020)Zhao, Song, Chen, and
  Jin}]{Zhao2020RUC_AIM3Description}
Zhao Y, Song Y, Chen S, Jin Q (2020) {RUC{\_}AIM3 at TRECVID 2020: Ad-hoc Video
  Search {\&} Video to Text Description}. In: \emph{TRECVID}

\bibitem[{Zhou et~al.(2018{\natexlab{a}})Zhou, Lapedriza, Khosla, Oliva, and
  Torralba}]{Zhou2018Places:Recognition}
Zhou B, Lapedriza A, Khosla A, Oliva A, Torralba A (2018{\natexlab{a}})
  {Places: A 10 Million Image Database for Scene Recognition}. \emph{IEEE
  Transactions on Pattern Analysis and Machine Intelligence} 40(6), 1452--1464

\bibitem[{Zhou et~al.(2018{\natexlab{b}})Zhou, Xu, and
  Corso}]{Zhou2018TowardsVideos}
Zhou L, Xu C, Corso JJ (2018{\natexlab{b}}) {Towards Automatic Learning of
  Procedures from Web Instructional Videos}. In: \emph{AAAI}, Association for
  the Advancement of Artificial Intelligence, pp. 7590--7598

\bibitem[{Zhou et~al.(2018{\natexlab{c}})Zhou, Zhou, Corso, Socher, and
  Xiong}]{Zhou2018End-to-EndTransformer}
Zhou L, Zhou Y, Corso JJ, Socher R, Xiong C (2018{\natexlab{c}}) {End-to-End
  Dense Video Captioning with Masked Transformer}. In: \emph{IEEE/CVF CVPR},
  IEEE, pp. 8739--8748

\bibitem[{Zhou et~al.(2019)Zhou, Kalantidis, Chen, Corso, and
  Rohrbach}]{Zhou2019GroundedDescription}
Zhou L, Kalantidis Y, Chen X, Corso JJ, Rohrbach M (2019) {Grounded Video
  Description}. In: \emph{IEEE/CVF CVPR}, IEEE, pp. 6571--6580

\bibitem[{Zhu et~al.(2017)Zhu, Park, Isola, and
  Efros}]{Zhu2017UnpairedNetworks}
Zhu JY, Park T, Isola P, Efros AA (2017) {Unpaired Image-to-Image Translation
  Using Cycle-Consistent Adversarial Networks}. In: \emph{IEEE ICCV}, IEEE, vol
  2017-October, pp. 2242--2251

\bibitem[{Zolfaghari et~al.(2018)Zolfaghari, Singh, and
  Brox}]{Zolfaghari2018ECO:Understanding}
Zolfaghari M, Singh K, Brox T (2018) {ECO: Efficient Convolutional Network for
  Online Video Understanding}. In: \emph{ECCV}, Springer International
  Publishing, pp. 713--730

\end{thebibliography}

\end{document}